\documentclass{style/naturep}

\usepackage[utf8]{inputenc}
\usepackage[T1]{fontenc}
\usepackage{lmodern} 
\usepackage{amsmath, amssymb} 
\usepackage{graphicx} 
\usepackage[table,x11names]{xcolor} 
\usepackage{geometry}
\usepackage{ragged2e}
\usepackage{setspace}
\usepackage{booktabs} 
\usepackage{multirow}
\usepackage{tabularx}
\usepackage{makecell}
\usepackage{float}
\usepackage{placeins}
\usepackage{anyfontsize}
\usepackage{enumitem}
\usepackage{kantlipsum}

\usepackage{algorithm}
\usepackage[noend]{algpseudocode}
\usepackage{algorithmicx}

\usepackage{bbm}
\usepackage{pifont}
\usepackage{xspace}
\usepackage{adjustbox}

\usepackage{caption}
\usepackage{subcaption}
\usepackage{standalone}
\usepackage{longtable}

\usepackage[colorlinks=true, linkcolor=blue, citecolor=blue, urlcolor=blue]{hyperref}
\usepackage{xurl}
\usepackage{afterpage} 

\definecolor{DeepGreen}{RGB}{20, 100, 60} 
\definecolor{maroon}{cmyk}{0,0.87,0.68,0.32}
\definecolor{gray}{rgb}{0.3,0.3,0.3}

\newcommand{\hheading}[1]{\noindent\textbf{#1}}
\newcommand\Heading[1]{%
  \vspace{1em}\noindent\textbf{\Large #1}\vspace{0.5em}%
}
\newcommand\heading[1]{%
  \noindent\textbf{\large #1}%
}

\makeatletter
\let\saved@includegraphics\includegraphics
\AtBeginDocument{\let\includegraphics\saved@includegraphics}

\makeatother

\title{
    \centering
    \includegraphics[height=2.2cm]{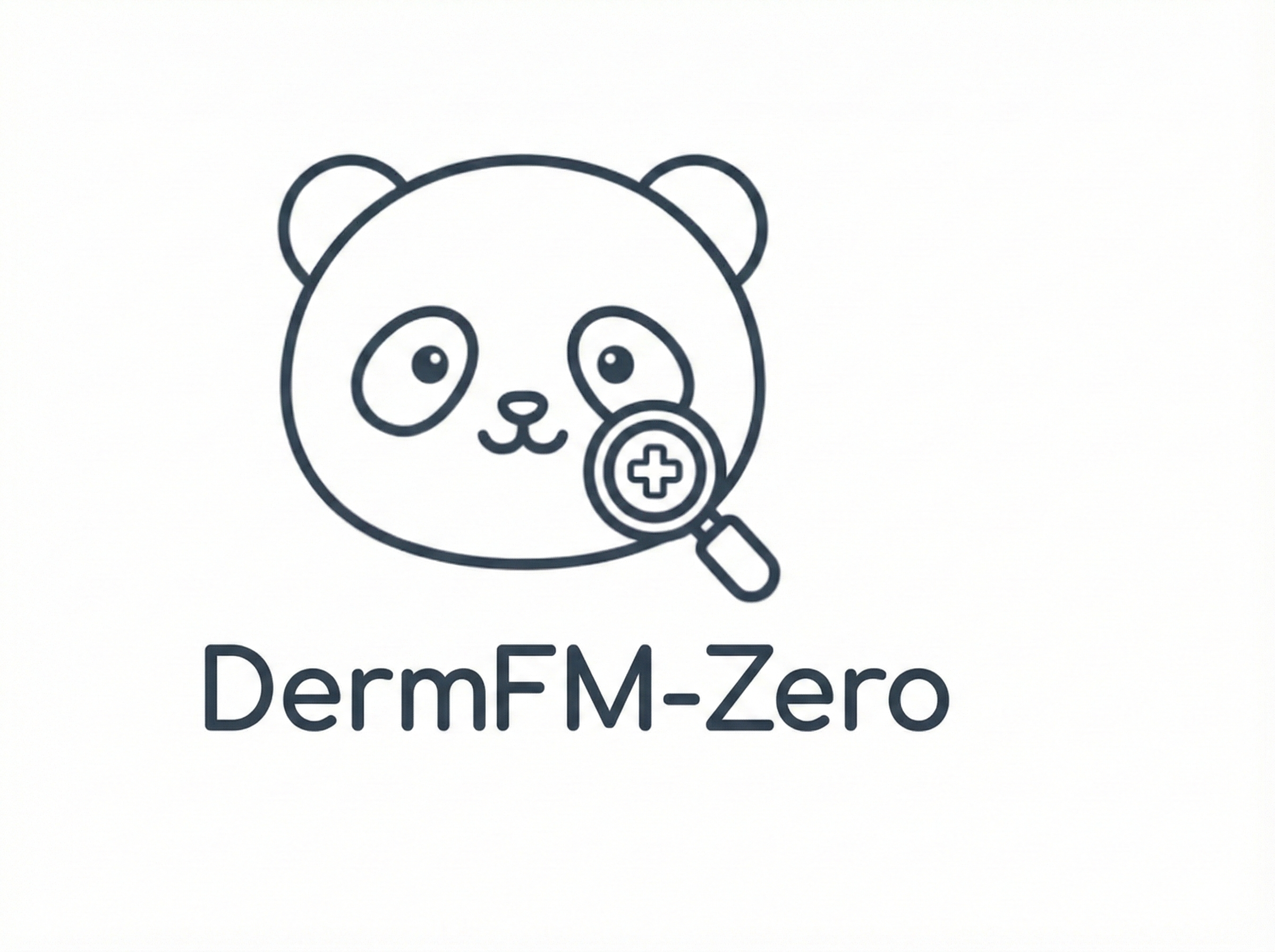} \\[0.1em] 
    
    \begin{spacing}{1.0}
        {\LARGE \sffamily \bfseries \textcolor{DeepGreen}{DermFM-Zero}:} \\[0.1em]
        {\Large \sffamily \bfseries A Vision-Language Foundation Model for Zero-shot Clinical Collaboration and Automated Concept Discovery in Dermatology}
    \end{spacing}
}

\begin{document}

\maketitle
\vspace{-2em} 
\begin{spacing}{1.25} 
\noindent Siyuan Yan$^{1,2*\ddagger}$, Xieji Li$^{1,2*}$, Dan Mo$^{1,2}$,  Philipp Tschandl$^{3}$,  Yiwen Jiang$^{1,4}$, Zhonghua Wang$^{1,2}$, Ming Hu$^{1,4}$, Lie Ju$^{1,5}$, Cristina Vico-Alonso$^{6}$,  Yizhen Zheng$^{2}$, Jiahe Liu$^{1,2}$,  Juexiao Zhou$^{7}$, Camilla Chello$^{8,9}$, Jen G. Cheung$^{10,11}$, Julien Anriot$^{12,13}$, Luc Thomas$^{12,14,15}$, Clare Primiero$^{8}$, Gin Tan$^{16}$, Aik Beng Ng$^{17}$, Simon See$^{17}$, Xiaoying Tang$^{18}$, Albert Ip$^{19}$, Xiaoyang Liao$^{20,21}$, Adrian Bowling$^{22}$, Martin Haskett$^{22}$, Shuang Zhao$^{23,24}$, Monika Janda$^{25}$,  H. Peter Soyer$^{8,26\dag}$, Victoria Mar$^{10,11,27\dag}$, Harald Kittler$^{3\dag}$, Zongyuan Ge$^{1,2,11,28\dag\ddagger}$
\end{spacing}
\vspace{-5mm} 
\begin{spacing}{1.25} 
\small
\begin{affiliations}
 \item AIM for Health Lab, Faculty of Information Technology, Monash University, Melbourne, Australia  
 \item Faculty of Information Technology, Monash University, Melbourne, Australia  
  \item Department of Dermatology, Medical University of Vienna, Vienna, Austria  
   \item Faculty of Engineering, Monash University, Melbourne, Australia  
    \item Institute of Ophthalmology, University College London, London, UK
    \item Dermatology Department. Fundacion Hospital 12 de Octubre. Madrid, Spain
    \item School of Data Science, The Chinese University of Hong Kong, Shenzhen
    \item Frazer Institute, The University of Queensland, Dermatology Research Centre, Brisbane, Australia  
    \item Department of Medical and Cardiovascular Sciences, Sapienza University of Rome, Rome, Italy
    \item Victorian Melanoma Service, Alfred Care Group, Bayside Health, Melbourne, Australia
    \item SkIIN Discovery Program, Monash University, Melbourne, Australia

    \item Claude Bernard Lyon-1 University, Lyon, France
    \item Centre Léon Berard, Lyon, France
    \item Dermatology department, Hôpital Lyon Sud, Hospices Civils de Lyon, Lyons, France
    \item Cancer Research Center of Lyon, Lyons, France  
  \item eResearch Centre, Monash University, Melbourne, Australia
 \item NVIDIA AI Technology Center, Singapore
 \item Department of Electronic and Electrical Engineering, Southern University of Science and Technology, Shenzhen, China
\item Surrey Hills Medical Centre, Melbourne, Australia
\item General Practice Ward/International Medical Center Ward, General Practice Medical Center, West China Hospital, Sichuan University, 610041, Chengdu, Sichuan, China
\item Teaching \& Research Section, General Practice Medical Center, West China Hospital, Sichuan University, 610041, Chengdu, Sichuan, China
\item Independent Researcher, Melbourne, Australia
\item Department of Dermatology, Xiangya Hospital, Central South University, Changsha, 410008, China.
\item Furong Laboratory (Precision Medicine), Changsha, 410008, China.
\item Centre for Health Services Research, Faculty of Medicine, The University of Queensland, Brisbane, Australia
 \item Dermatology Department, Princess Alexandra Hospital, Brisbane, Australia
 \item School of Public Health and Preventive Medicine, Monash University, Melbourne, Australia
 \item Airdoc-Monash Research, Monash University, VIC 3800, Australia
    \\$^*$Equal contribution: Siyuan Yan, Xieji Li
    \\$\dagger$These authors contributed equally as senior authors
\\$^\ddagger$Corresponding author: Siyuan Yan (siyuan.yan@monash.edu); Zongyuan Ge (zongyuan.ge@monash.edu)
 
\end{affiliations}
\end{spacing}

\Heading{Abstract}
\begin{spacing}{1.38}

\noindent Medical foundation models have shown promise in controlled benchmarks, yet widespread deployment remains hindered by reliance on task-specific fine-tuning. Here, we introduce DermFM-Zero, a dermatology vision-language foundation model trained via masked latent modelling and contrastive learning on over 4 million multimodal data points. We evaluated DermFM-Zero across 20 benchmarks spanning zero-shot diagnosis and multimodal retrieval, achieving state-of-the-art performance without task-specific adaptation. We further evaluated its zero-shot capabilities in three multinational reader studies involving over 1,100 clinicians. In primary care settings, AI assistance enabled general practitioners to nearly double their differential diagnostic accuracy across 98 skin conditions. In specialist settings, the model significantly outperformed board-certified dermatologists in multimodal skin cancer assessment. In collaborative workflows, AI assistance enabled non-experts to surpass unassisted experts while improving management appropriateness. Finally, we show that DermFM-Zero's latent representations are interpretable: sparse autoencoders unsupervisedly disentangle clinically meaningful concepts that outperform predefined-vocabulary approaches and enable targeted suppression of artifact-induced biases, enhancing robustness without retraining. These findings demonstrate that foundation models can provide effective, safe, and transparent zero-shot clinical decision support. 
\end{spacing}

\vspace{2em}
\begin{center}
{\small
\textcolor{DeepGreen}{\sffamily \textbf{$\blacktriangleright$~\href{https://github.com/SiyuanYan1/DermFM-Zero}{\textcolor{DeepGreen}{Open-source code repository}}}}
}
\end{center}
\vspace{0.8em}

\newpage
\begin{spacing}{1.35}

\noindent Foundation models (FMs) have markedly advanced medical artificial intelligence, achieving impressive performance in controlled evaluations \cite{path_fm1, panderm, eyefm}. Yet a critical gap remains between these technical achievements and their clinical translation. Current assessments of human-AI collaboration rely predominantly on models with task-specific fine-tuning \cite{panderm,eyecarefm}. While this optimizes performance for defined tasks, it requires curated datasets for every clinical question, which is often impractical, particularly for the long tail of rare conditions. This creates a paradox: foundation models are designed as versatile generalists, yet they are currently evaluated as narrow specialists. In clinical practice, clinicians encounter diverse presentations and cannot pause to retrain models for each scenario. A critical evidence gap thus remains: Can a foundation model meaningfully support clinical decision-making across diverse settings without task-specific customization?

Dermatology provides a particularly rigorous testbed for this paradigm. Unlike clinical tasks that focus on screening for specific high-prevalence pathologies \cite{eyecarefm,radiology_fm}, dermatology encompasses an open-ended distribution of hundreds of conditions with overlapping visual features. Diagnostic workflows vary substantially across healthcare settings, from differential diagnosis in primary care with standard photography documentation \cite{liu2020deep,fair2} to diagnostic stratification of neoplasms in specialist centers using dermoscopy \cite{ham10000,humanai}. In this heterogeneous environment, fine-tuning models for every category, imaging modality, or task is inherently unscalable. This necessitates zero-shot capabilities \cite{clip, chextzero, contch}, defined as the ability of a zero-shot model to adapt to diverse clinical contexts without task-specific training data.

However, two critical limitations impede this zero-shot paradigm. First, existing domain-specific models \cite{panderm, swavderm} remain image-centric and incapable of inference across unseen conditions. Vision-language models, such as MONET \cite{monet} and DermLIP \cite{derm1m}, have emerged but remain constrained by limited training scale and generic architectures that fail to capture disease diversity in the real world. Second, rigorous evaluations of zero-shot models in collaborative clinical workflows have not been conducted. Without such validation, the real-world utility of zero-shot capabilities remains theoretical.

\begin{figure*}[!ht]
\centering
\includegraphics[width=0.8\textwidth]{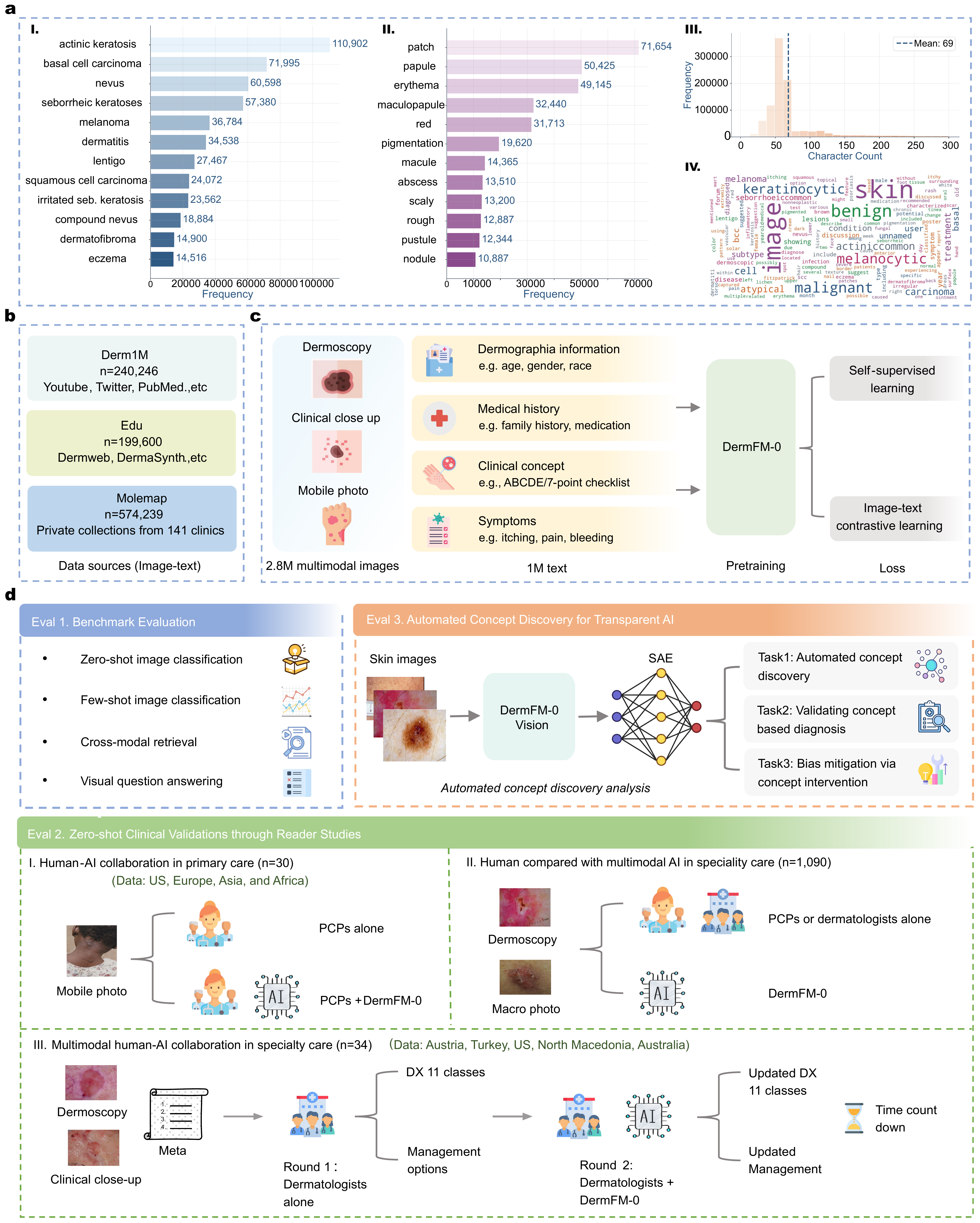}
\caption{\small \textbf{DermFM-Zero pretraining dataset and evaluation framework.} 
\textbf{a, b,} The vision-language pretraining dataset. 
\textbf{a,} Pretraining data statistics, showing: \textbf{(i)} top skin conditions, \textbf{(ii)} clinical concepts, \textbf{(iii)} text length distribution, and \textbf{(iv)} common corpus terms. 
\textbf{b,} Image-text data sources: curated public (Derm1M, n=240,246; Edu, n=199,600) and private (MoleMap; n=574,239) collections. 
\textbf{c,} DermFM-Zero pretraining schematic, using multimodal data (dermoscopy, clinical, mobile photos) and text (demographics, medical history, symptoms) with unimodal self-supervised and multimodal contrastive learning objectives. 
\textbf{d,} The three-stage evaluation framework. 
\textbf{Eval 1:} Evaluation on 17 benchmarks (e.g., zero/few-shot classification, cross-modal retrieval, VQA). 
\textbf{Eval 2:} Clinical validation via three zero-shot reader studies: \textbf{(I)} Human-AI collaboration in primary care (n=30 PCPs vs. PCPs + DermFM-Zero in skin condition differential diagnosis). \textbf{(II)} Standalone AI vs. 1,073 clinicians for multimodal skin cancer diagnosis. \textbf{(III)} Human-AI collaboration in specialty care for skin cancer diagnosis and management (n=34). 
\textbf{Eval 3:} Automated concept discovery for transparent AI applications using SAE. All icons are from Flaticon.com.}
\label{fig1}
\end{figure*}

Here, we developed DermFM-Zero, a multimodal vision-language foundation model pretrained on over 4 million dermatological images and image-text pairs. The training data encompassed diverse multimodal information, including dermoscopy and clinical photography paired with clinical concept descriptions, patient demographics, medical history, and diagnostic terminology (\textbf{Fig.\ref{fig1}a-c}, \textbf{Extended Data Fig.\ref{supp_word_cloud}}, and \textbf{Extended Data Table~\ref{tab:all_disease}}). Through a two-stage training method (\textbf{Extended Data Fig.~\ref{supp_method1}} and \textbf{Methods}), we first employed masked latent modelling \cite{caev2} to enable the vision encoder to capture fine-grained morphological structures from 3 million unlabeled dermatological images. Second, we aligned these visual representations with medical knowledge through bootstrapped contrastive learning on 1 million image-text pairs, paired with a domain-specialized text encoder (PubMedBERT \cite{pubmedbert}). This approach enables DermFM-Zero to function as a zero-shot clinical assistant across diverse healthcare settings.

We evaluated DermFM-Zero through a comprehensive three-stage framework (\textbf{Fig.~\ref{fig1}d}). First, we benchmarked DermFM-Zero against state-of-the-art FMs across diverse zero-shot, few-shot, and multimodal fine-tuning tasks to validate the quality of pretrained representations. Second, we conducted large-scale, multi-country reader studies ($n > 1{,}100$ clinicians) to assess the utility of the zero-shot assistant in human--AI collaborative workflows, focusing on differential diagnoses of general dermatology conditions in primary care and multimodal skin cancer management in specialist settings.  Third, we applied Sparse Autoencoders (SAEs) \cite{sae1,sae2} to investigate whether clinically meaningful concepts emerge without supervision. This approach enables the discovery of latent clinical concepts and targeted interventions to improve model robustness.  In summary, our study provides empirical evidence that a zero-shot foundation model can effectively augment clinical decision-making across diverse healthcare settings, offering a blueprint for shifting the focus from model retraining to zero-shot collaboration.

\Heading{Results}
\section*{Benchmarking zero-shot capabilities on diverse clinical tasks}
A key feature of vision-language foundation models is the zero-shot capability\cite{chextzero,clip}, which allows them to perform downstream tasks without retraining. By pretraining aligned vision-language representations, these models can be directly applied to downstream tasks such as image classification and cross-modal retrieval (retrieving related images from text queries or vice versa).

\begin{figure*}
\centering
\vspace{-12mm}
\includegraphics[width=0.95\textwidth]
{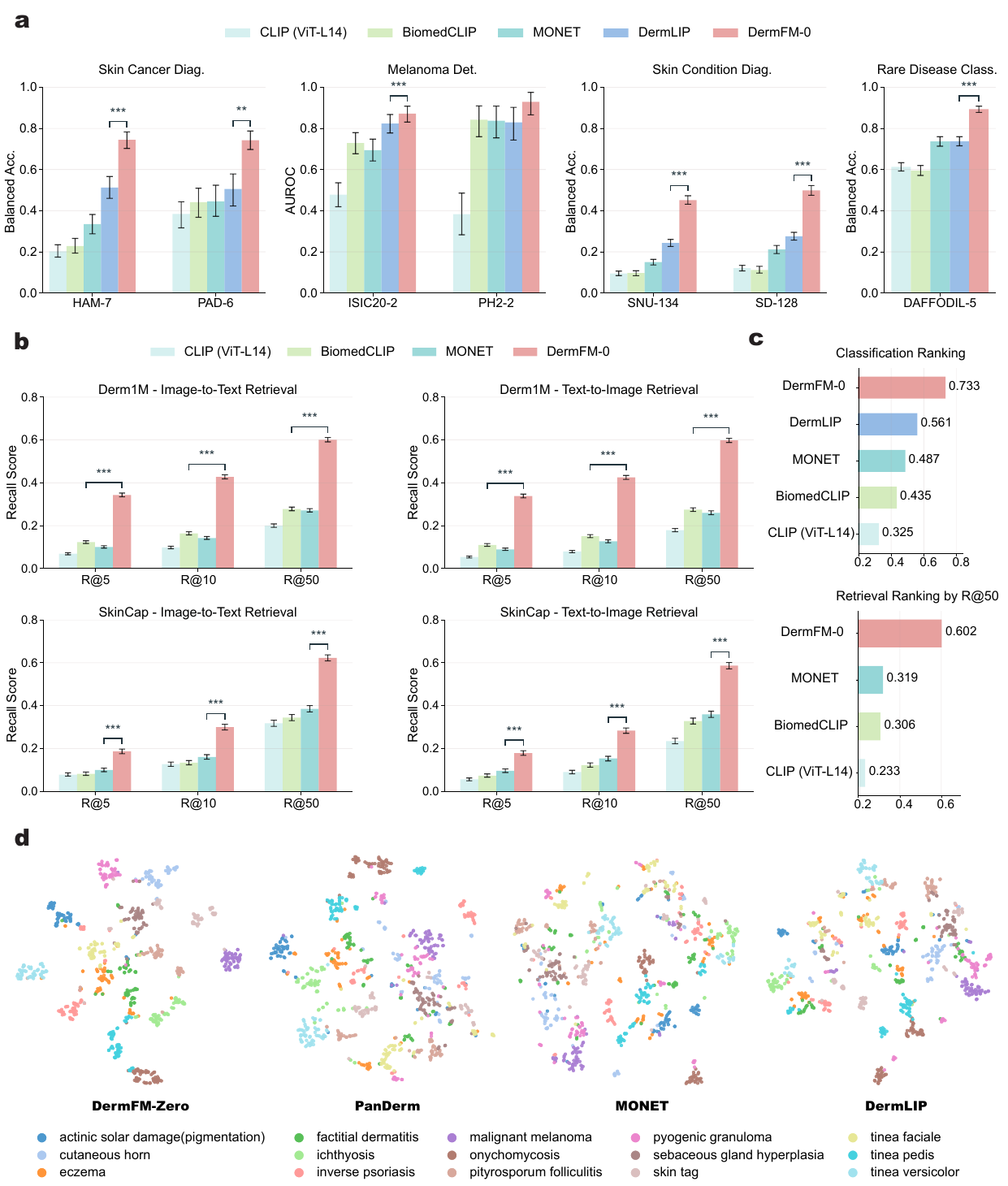}
\caption{\small \textbf{Zero-shot benchmark evaluation.} 
\textbf{a,} Zero-shot image classification performance of DermFM-Zero and other vision-language foundation models across diverse modalities, tasks, and datasets. Metrics: AUROC for binary (c=2) and balanced accuracy for multi-class (c>2) datasets. 
\textbf{b,} Zero-shot image-to-text and text-to-image retrieval performance on the Derm1M and SkinCap datasets, measured using Recall@K (K={5, 10, 50}). 
\textbf{c,} Summary ranking of models by average performance. Top: average zero-shot classification (from \textbf{a}). Bottom: average cross-modal retrieval (based on R@50, from \textbf{b}). 
\textbf{d,} T-SNE visualisation of feature embeddings from DermFM-Zero and other models for the top 20 classes of the SD-128 dataset. 
In \textbf{a, b}, bar centres represent the mean value and error bars show 95\% CIs (computed via non-parametric bootstrapping, 1,000 replicates). Pairwise statistical significance was determined by a two-sided t-test (*\textit{P} < 0.05, **\textit{P} < 0.01, ***\textit{P} < 0.001).
}
\label{fig2}
\end{figure*}

We first evaluated DermFM-Zero's zero-shot performance across four major dermatology tasks on seven public datasets: multi-class skin cancer diagnosis (HAM-10000\cite{ham10000}, PAD-UFES-20\cite{pad}), melanoma detection (ISIC2020\cite{isic}, PH2\cite{ph2}), extensive skin condition diagnosis (SNU-134 \cite{snu134}, SD-128 \cite{sd198}), and rare disease classification (DAFFODIL-5 \cite{Daffodil}). DermFM-Zero outperformed state-of-the-art vision-language foundation models across all benchmarks, with DermLIP\cite{derm1m} generally ranking second (\textbf{Fig~\ref{fig2}a} and \textbf{Extended Data Table~\ref{tab:zero-shot})}. Specifically, for multi-class skin cancer diagnosis using dermoscopy and clinical photographs, DermFM-Zero achieved balanced accuracy of 0.744 (95\% CI 0.702-0.783) and 0.743 (95\% CI 0.697-0.787), surpassing the next-best model by 23.7\% ($P < 0.001$) and 23.2\% ($P < 0.01$), respectively.

We further validated DermFM-Zero on the challenging tasks of diagnosing diverse skin conditions and rare diseases. On the classification of 134 and 128 skin conditions from clinical photographs, DermFM-Zero achieved balanced accuracy of 0.452 (95\% CI 0.431-0.472) and 0.498 (95\% CI 0.475-0.522), exceeding the next-best model (DermLIP) by 22.3\% and 20.9\% (all $P < 0.001$). For rare disease classification, including Stevens-Johnson syndrome and toxic epidermal necrolysis (combined incidence approximately 2-7 cases per million annually), DermFM-Zero achieved a balanced accuracy of 0.893 (95\% CI 0.878-0.908), outperforming the next-best model by 15.6\% ($P < 0.001$). This performance is particularly notable given the scarcity of training data for these life-threatening conditions. On average, DermFM-Zero exceeded the second-best model, DermLIP, by a significant margin (73.3\% vs. 56.1\%) in zero-shot image classification (\textbf{Fig.~\ref{fig2}c}).

To investigate the quality of the pretrained representation, we visualized the feature embeddings produced by the vision encoders of DermFM-Zero, alongside top-performing vision-language models (MONET and DermLIP) and a vision-centric foundation model (PanDerm \cite{panderm}) using t-SNE (\textbf{Fig.~\ref{fig2}d}). Focusing on the top 20 common classes in the SD-128 dataset, we observed that DermFM-Zero generated significantly more distinct and compact clusters compared to other foundation models. This structural separation suggests that DermFM-Zero acquires discriminative, clinically aligned representations, directly supporting its zero-shot capabilities.

To assess semantic alignment between visual features and medical terminology, we evaluated zero-shot cross-modal retrieval on the Derm1M~\cite{derm1m} validation set and SkinCap~\cite{skincap} (\textbf{Fig.~\ref{fig2}b}). On Derm1M, DermFM-Zero achieved R@50 scores of 0.601 (95\% CI 0.591--0.611) for image-to-text and 0.598 (95\% CI 0.588--0.607) for text-to-image retrieval, significantly outperforming the next-best model BiomedCLIP~\cite{biomedclip} by 32.3\% on both tasks ($P < 0.001$). Results were consistent on SkinCap, where DermFM-Zero (I2T: 0.623, 95\% CI 0.608--0.637; T2I: 0.586, 95\% CI 0.571--0.601) surpassed MONET~\cite{monet} by 23.8\% and 22.6\%, respectively ($P < 0.001$). Overall, DermFM-Zero achieved an average R@50 of 60.2\% compared to 31.9\% for MONET (\textbf{Fig.~\ref{fig2}c}). Further results, including qualitative examples for retrieval and visual question answering tasks are presented in \textbf{Extended Data Table~\ref{tab:text_to_image_derm1m_validation}-\ref{tab:image_to_text_skincap}; Extended Data Fig.\ref{supp_retrieval}-\textbf{\ref{supp_vqa}}}, respectively.

\section*{Adaptability to low-resource and multimodal clinical tasks} 

To demonstrate the feature quality of DermFM-Zero, we assessed its adaptability in data-constrained and complex multimodal settings across a wide range of diagnostic and prognostic tasks (\textbf{Extended Data Fig. \ref{supp_lp}--\ref{supp_multimodal}}; \textbf{comprehensive results in Extended Data Tables \ref{tab:label_efficiency_ham}--\ref{tab:zs_prompt}}). First, we evaluated label efficiency via linear probing to simulate low-resource environments. DermFM-Zero consistently outperformed domain-specific baselines (\textbf{Extended Data Fig. \ref{supp_lp}a}), achieving 9.6\% average improvement over the 
second-best model (PanDerm) with only 10\% of labeled data (\textbf{Extended Data Fig. \ref{supp_lp}b}). Despite being 23$\times$ smaller (304M vs. 7B parameters), DermFM-Zero outperformed the general-domain DINOv3 \cite{dinov3} model across nearly all training fractions (\textbf{Extended Data Fig. \ref{supp_lp}c-d})

Second, we evaluated multimodal fine-tuning by integrating dermoscopic images, clinical photographs, and text across diagnosis and prognosis tasks (\textbf{Methods; Extended Data Fig.~\ref{supp_method2}} and \textbf{\ref{supp_multimodal}a--c}). On skin cancer diagnosis (Derm7pt \cite{derm7pt}, PAD \cite{pad}) and skin condition classification (SCIN \cite{scin}), DermFM-Zero consistently outperformed baselines (\textbf{Extended Data Fig. \ref{supp_multimodal}d--f, left}; all ($P < 0.05$)). Modality ablation confirmed complementary information integrating clinical photography, dermoscopy, 
and text yielded significant gains over single-modality inputs (e.g., Macro F1 +8.1\% on Derm7pt; $P < 0.001$) (\textbf{Extended Data Fig. \ref{supp_multimodal}d--f, right}). For melanoma metastasis prediction from dermoscopic images (n=302 patients), multimodal integration showed superiority in both binary and multi-class settings (\textbf{Extended Data Fig. \ref{supp_multimodal}g}). In survival analysis, the model stratified patients 
by recurrence risk (Log-rank $P < 0.001$) and served as a significant independent predictor (Hazard Ratio = 4.70; $P < 0.001$) beyond standard clinical variables (\textbf{Extended Data Fig. \ref{supp_multimodal}h-i}), with robust 3-, 5-, and 7-year predictive power (AUCs: 0.930, 0.945, 0.912; \textbf{Extended Data Fig. \ref{supp_multimodal}j}).

\section*{Reader studies}

We conducted three reader studies to validate DermFM-Zero's zero-shot effectiveness in augmenting clinical decision-making across care settings, from primary care through specialist benchmarking to collaborative specialist care.

\subsection{Reader study 1: Human-AI collaboration for general dermatology conditions in primary care.}

\begin{figure*}[h]
\centering
\vspace{-12mm}
\includegraphics[width=0.9\textwidth]
{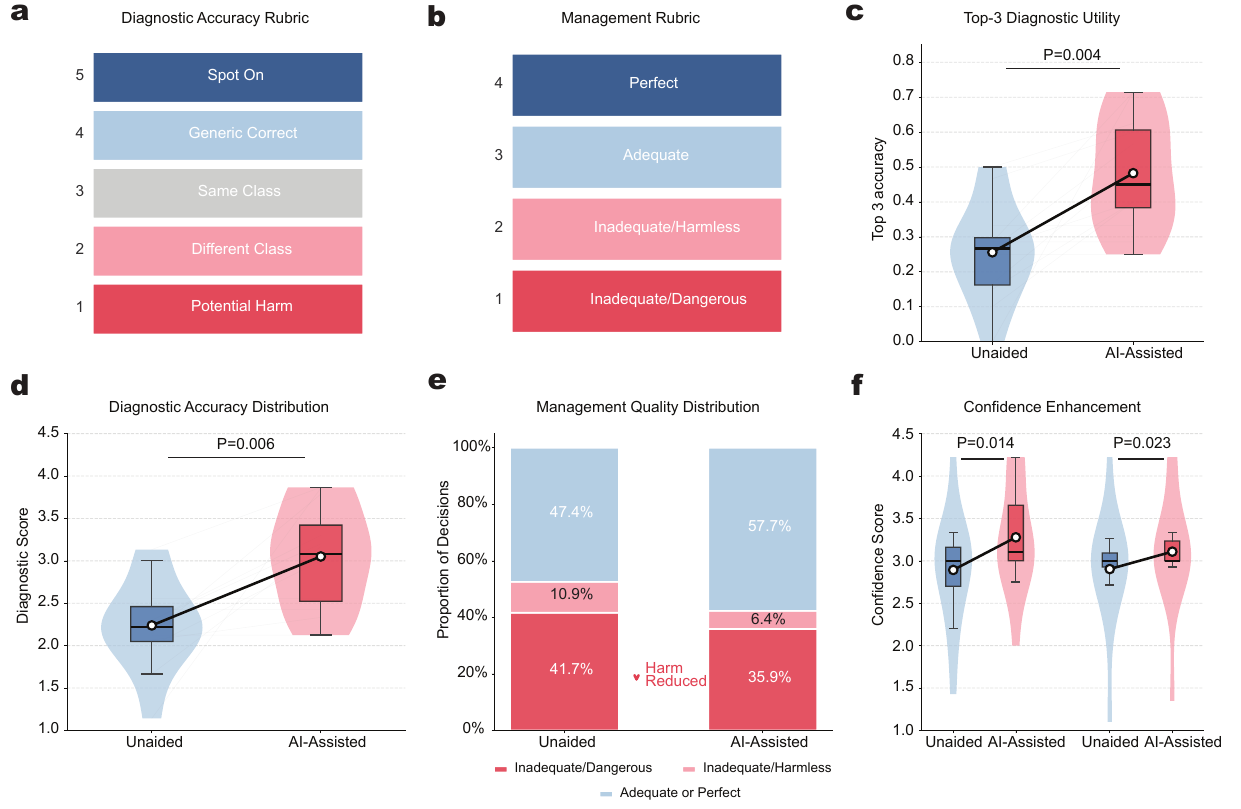}
\caption{\textbf{Reader study 1: Impact of DermFM-Zero zero-shot assistance on diagnostic accuracy and safety in primary care.}
\textbf{a,b,} Evaluation rubrics for diagnostic accuracy (\textbf{a}) and management quality (\textbf{b}), categorizing decisions from potential harm (score 1) to optimal care (score 4 or 5).
\textbf{c,} Top-3 diagnostic utility, defined as the presence of the correct diagnosis within the top three differentials.
\textbf{d,} Overall diagnostic accuracy scores comparing unaided versus AI-assisted performance.
\textbf{e,} Global shift in management decision quality. Stacked bars show the proportion of decisions classified as Dangerous (red), Harmless (grey), and Adequate/Perfect (teal), illustrating a structural shift from potential harm to clinical competence.
\textbf{f,} Reduction in harm rate (proportion of Inadequate/Dangerous decisions) per reader.
Comparisons unaided vs. with DermFM-Zero support ($n=30$ primary care physicians) evaluating complex cases across 98 skin conditions using free-text inputs. Thick black lines connect mean values; thin grey lines represent individual readers. \textit{P}-values from one-sided Wilcoxon signed-rank test. $*P < 0.05$, $**P < 0.01$.}

\label{fig_rs1}
\end{figure*}

We first assessed DermFM-Zero's impact on clinical decision-making in primary care, where diagnostic challenges are most acute. 30 general practitioners (GPs) evaluated complex clinical photo cases spanning 98 skin conditions in a simulated real-world setting (\textbf{Extended Data Fig.~\ref{fig:supp_r1_platform}} and \textbf{Methods}). For each case, GPs provided free-text differential diagnoses, management options, and confidence ratings---both independently and with DermFM-Zero's zero-shot assistance. Outcomes were assessed by dermatologists using expert-defined rubrics for diagnostic accuracy (scores 1--5; \textbf{Fig.~\ref{fig_rs1}a}) and management safety (scores 1--4; \textbf{Fig.~\ref{fig_rs1}b}). AI assistance produced substantial diagnostic improvements, with top-3 accuracy increasing from 0.266 (95\% CI 0.166--0.346) to 0.482 (95\% CI 0.380--0.585; $P = 0.004$; \textbf{Fig.~\ref{fig_rs1}c}) and mean diagnostic scores rising from 2.24 (95\% CI 1.90--2.58) to 3.05 (95\% CI 2.67--3.43; $P = 0.006$; \textbf{Fig.~\ref{fig_rs1}d}). Crucially, this diagnostic precision translated into benefits for management efficacy and safety. The proportion of appropriate management decisions increased significantly from 0.504 (95\% CI 0.350--0.657) to 0.592 (95\% CI 0.431--0.754; $P = 0.018$; \textbf{Fig.~\ref{fig_rs1}e}). Moreover, AI support demonstrated a distinct risk-mitigation effect, reducing potentially harmful management decisions from 0.400 (95\% CI 0.242--0.558) to 0.341 (95\% CI 0.191--0.491; $P = 0.048$; \textbf{Fig.~\ref{fig_rs1}f}). AI assistance also significantly boosted clinician confidence in both diagnosis (2.89 to 3.28; $P = 0.014$) and management (2.90 to 3.11; $P = 0.023$). Comprehensive results are provided in \textbf{Extended Data Table~\ref{tab:rs1_comprehensive_analysis}}.

\subsection{Reader study 2A: Expert-Level benchmarking in specialist care.}

\begin{figure*}[h]
\centering
\vspace{-12mm}
\includegraphics[width=0.95\textwidth]
{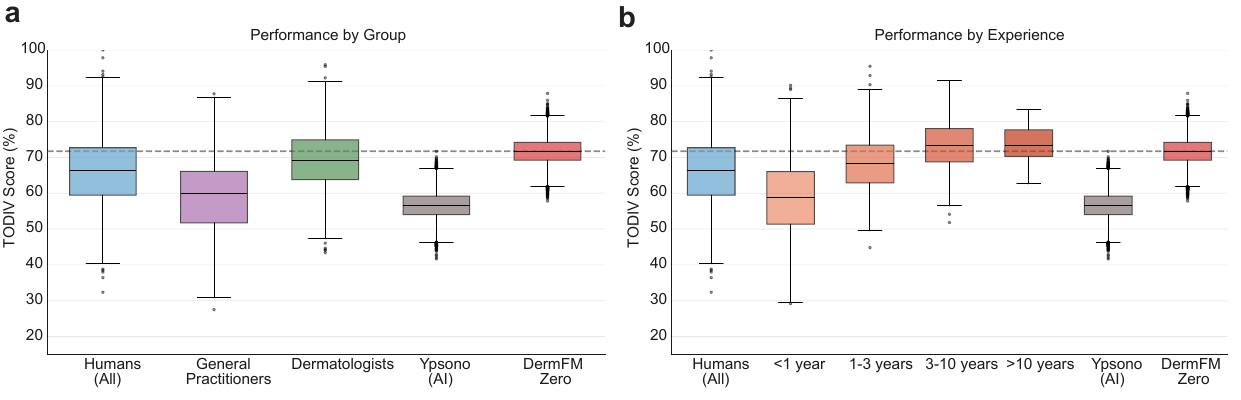}
\caption{\small \textbf{Reader study 2A: Diagnostic performance of DermFM-Zero zero-shot AI compared to clinicians in multi-modal skin cancer assessment.}
\textbf{a,} Performance comparison across reader groups: all human readers (n=1,090), general practitioners (n=284), and dermatologists (n=762), compared against Ypsono AI (a fintuned ResNet50) \cite{humanai} baseline and DermFM-Zero zero-shot model. TODIV scores measure diagnostic accuracy. 
\textbf{b,} Performance stratified by clinical experience levels: <1 year (n=454), 1-3 years (n=305), 3-10 years (n=260), and >10 years (n=71) of dermatological experience, compared with AI systems. Dashed horizontal line is the mean performance of DermFM-Zero (mean=71.74\%). In all boxplots, horizontal lines represent medians, boxes show interquartile ranges (25th to 75th percentiles), whiskers extend to 1.5 times the interquartile range, and dots represent outliers.}

\label{fig_4}
\end{figure*}

We next benchmarked DermFM-Zero's zero-shot performance against specialist-level expertise for skin cancer classification. Compared to an international cohort of 1,090 clinicians (284 general practitioners and 762 dermatologists) evaluating paired dermoscopic and clinical images, DermFM-Zero achieved a diagnostic accuracy of 0.717 (95\% CI 0.717--0.718), significantly exceeding the collective clinician average of 0.663 (95\% CI 0.657--0.669) by 5.4\% ($P < 0.001$; \textbf{Fig.~\ref{fig_4}a} and \textbf{Extended Data Table~\ref{tab:rs2a_performance_by_group}}). Stratification by specialty revealed that DermFM-Zero significantly outperformed general practitioners (0.596; 95\% CI 0.583--0.609) and board-certified dermatologists (0.694; 95\% CI 0.688--0.700) by 12.1\% and 2.3\% ( All $P < 0.001$), respectively. In contrast, a legacy supervised baseline model (a ResNet50 model finetuned for classification of 9 classes with approximately 39000 images, served through Dermonaut's Ypsono tool \cite{humanai}) underperformed the clinician average by 9.6\%. Analysis by experience level demonstrated that DermFM-Zero significantly surpassed clinicians with fewer than 3 years of experience ($P < 0.001$; \textbf{Fig.~\ref{fig_4}b}), while underperforming clinicians with 3--10 years of experience by 1.4\% ($P < 0.001$) and those with $>10$ years by 1.9\% ($P < 0.001$).

\subsection{Reader study 2B: Multimodal human--AI collaboration in specialist care.}

\begin{figure*}[h]
\centering
\vspace{-12mm}
\includegraphics[width=0.95\textwidth]
{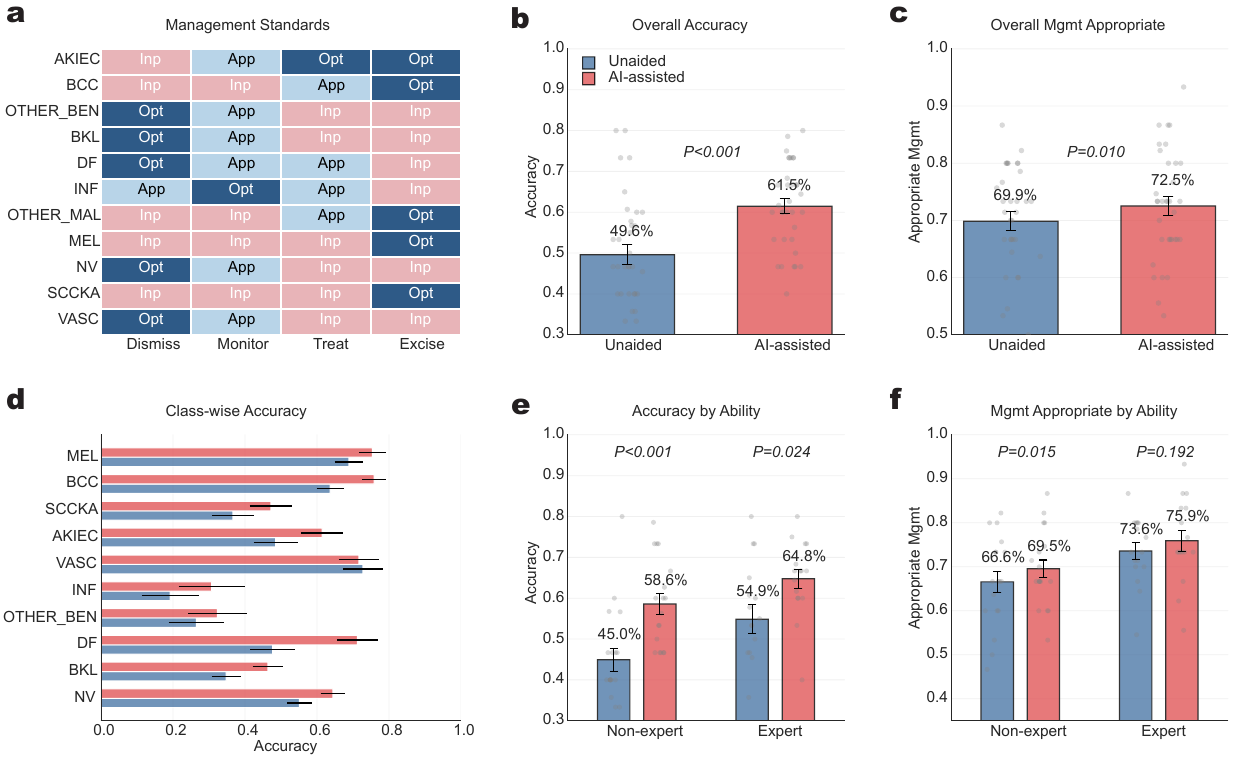}
\caption{\textbf{Reader study 2B: Impact of DermFM-Zero zero-shot assistance on multimodal skin lesion diagnosis and management}
\textbf{a,} Management decision framework showing optimal management strategies (Inp, inappropriate; App, appropriate; Opt, optimal) for 11 skin lesion classes across four decision categories (Dismiss, Monitor, Treat, Excise). AKIEC, actinic keratosis/intraepithelial carcinoma; BCC, basal cell carcinoma; OTHER\_BEN, other benign lesions; BKL, benign keratosis-like lesions; DF, dermatofibroma; INF, inflammatory conditions; OTHER\_MAL, other malignant lesions; MEL, melanoma; NV, melanocytic nevus; SCCKA, squamous cell carcinoma/keratoacanthoma; VASC, vascular lesions.
Comparisons without vs. with DermFM-Zero support (n=34 readers) for: \textbf{b,} overall diagnostic accuracy, \textbf{c,} overall management appropriateness, and \textbf{d,} class-wise diagnostic accuracy across 11 categories. \textbf{e-f,} Accuracy and management appropriateness stratified by non-expert (n=18) and expert (n=16) readers. \textit{P}-values from two-sided paired t-test. Error bars, 95\% CIs; bar centers, means; gray dots, individual readers.}

\label{fig_5}
\end{figure*}

Building on these benchmarking results, we evaluated DermFM-Zero's clinical utility in a multimodal collaborative specialist setting. 34 clinicians assessed complex cases comprising paired dermoscopic and clinical images across 11 lesion classes suspected of malignancy (\textbf{Extended Data Fig.~\ref{fig:specialist_r2b_platform}} and \textbf{Methods}). Management decisions were scored using a framework mapping diagnoses to 4 clinical actions graded from inappropriate to optimal (\textbf{Fig.~\ref{fig_5}a}). AI assistance significantly enhanced overall diagnostic accuracy from 0.50 (95\% CI 0.45--0.54) to 0.61 (95\% CI 0.58--0.65; $P < 0.001$; \textbf{Fig.~\ref{fig_5}b}). Critically, these diagnostic improvements were accompanied by an increase in overall management appropriateness, rising from 0.70 (95\% CI 0.67--0.73) to 0.73 (95\% CI 0.69--0.76; $P = 0.010$; \textbf{Fig.~\ref{fig_5}c}). Diagnostic gains spanned 10 of 11 lesion classes, with particularly strong effects for basal cell carcinoma and dermatofibroma ($P < 0.05$; \textbf{Fig.~\ref{fig_5}d}). 

Stratification by competency revealed a profound ``skill-leveling'' effect: non-experts derived the greatest benefit, improving diagnostic accuracy from 0.45 (95\% CI 0.39--0.51) to 0.59 (95\% CI 0.54--0.64; $P < 0.001$; \textbf{Fig.~\ref{fig_5}e}) and surpassing unassisted expert performance (0.55, 95\% CI 0.48--0.62). While experts' accuracy also improved from 0.55 (95\% CI 0.48--0.62) to 0.65 (95\% CI 0.60--0.69) ($P = 0.024$), DermFM-Zero effectively bridged the expertise gap in patient care: AI-assisted non-experts achieved management appropriateness rates (0.70, 95\% CI 0.65--0.74) comparable to those of unassisted experts (0.74, 95\% CI 0.70--0.77), reflecting a significant performance improvement within the non-expert group ($P = 0.015$; \textbf{Fig.~\ref{fig_5}f}). Detailed results are provided in \textbf{Extended Data Tables~\ref{tab:rs2a_performance_by_group}--\ref{tab:rs2b_management_by_experience}}.

\section*{DermFM-Zero enables automated discovery of clinically meaningful concepts}

\begin{figure*}[h]
\centering
\vspace{-12mm}
\includegraphics[width=0.95\textwidth]
{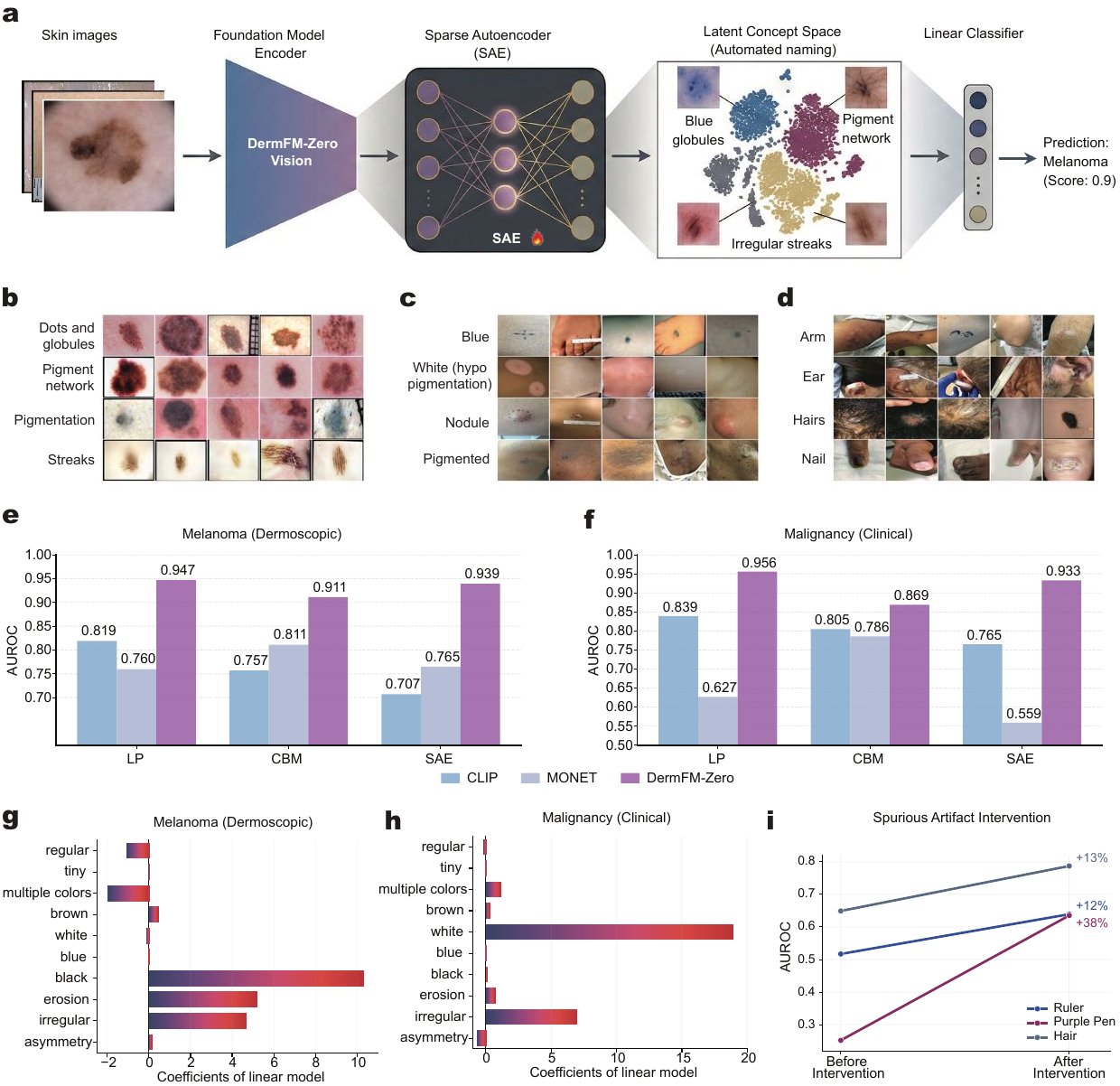}
\caption{\textbf{Automated concept discovery and human intervention.} 
\textbf{a,} Schematic of automated concept discovery. A sparse autoencoder (SAE) disentangles zero-shot DermFM-Zero vision features into a sparse latent space where individual neurons represent distinct concepts (e.g., 'Blue globules', 'Pigment network') automatically named via the text encoder and used to train a linear classifier.
\textbf{b--d,} Automatically discovered concepts for dermoscopic (Derm7Pt, melanoma/nevus, n=827; \textbf{b}) and clinical photo datasets (F17K and DDI, malignancy/benign, n=3,056; \textbf{c, d}).
\textbf{e, f,} Performance comparison of melanoma (\textbf{e}) and malignancy (\textbf{f}) prediction using three approaches: Linear Probing (LP) on zero-shot DermFM-Zero features; Concept Bottleneck Models (CBMs) + LP using MONET-annotated concepts from expert-curated vocabulary; and SAE + LP using sparse concepts discovered and named by SAE. Unlike CBMs constrained by predefined concepts, SAE automatically uncovers concepts inherently meaningful for diagnosis..
\textbf{g, h,} Linear model coefficients in DermFM-Zero SAE-CBM for melanoma (\textbf{g}) and malignancy (\textbf{h}) prediction. Positive coefficients indicate target class associations; negative coefficients indicate benign associations.
\textbf{i,} Human intervention on artifact-confounded ISIC subsets: 'Ruler' (n=500), 'Purple Pen' (n=246), 'Hair' (n=649). Performance (AUROC) before and after manually suppressing top-5 artifact-associated neurons.}
\label{fig_6}
\end{figure*}

To probe the internal structure of DermFM-Zero's learned representations, we investigated whether the model spontaneously encodes interpretable medical concepts. We adopted a ``discover-then-name'' setting~\cite{sae1} using Sparse Autoencoders (SAEs) to disentangle DermFM-Zero's pretrained vision features into a sparse latent space. Individual latent neurons, each corresponding to a distinct visual pattern, were automatically named using the text encoder and used to train linear classifiers (\textbf{Fig.~\ref{fig_6}a}). 

We applied this approach to dermoscopic (Derm7pt, melanoma/nevus, $n=827$) and clinical (F17K and DDI, malignancy/benign, $n=3{,}056$) image datasets. The discovered concepts showed strong alignment with established medical semantics despite the absence of explicit concept supervision (\textbf{Fig.~\ref{fig_6}b--d}). In dermoscopic images, individual neurons consistently activated on diagnostically relevant structures, such as ``blue globules'' and ``pigment networks''. DermFM-Zero demonstrated superior concept retrieval performance across dermoscopic features, such as streaks (Precision@50: 0.64 vs. CLIP 0.22, MONET 0.58) and pigment network (0.94 vs. CLIP 0.52, MONET 0.80; \textbf{Extended Data Table~\ref{tab:concept_retrieval}}). In clinical photographs, the SAE extracted both lesion-specific visual features (``color variation'', ``morphology'') and anatomical context (``arm'', ``ear'', ``face''), suggesting that DermFM-Zero encodes concept and contextual information in a disentangled manner.

To assess the diagnostic utility of the discovered concepts, we constructed Concept Bottleneck Models (CBMs)~\cite{cbm}, in which images are first mapped to interpretable clinical concepts and then used to predict clinical outcomes via a linear classifier. We compared our unsupervised SAE-CBM with CBMs built on predefined, expert-curated dermatology concept vocabularies (MONET settings)~\cite{monet} (\textbf{Fig.~\ref{fig_6}e,f}). For melanoma prediction, the SAE-CBM based on DermFM-Zero achieved an AUROC of 0.939, outperforming the predefined-concept MONET baseline (AUROC 0.765) and approaching a black-box linear probe trained directly on DermFM-Zero features (AUROC 0.947). Across experiments, standard CBMs constrained to predefined concept vocabularies consistently underperformed the unsupervised SAE-based approach on DermFM-Zero. This behavior was specific to DermFM-Zero. When applied to CLIP and MONET, SAE-CBM underperformed standard CBMs (CLIP: 0.707 vs. 0.757; MONET: 0.765 vs. 0.811 AUROC). These results indicate that automated concept discovery is effective only when the underlying foundation model learns sufficiently structured and clinically aligned representations. Similar trends were observed for malignancy prediction (\textbf{Fig.~\ref{fig_6}f}).

Analysis of the linear classifier weights revealed clinically coherent patterns in the discovered concepts. For melanoma prediction, positive weights were assigned to established diagnostic markers such as ``erosion'', ``black'', and ``irregularity'', while benign-associated concepts including ``regular'' and ``tiny'' were assigned negative weights (\textbf{Fig.~\ref{fig_6}g}). Similar clinically meaningful associations were observed for malignancy prediction (\textbf{Fig.~\ref{fig_6}h}).

Beyond interpretability, the discovered concepts enabled targeted model intervention. Using ISIC subsets containing non-diagnostic visual artifacts, including rulers ($n=500$), purple pen markings ($n=246$), and hair occlusion ($n=649$), we identified the top five neurons most strongly activated by each artifact type and suppressed their activations at inference time. This intervention, which is not feasible in black-box classifiers, led to substantial improvements in diagnostic robustness, with AUROC increases ranging from 12\% to 38\% across artifact categories (\textbf{Fig.~\ref{fig_6}i}). These results show that the interpretable representations learned by DermFM-Zero can be used to mitigate visual artifact biases without retraining.

\Heading{Discussion}

\noindent The clinical translation of foundation models (FMs) remains limited despite demonstrated performance in benchmark evaluations and human-AI collaboration studies~\cite{panderm,eyecarefm,contch,biomedgpt}. Current medical FMs typically require task-specific fine-tuning, a requirement incompatible with dynamic clinical settings where diverse, heterogeneous cases arise without the opportunity for retraining. While several vision-language FMs in pathology~\cite{contch,musk,titan} and ophthalmology~\cite{retizero} have attempted to address this via zero-shot inference, they largely remain proof-of-concept and have not been evaluated in rigorous human-AI collaboration studies. Moreover, most existing FMs lack intrinsic interpretability, operating as ``black boxes'' that impede clinical trust and prevent the identification of failure modes. These limitations create a fundamental gap between technical advances and practical clinical deployment.

In this study, we developed DermFM-Zero, a vision-language foundation model designed to function as a transparent, zero-shot clinical assistant for dermatology. To address the data bottleneck limiting medical vision-language FMs, we employed a strategic two-stage training curriculum. Initially, we pretrained the vision encoder on over 3 million unlabeled multimodal dermatological images using masked latent modelling~\cite{panderm,caev2} to capture fine-grained morphology. We then refined these representations through bootstrapped image-text contrastive learning on 1 million curated pairs derived from educational resources and multi-country clinical data, covering over 400 skin conditions. To further enhance semantic understanding, we integrated PubMedBERT~\cite{pubmedbert} with an extended token window to process detailed clinical descriptions. To address the challenge of interpretability, we introduced a framework using Sparse Autoencoders~\cite{sae1,sae2} to automatically discover clinically relevant concepts within learned representations, moving beyond the predefined vocabularies used in prior approaches~\cite{monet}. We validated DermFM-Zero through a comprehensive evaluation framework spanning diverse benchmarks for zero-shot generalization, multinational reader studies for clinical utility, and interpretability analyses demonstrating transparent diagnosis and actionable bias mitigation.

We first benchmarked DermFM-Zero against state-of-the-art FMs to validate its zero-shot generalization across diverse clinical tasks, ranging from common skin cancer screening to rare disease classification and cross-modal retrieval. Unlike previous image-centric models~\cite{panderm,eyefm} or vision-language models constrained by limited scale~\cite{monet,derm1m}, DermFM-Zero demonstrated significantly superior performance across these scenarios without task-specific retraining. This capability stems directly from our alignment strategy anchored in high-quality educational and multi-country clinical data, as evidenced by robust cross-modal retrieval and the formation of compact, discriminative feature clusters in the latent space. Crucially, the quality of these representations translated to superior transfer learning efficiency. Despite being 23$\times$ smaller than the general-domain DINOv3-7B model, DermFM-Zero outperformed it across training data fractions. This finding implies that domain-specific pretraining produces more clinically relevant representations than scaling model parameters in natural image domains, which is critical for high-stakes applications. Beyond diagnostic breadth, the model demonstrated the depth required for complex prognostication. By integrating dermoscopy with patient metadata, DermFM-Zero achieved accurate melanoma metastasis prediction and prognosis, significantly outperforming conventional unimodal methods~\cite{panderm} and standard clinical variables.

Moving beyond standalone metrics, we assessed the clinical utility of DermFM-Zero through zero-shot reader studies designed to evaluate human-AI collaboration. In primary care settings, where diagnostic challenges are exacerbated by limited specialized training, general practitioners evaluated clinical photo cases sourced from diverse populations across multiple countries. DermFM-Zero functioned as a critical mechanism for risk mitigation. AI support improved diagnostic accuracy and shifted clinical behavior toward safer management decisions by reducing potentially harmful errors. This underscores the potential of zero-shot FMs to support general practitioners in community health settings where specialist access is limited. In specialist care requiring paired dermoscopic and clinical images for accurate management, DermFM-Zero demonstrated expert-level proficiency in standalone assessments. The model significantly outperformed an international cohort of 1,090 clinicians. Importantly, collaborative evaluations revealed a distinctive skill-leveling effect. AI assistance enabled non-experts to surpass the diagnostic performance of unassisted experts (58.6\% vs 54.9\%) and narrowed the disparity in management appropriateness. This effect, previously observed primarily with fine-tuned models \cite{humanai,panderm}, was achieved here in a zero-shot fashion without retraining. Furthermore, consistent with prior studies \cite{humanai,panderm}, human-AI collaboration matched but did not consistently exceed standalone AI performance. This reflects appropriate clinical judgment where clinicians selectively use AI outputs as a validation tool rather than a directive, maintaining their role as the final decision-maker. Collectively, these findings validate that foundation models can effectively augment clinical decision-making and democratize expert-level care without task-specific retraining.

An important capability of DermFM-Zero is its emergent interpretability through automated concept discovery. Previous transparent foundation models, such as MONET~\cite{monet}, rely on mapping internal representations to predefined concept vocabularies. However, forcing alignment with human-defined concepts can introduce a semantic disconnect from the model’s learned representation space, potentially compromising both faithfulness and diagnostic accuracy. In contrast, DermFM-Zero employs Sparse Autoencoders (SAEs) to identify intrinsic concepts in a ``discover-then-name'' manner~\cite{sae1}. This approach aligns interpretability with the model’s intrinsic representations, enabling explanations that more faithfully reflect the rationale underlying predictions while largely preserving diagnostic performance. Importantly, our results demonstrate that the effectiveness of automated concept discovery is not universal, but critically depends on the quality of the underlying foundation model. SAE-based classifiers built on DermFM-Zero outperformed predefined concept vocabularies and closely matched black-box baselines, whereas the same approach resulted in inferior performance when applied to general-domain models (CLIP) or existing medical foundation models (MONET). These contrasting outcomes indicate that unsupervised concept discovery requires representation spaces that are both well-structured and semantically aligned with clinical knowledge, a property enabled by DermFM-Zero’s domain-specific pretraining. Beyond interpretability, we translated this emergent capability into actionable mechanisms for model auditing and correction. Deep learning models are known to exploit spurious correlations with non-diagnostic visual artifacts~\cite{trustderm,dc}, such as rulers or skin markings, which undermine reliability. By leveraging the discovered concepts to identify and selectively suppress artifact-associated neurons at inference time, we substantially mitigated these biases and improved diagnostic performance by up to 38\% on artifact-confounded datasets. This editable, concept-level intervention provides a transparent alternative to black-box retraining, supporting the development of more trustworthy and controllable medical AI systems.

Our study has limitations that should inform the interpretation of these results. First, although our reader studies were designed to closely mimic the decision-making complexity of clinical practice, they remain retrospective simulations within controlled environments. As such, they cannot fully replicate the dynamic constraints and workflow interruptions inherent to real-world patient care. Future validation should prioritize prospective reader studies~\cite{prov-ai,eyecarefm} or ``silent'' deployments~\cite{real_path} to provide evidence of clinical utility in live practice. Second, while the pretraining data of DermFM-Zero covers over 400 skin conditions, which is a scale comparable to or exceeding prior studies~\cite{panderm,liu2020deep}, this represents only a fraction of the full dermatological spectrum that encompasses over 2,000 entities. To address this long tail, future work should leverage the model's unsupervised representation capabilities to construct a dynamic, FM-driven ontology. By identifying and organizing new visual patterns from vast unlabeled data, we can expand the medical lexicon used by agentic systems. This capability allows agents to bridge the gap between visual observations and medical literature, enabling sophisticated reasoning and retrieval even for conditions they have not been explicitly supervised to recognize~\cite{deeprare}. Third, although we analysed standalone performance across multinational cohorts, we did not specifically investigate skin-tone-based fairness within the human-AI collaboration dynamic. While DermFM-Zero was pretrained on a geographically diverse dataset to mitigate algorithmic bias, we acknowledge that equitable standalone performance may not guarantee unbiased human-AI interaction \cite{skintone-humanai}. Therefore, further investigation is required to understand how diverse clinician groups interact with model outputs to ensure equitable patient benefits. Finally, we have not yet integrated the transparent capabilities of DermFM-Zero directly into the human-AI collaboration workflow. While previous studies~\cite{humanai3} indicate that concept-based explanations can improve clinician confidence, the specific impact of DermFM-Zero's automated concept discovery and intervention mechanisms on clinical decision-making remains unassessed. Future work requires the comprehensive design of new collaborative paradigms to rigorously evaluate how these interpretable features influence decision-making and trust.

In conclusion, we developed and validated DermFM-Zero, a vision-language foundation model that demonstrates zero-shot clinical capabilities and automated interpretability for dermatology. Through rigorous evaluation across technical benchmarks, multinational reader studies, and interpretability analysis, we show that domain-specific foundation models can assist clinical decision-making across diverse settings without the need for task-specific retraining. By combining the breadth of zero-shot inference with mechanisms for transparent and actionable insights, DermFM-Zero offers a framework for the next generation of medical AI systems that are data-efficient and more transparent. These capabilities represent a step forward in integrating foundation models into clinical workflows, facilitating the transition from technical validation to real-world human-AI collaboration.

\end{spacing}
\clearpage
\begin{spacing}{1.35}
\Heading{Methods}

\heading{Ethics statement}\\
The Molemap dataset was utilised under a research agreement with the Monash eResearch Centre, with approval granted by the Monash University Human Research Ethics Committee (MUHREC). For the ACEMID\_path study, data accrued for the registered trial ACTRN12619001706167 was shared under approval from the Alfred Hospital Ethics Committee (746/23), operating under the primary protocols of the Metro South Human Research Committee (HREC/2019/QMS/57206) and the University of Queensland Human Research Ethics Committee (2019003077). The MYM study received approval from the Metro South Health Human Research Ethics Committee (HREC/16/QPAH/125) on April 21, 2016, with additional approvals from the University of Queensland (2016000554), Queensland University of Technology (1600000515), and QIMR Berghofer (P2271). Similarly, the HOP study was approved by the Metro South Health HREC (HREC/17/QPAH/816) and The University of Queensland HREC (2018000074).
The ComBineMel (Computer biomarkers evaluation of invasive melanoma) study operates in compliance with the National Statement on Ethical Conduct in Human Research (2007) and was approved by the Alfred Hospital Ethics Committee (HREC/98200/Alfred-2023). The SDDI\_Alfred study also received clearance from the Alfred Hospital Ethics Committee (198/19) for the use of sequential dermoscopic imaging, utilizing strictly de-identified retrospective data without active patient involvement. The NSSI dataset, part of the Brisbane Naevus Morphology Study (2009–2014), was approved by the Princess Alexandra Hospital Human Research Ethics Committee and adhered to the Declaration of Helsinki protocols. Finally, the SDDI2 dataset obtained approval from the Ethics Review Board of the Medical University of Vienna.

\heading{Pretraining data curation}\\
To develop the multimodal DermFM-Zero foundation model, we curated two distinct datasets tailored to our strategic two-stage training curriculum: a large-scale unlabeled corpus for self-supervised representation learning (SSL) and a high-quality image-text paired dataset for cross-modal contrastive learning.

\noindent\textbf{Dataset for Stage 1 (Unlabeled Visual Representation Learning):}
The dataset for the initial SSL pretraining phase consisted of over 3 million unlabeled multimodal dermatological images. This represents a comprehensive expansion of the dataset used for PanDerm-v1~\cite{panderm}, aggregated from 15 distinct data sources across four imaging modalities. We incorporated two additional sources: (1) 403,563 image-text pairs from the Derm1M dataset~\cite{derm1m}, and (2) 296,467 images from in-house educational resources, including atlases, textbooks, and clinical guidelines. This comprehensive corpus was designed to maximize visual diversity and ensure robust low-level feature learning across the full spectrum of dermatological presentations. Patient metadata was excluded from the training dataset to maintain focus on visual features.

\noindent\textbf{Dataset for Stage 2 (Vision-Language Alignment):}
For the subsequent vision-language alignment phase, we curated a high-quality subset of 1,014,085 image-text pairs. This selection prioritized samples with rich semantic content to facilitate learning of complex medical concepts. The dataset comprised three complementary sources: (1) a filtered high-quality subset (n=240,246) from Derm1M~\cite{derm1m}, sourced from diverse open-access educational websites (e.g., PubMed and YouTube educational channels) with metadata spanning over 390 skin conditions; (2) an expanded collection of educational resources (n=199,600) from dermatology atlases, textbooks, and clinical guidelines; and (3) private clinical data from the in-house Molemap collection (n=574,239 images from 78,760 patients), acquired from 141 dedicated community clinics across Australia and New Zealand between 2010 and 2019. The Molemap data featured fine-grained diagnostic labels across 65 skin disease classes, including inflammatory conditions, infections, benign proliferations, melanocytic lesions, and eczema.

\heading{Model development}\\
\noindent\textbf{Step 1: Unimodal masked pretraining.}
We utilised a masked latent modelling approach \cite{caev2} to initialize the vision encoder. The architecture comprises a ViT-Large encoder $f$, a lightweight regressor $h$, and a frozen CLIP-Large \cite{clip} teacher $T$. An input image $\mathbf{x}$ is partitioned into patches, which are divided into visible sets $\mathcal{V}$ and masked sets $\mathcal{M}$. The encoder extracts latent representations $\mathbf{z}^v$ from visible patches, while the regressor predicts the representations $\mathbf{z}^m$ of masked patches. The objective is to minimize the cosine distance $D(\cdot, \cdot)$ between the student's outputs and the teacher's target features $\bar{\mathbf{z}}$ for both visible and masked regions:
\begin{equation}
\mathcal{L}_{\text{MIM}} = \underbrace{\sum_{i \in \mathcal{V}} D(\mathbf{z}_i^v, \bar{\mathbf{z}}_i^v)}_{\text{Visible Alignment}} + \underbrace{\sum_{j \in \mathcal{M}} D(\mathbf{z}_j^m, \bar{\mathbf{z}}_j^m)}_{\text{Masked Alignment}}
\end{equation}
\textbf{Masked latent training settings.}
We initialized the encoder with ImageNet-1K weights and trained on 3 million unlabeled multimodal skin images. We used a 50\% masking ratio. Image augmentations included random resized cropping and horizontal flipping. We trained for 500 epochs using the AdamW optimizer with $\beta_1=0.9, \beta_2=0.95$, a weight decay of 0.05, and an initial learning rate of $1.5 \times 10^{-3}$ with a 20-epoch linear warmup. The effective batch size was 1,920 across four 80-GB NVIDIA H100 GPUs.

\heading{Step 2: Multimodal contrastive pretraining.}\\
The second step utilised contrastive learning to align the pretrained visual encoder in stage 1 with medical semantics. We replaced the standard text encoder with PubMedBERT and extended the tokenizer sequence length to 256 to capture detailed clinical descriptions. Given a batch of $N$ image-text pairs, we computed the normalized image embeddings $\mathbf{u}$ and text embeddings $\mathbf{v}$. The model was optimized using the symmetric InfoNCE loss to maximize the similarity of matched pairs:

\begin{equation}
\mathcal{L}_{\text{Con}} = - \frac{1}{N} \sum_{i=1}^{N} \left( \log \frac{\exp(\mathbf{u}_i^\top \mathbf{v}_i / \tau)}{\sum_{j=1}^{N} \exp(\mathbf{u}_i^\top \mathbf{v}_j / \tau)} + \log \frac{\exp(\mathbf{v}_i^\top \mathbf{u}_i / \tau)}{\sum_{j=1}^{N} \exp(\mathbf{v}_i^\top \mathbf{u}_j / \tau)} \right)
\end{equation}

\noindent\textbf{Contrastive training settings.}
We pretrained the model on the curated dataset of 1 million image-text pairs for 30 epochs. The input resolution was $224 \times 224$ pixels. We used the AdamW optimizer with a peak learning rate of $1 \times 10^{-4}$ and a cosine decay schedule. The batch size was set to 4,096 (4× gradient accumulation) to ensure sufficient negative samples for contrastive learning.

\noindent\textbf{Domain-specific design.}
To adapt general foundation model architectures to the nuances of dermatology, we implemented four key domain-specific modifications. First, we employed domain-aware multi-source data optimisation to balance the contribution of diverse imaging modalities and heterogeneous data sources (image-caption vs. image-metadata and labels) during training, preventing model bias toward dominant data types. Second, we utilised bootstrapped contrastive learning to purify the training corpus; a baseline model filtered out noisy image-text pairs with low alignment scores to ensure high semantic fidelity. Third, we prioritized domain-specific unimodal encoder pretraining before multimodal alignment, ensuring the model first mastered fine-grained lesion morphology independent of textual associations. Finally, to address the complexity of medical language, we replaced the standard CLIP text encoder (GPT-2 with BPE tokenizer) with PubMedBERT using a WordPiece tokenizer (30k vocabulary) and extended the maximum text sequence length from 77 to 256 tokens. This adaptation allows DermFM-Zero to process complex medical terminology and comprehensive clinical descriptions that exceed the capacity of general-domain text encoders.


\noindent\textcolor{black}{\heading{Benchmark datasets}}\\
We evaluated DermFM-Zero across multiple publicly available dermatology datasets spanning four evaluation settings. For zero-shot classification, we used seven datasets. Dermoscopic datasets include HAM-7\cite{ham10000} with 1,503 images across seven skin cancer and benign lesion types, PH2-2\cite{ph2} with 200 images for melanoma versus nevus classification, and ISIC20-2\cite{isic2020} with 4,969 images for binary melanoma detection. Clinical photography datasets comprise DAFFODIL-5\cite{Daffodil} offering 1,910 images spanning five rare inflammatory and pigmentary conditions, PAD-6\cite{pad} with 461 images covering six categories of skin cancers and benign lesions, SNU-134\cite{snu134} featuring 2,101 images across 134 common disease classes, and SD-128\cite{sd198} encompassing 1,405 images representing 128 diagnostic categories. For zero-shot cross-modal retrieval, we evaluated on Derm1M validation set\cite{derm1m} containing 9,806 image-text pairs and SkinCap\cite{skincap} with 3,989 annotated pairs spanning 178 diseases. For linear probing, we utilised complete train/validation/test splits stratified by class labels from four datasets: HAM-7 with 10,015 dermoscopic images, ISIC20-2 with 33,126 dermoscopic images, PAD-6 with 2,298 clinical images, and SD-128 with 5,619 clinical images, where test sets matched those in zero-shot evaluation. For visual question answering, we constructed two datasets. Derm7pt-VQA comprises 4,000 question-answer pairs generated from Derm7pt\cite{derm7pt} metadata using handcrafted templates (60/20/20 train/validation/test split, no image overlap), spanning 49 unique answers across seven question types including concept existence, diagnosis, texture, and management. SkinCap-VQA contains 4,665 samples (60/20/20 split) derived from Derm1M\cite{derm1m} instruction set filtered for SkinCap-sourced images, with 188 unique answers after removing multiple-choice options and filtering low-frequency responses. For multimodal fine-tuning, we employed five datasets integrating multiple data modalities. Derm7pt\cite{derm7pt} provides 1,011 cases combining clinical and dermoscopic images across 5 skin cancer categories with 7-point diagnostic criteria. PAD\cite{pad} contains 2,298 cases pairing clinical images with lesion metadata across six diagnostic categories. SCIN-20\cite{scin} includes 2,140 cases combining clinical images with demographic and symptom metadata for 20 common conditions. PUMCH\cite{pumch} encompasses 1,174 patients with paired clinical and dermoscopic images across eight inflammatory disease categories. Combinmel comprises 663 cases integrating dermoscopic images with clinicopathological metadata for binary metastasis prediction. For automated concept discovery, we used three specialized datasets. Clinical malignancy classification employed 3,056 images from F17k\cite{f17k} and DDI\cite{ddi}, while dermoscopic melanoma detection used 827 images from Derm7pt. ISIC-Intervention\cite{isic2016, isic2017, isic2018, isic2019, isic2020} comprised three artifact-biased datasets: hair occlusion with 1,120 images, ruler markings with 1,111 images, and pen annotations with 723 images - for evaluating robustness against spurious correlations. Detailed dataset descriptions are provided in Supplementary Methods.

\noindent\textcolor{black}{\heading{Eval 1: Benchmark evaluation}}

\noindent\textcolor{black}{\hheading{Competing baselines.}} We compared DermFM-Zero against models spanning three training settings based on data domain and specialization. General-domain models include CLIP-L/14\cite{clip} trained on 400 million internet image-text pairs and DINOv3\cite{dinov3} series(7B and L16) pretrained via self-supervised learning on web images. Biomedical-domain models comprise BioMedCLIP\cite{biomedclip}, a vision-language foundation model pretrained on 15 million biomedical research articles across multiple medical specialties. Dermatology-specific models include MONET\cite{monet} pretrained on 105,550 dermatology image-text pairs for medical concept identification, DermLIP\cite{derm1m} trained on 403,563 educational materials, MAKE\cite{make} employing knowledge-enhanced contrastive learning on the Derm1M\cite{derm1m} dataset, and PanDerm\cite{panderm}, a dermatological vision foundation model pretrained via self-supervised learning. For multimodal fine-tuning tasks, we augmented PanDerm with a GPT-2\cite{gpt2} text encoder to enable text processing capabilities.

\noindent\textcolor{black}{\hheading{Evaluation metrics.}} For zero-shot classification and linear probing, we reported weighted F1 score (W\_F1), which accounts for class imbalance by weighting each class's F1 by its support; area under the receiver operating characteristic curve (AUROC), measuring discrimination ability between classes; balanced accuracy (BACC), computed as the average sensitivity across all classes; and sensitivity (SENS), representing the true positive rate. Linear probing additionally included macro F1 score (M\_F1), the unweighted average of per-class F1 scores. For zero-shot cross-modal retrieval, we measured Recall@K for K $\in$ \{1, 5, 10\}, indicating the proportion of relevant items retrieved within the top-K results, and reported mean recall averaged over all K values. For multimodal fine-tuning, we used W\_F1, AUROC, BACC, and M\_F1 as primary metrics. For automated concept discovery, we used AUROC to evaluate binary classification performance on melanoma detection and malignancy prediction tasks.

\noindent\textcolor{black}{\hheading{Zero-shot learning.}} We evaluated DermFM-Zero's zero-shot capabilities on image classification and cross-modal retrieval tasks following the CLIP\cite{clip} framework. In zero-shot classification, each test image was encoded into a visual embedding using the vision encoder. To represent disease classes, we applied multiple prompt templates for each condition (e.g., ``a photo of [disease name]''). The text encoder computed embeddings for all templates of a given class, which were then averaged to create a single representative class embedding. Predictions were determined by computing cosine similarity between the image embedding and each mean class embedding, with the class exhibiting maximum similarity selected as the prediction. For multi-class datasets, similarity scores were normalized via softmax to yield class probabilities. 

\noindent\textcolor{black}{\hheading{Linear probing.}} Linear probing evaluates the quality of learned visual representations by training a linear classifier on top of frozen features. We used only the vision encoder from DermFM-Zero to extract visual embeddings for all training, validation, and test images, with the encoder weights frozen throughout the process. A logistic regression classifier was then trained on these fixed visual features to predict disease labels. We employed the L-BFGS optimization algorithm with a maximum of 1,000 iterations. The L2 regularization strength was set to $MC/100$, where $M$ represents the visual feature dimension and $C$ denotes the number of disease classes. 

\noindent\textcolor{black}{\hheading{Cross-modal retrieval.}} For cross-modal retrieval, we evaluated both image-to-text and text-to-image directions. Given a query image, text candidates were ranked by cosine similarity between their text embeddings and the image embedding. Conversely, given a query text, image candidates were ranked against the text embedding. Performance was measured using Recall@K, which captures whether the correct match appears within the top K retrieved results.


\noindent\textcolor{black}{\hheading{Visual question answering.}} Visual question answering requires generating answers conditioned on both visual content and textual queries. We framed this task as a multi-class classification task, where the model predicts answers from all unique responses in the dataset. For each question-image pair, we extracted visual features using DermFM-Zero's vision encoder and textual features using the text encoder. The visual features were aggregated via mean pooling across spatial dimensions and projected through a linear layer, while textual features were similarly projected through a linear layer. These visual and textual features were then fused using a cross-attention mechanism implemented with a meta-fusion transformer (8 attention heads, 4 layers). The resulting representation and visual feature were concatenated, then fed into an MLP classifier to predict the answer class. Both encoders were fine-tuned end-to-end during training. We used the Adam optimizer with learning rate 1e-5, $\beta_1=0.9$, and $\beta_2=0.999$, training for 50 epochs with batch size 32 and gradient accumulation over 2 steps. Cross-entropy loss was computed on the training set, with validation accuracy monitored to select the best checkpoint for test evaluation.

\noindent\textcolor{black}{\heading{Eval 2: Multimodal finetuning}}

\noindent\textcolor{black}{\hheading{Multimodal finetuning framework.}} Multimodal finetuning evaluates DermFM-Zero's ability to integrate signals from different modalities for diagnostic prediction. We denote clinical images as C, dermoscopic images as D, and textual metadata as T.
For datasets combining a single imaging modality with structured metadata (C+T or D+T), images are encoded through DermFM-Zero's vision encoder, while metadata are converted to text via templates and encoded by the text encoder, followed by a linear projection layer. The meta-fusion transformer then fuses these representations, with textual features as queries attending to visual features as keys and values via cross-attention across multiple layers.
For datasets with all three modalities (C+D+T), the architecture employs parallel fusion pathways. Clinical and dermoscopic images are encoded through two separate vision encoders, both initialized from DermFM-Zero's pretrained weights. The cross-transformer then performs bidirectional cross-attention between these visual representations across multiple layers, producing clinical-to-dermoscopic and dermoscopic-to-clinical fused features. Simultaneously, the meta-fusion transformer integrates textual features (as queries) with concatenated original visual features (as keys and values). The final representation concatenates the two cross-fused visual features and the textual-fused feature feeding into an MLP classifier.
All encoder weights are fine-tuned jointly during training, with the best checkpoint selected based on validation performance.

\noindent\textcolor{black}{\hheading{Training configuration for multimodal finetuning}}. For all multimodal finetuning experiments, we trained for 50 epochs using the Adam optimizer with learning rate 1e-5, $\beta_1=0.9$, $\beta_2=0.999$, and a cosine annealing learning rate scheduler with maximum period of 50 epochs. Training used batch size 32 with gradient accumulation over 2 steps. The cross-transformer was configured with 8 attention heads and 2 layers, while the meta-fusion transformer used 8 attention heads and 4 layers, except where noted below.


\noindent\textcolor{black}{\hheading{Multimodal evaluation on Derm7pt.}} Derm7pt contains 1,011 skin cancer cases(413/203/395 train/val/test split, following derm7pt official\cite{derm7pt} split) with paired clinical and dermoscopic images, 7-point diagnostic checklists (pigment network, dots/globules, streaks, regression areas, blue-whitish veil, vascular structures, pigmentation patterns), and patient metadata including sex, lesion elevation, and anatomical location. We integrated all three modalities (C+D+T) and selected the best checkpoint based on validation accuracy.

\noindent\textcolor{black}{\hheading{Multimodal evaluation on SCIN.}} SCIN comprises 2,140 cases across 20 common inflammatory and infectious conditions (1,284/428/428 train/val/test split, stratified by class). Metadata includes demographics (age group, sex, race/ethnicity, skin tone), lesion characteristics (texture, body location), and patient-reported symptoms, which were paired with clinical photographs. We employed the C+T multimodal architecture and selected the best checkpoint based on validation accuracy.

\noindent\textcolor{black}{\hheading{Multimodal evaluation on PAD.}} PAD includes 2,298 skin lesion cases (1,838/230/230 train/val/test split, stratified by lesion type) pairing clinical images with metadata comprising patient demographics (age, gender), lesion location (body region), risk factors, morphological features (diameter, elevation), clinical symptoms (itch, pain, bleeding), and temporal changes (growth, color change). We employed the C+T multimodal architecture and selected the best checkpoint based on validation accuracy.

\noindent\textcolor{black}{\hheading{Multimodal evaluation on Combinmel.}} The Combinmel dataset comprises 663 melanoma cases from 364 patients recruited across 10 hospital sites in Australia and Europe for binary metastasis prediction. Metadata include clinicopathological features (Breslow thickness, ulceration status, mitosis, lymphovascular invasion, microsatellites, melanoma subtype) and patient information (anatomical location, age, sex), paired with dermoscopic images. We evaluated using patient-stratified 5-fold cross-validation (70/10/20 train/val/test split per fold). For this dataset, the cross-transformer was configured with 4 layers instead of 2. We employed the D+T multimodal architecture and selected the best checkpoint based on validation AUROC with an early stopping patience of 10 epochs.

\heading{Eval 3: Reader study} \\
\textbf{RS1: Human-AI collaboration in primary care.}
This reader study was conducted via a secure web-based interface (APEX Study) developed in-house at Monash University to simulate real-world teledermatology workflows (\textbf{Extended Data Fig.~\ref{fig:supp_r1_platform}}). The study recruited 30 general practitioners (GPs) from diverse geographic backgrounds (Australia, China, and Europe). Participants evaluated clinical cases randomly sampled from a bank of 150 cases sourced from public datasets (e.g., SNU-134, PASSION, SKINTONE) with ground truth revised by board-certified dermatologists. Each case consisted of a single clinical photograph simulating a store-and-forward consultation scenario spanning 98 distinct skin conditions. 

The study employed a sequential design. In the \textit{Pre-AI} phase, participants assessed the case independently using the web interface to provide: (1) a primary diagnosis via free-text entry, (2) differential diagnoses, (3) diagnostic confidence on a 5-point scale, and (4) a management plan (\textbf{Extended Data Fig.~\ref{fig:supp_r1_platform}a}). Immediately following the initial submission, the \textit{Post-AI} phase commenced, where participants were presented with DermFM-Zero's zero-shot top-3 predicted diagnoses and allowed to revise their diagnosis and management decisions (\textbf{Extended Data Fig.~\ref{fig:supp_r1_platform}b}). To simulate real-world diagnostic workflows, free-text diagnoses entered by readers were standardized using automated mapping and graded by a GPT-4o-based evaluation agent, followed by dermatologist review and confirmation. Outcomes were assessed by expert dermatologists using detailed scoring rubrics. Diagnostic accuracy was evaluated on a 5-point scale: 1 (Potential Harm), 2 (Different Class), 3 (Same Class), 4 (Generic Correct), and 5 (Spot On). Management quality was assessed on a 4-point scale: 1 (Inadequate/Dangerous), 2 (Inadequate/Harmless), 3 (Adequate), and 4 (Perfect).

\noindent\textbf{RS2A: Human vs AI for multimodal skin cancer diagnosis.}
To benchmark DermFM-Zero against a large cohort of clinicians in a specialized multimodal setting, we utilised the in-house Test Of Dermoscopy for International Validation (TODIV) dataset, comprising 1,117 standardized clinical cases. Each case contained a clinical description (demographics, risk factors), a macroscopic photograph, and a polarized dermoscopic image. The study included 1,090 clinicians (comprising 284 general practitioners and 762 dermatologists) with varying levels of experience. Participants classified each lesion into one of nine diagnostic categories (melanoma, nevus, basal cell carcinoma, squamous cell carcinoma or actinic keratosis, seborrheic keratosis or solar lentigo, dermatofibroma, hemangioma or angiokeratoma, other benign, and other malignant). DermFM-Zero was evaluated in a zero-shot multimodal setting (Clinical + Dermoscopy) on the same set of 1,117 cases. Additionally, we compared performance against a supervised baseline model (Ypsono~\cite{humanai}) to assess the advantage of the foundation model approach over traditional task-specific CNNs.

\textbf{RS2B: Human-AI collaboration in specialty care for skin cancer diagnosis and management.}
We conducted this study using DermaChallenge, a website hosted by the Medical University of Vienna and utilised in prior research~\cite{panderm,humanai}. We recruited 34 clinicians, stratified into non-experts ($n=18$; including dermatology residents and generalists) and experts ($n=16$; board-certified dermatologists), to evaluate complex skin lesions suspected of malignancy using the test set of MILK10K~\cite{milk10k}. The dataset consisted of cases spanning 11 distinct lesion classes. Each case presented paired clinical and dermoscopic images with basic meta information.

To support specialist-level assessment, the website featured a dual-modality viewer with an interactive slider, allowing participants to seamlessly toggle between clinical macro images and dermoscopy (\textbf{Extended Data Fig.~\ref{fig:specialist_r2b_platform}a}). The study utilised a two-step sequential design. First, participants diagnosed the lesion and selected a management strategy (Dismiss, Monitor, Treat, or Excise) based on visual inspection and basic meta-information. Subsequently, they were presented with DermFM-Zero's AI support, visualized as probability bars for the top-3 predictions, and were given the opportunity to update their decisions (\textbf{Extended Data Fig.~\ref{fig:specialist_r2b_platform}b}). Management decisions were evaluated against a predefined matrix mapping diagnoses to optimal clinical actions, graded as Inappropriate, Appropriate, or Optimal. This experimental setup allowed us to assess the impact of DermFM-Zero on bridging the expertise gap within multimodal diagnostic settings.

\noindent\textcolor{black}{\heading{Automated concept discovery}}

To assess whether DermFM-Zero internally organizes visual information into clinically interpretable features, we employed a three-stage approach combining sparse autoencoder training, concept filtering, and automated naming. The first stage involved training a sparse autoencoder (SAE) to decompose DermFM-Zero's vision features into localized visual patterns. The SAE architecture consists of a linear encoder with weights $W_E$, a ReLU activation function $\phi$, and a linear decoder with weights $W_D$. For a given input visual feature $x$, the reconstruction process is computed as: $\mathrm{SAE}(x) = W_D^{\top}\phi(W_E^{\top}x)$. We trained the SAE on DermFM-Zero's visual features extracted from Derm1M\cite{derm1m} by minimizing a reconstruction loss combined with L1 sparsity regularization: $\mathcal{L} = \|\mathrm{SAE}(x) - x\|_2^2 + \lambda\|\phi(W_E^{\top}x)\|_1$, with $\lambda$ = 3e-5. Here, we set the latent dimension to 8 times the input feature dimension. The SAE training used the Adam optimizer with learning rate $5 \times 10^{-4}$, $\beta_1=0.9$, $\beta_2=0.999$, and batch size 4,096. This sparsity constraint forces each SAE latent dimension to specialize in capturing distinct morphological concepts. In the second stage, we identified diagnostically relevant concepts by training an L1-regularized logistic regression classifier on the SAE latent activations to predict disease labels on diagnostic datasets. We used stochastic gradient descent with log loss, L1 penalty strength $\alpha=0.001$. The L1 regularization induces sparsity in the classifier weights, setting uninformative concept weights to zero and retaining only diagnostically relevant concepts. Only concepts with non-zero weights were retained for subsequent interpretation. The final stage involved mapping these retained concepts to a predefined medical vocabulary $\mathcal{V}$. Each vocabulary term was encoded using DermFM-Zero's text encoder, while each SAE concept was represented by its corresponding decoder weight column $p_c \in W_D$. The Hungarian algorithm determined the optimal one-to-one assignment $\pi^*$ that maximizes the total cosine similarity $\sum_{c} \mathrm{sim}(p_c, e_{\pi(c)})$, where $e_{\pi(c)}$ denotes the embedding of the matched vocabulary term. This procedure yielded human-interpretable names for each diagnostically relevant visual concept.

\noindent\textcolor{black}{\heading{T-SNE visualization.}} 

To visualize the learned feature representations, we applied t-distributed stochastic neighbor embedding (t-SNE) to DermFM-Zero's visual features. We selected the 15 most frequent disease classes from SD-128\cite{sd198}, comprising 900 images in total. Features were projected into two dimensions using t-SNE with perplexity 8, learning rate 1000, and cosine distance metric, initialized with PCA and run for 2,500 iterations.

\heading{Statistical analysis}\\

For zero-shot classification, linear probing, and cross-modal retrieval tasks, we employed non-parametric bootstrapping with 1,000 replicates to estimate 95\% confidence intervals (CIs) for performance metrics. To compare DermFM-Zero against baseline models, we conducted two-sided t-tests on bootstrap distributions. For multimodal fine-tuning tasks employing 5-fold cross-validation, we aggregated results as the mean and standard deviation (SD) across folds. The 95\% confidence intervals (CIs) were calculated as $\text{mean} \pm 1.96 \times \text{standard error (SE)}$, where $\text{SE} = \text{SD} / \sqrt{N}$ ($N=5$). Pairwise statistical significance for these tasks was assessed using two-sided $t$-tests. In Reader Study 1, we utilised one-sided Wilcoxon signed-rank tests to compare paired diagnostic and management decisions (unaided vs. AI-assisted). For Reader Studies 2A and 2B, performance differences were evaluated using two-sided $t$-tests (unpaired for Study 2A; paired for Study 2B). Continuous metrics are reported as means with 95\% CIs, supplemented by medians and interquartile ranges (IQRs) where data distribution warranted non-parametric description. For survival analysis (CombinMel), differences in recurrence-free intervals between risk groups were evaluated using log-rank tests. Hazard ratios (HRs) with 95\% CIs were estimated using Cox proportional hazards regression models. Predictive performance was assessed via time-dependent ROC curves generated for 3-, 5-, and 7-year time horizons. For concept discovery experiments, we compared AUROC metrics across different interpretability approaches using two-sided $t$-tests. All statistical analyses were conducted using Python scientific computing libraries, including SciPy (v1.13--1.15), NumPy (v1.26), and scikit-learn (v1.4--1.6). A $P$-value of $< 0.05$ was considered statistically significant.






\noindent\textcolor{black}{\heading{Computing hardware and software}}

All experiments were conducted using NVIDIA GPUs with CUDA acceleration. Model pretraining was performed on a single NVIDIA A100-SXM4-80GB GPU using Python 3.10.13, PyTorch 2.4.1, CUDA 11.8, and standard scientific computing libraries (NumPy 2.2.6, SciPy 1.15.2, pandas 2.2.3, scikit-learn 1.6.1, Pillow 11.2.1). Downstream evaluations including zero-shot classification, linear probing, and multimodal fine-tuning were conducted on single NVIDIA RTX 6000 Ada Generation GPUs (48 GB GPU memory) using Python 3.9.20, PyTorch 2.8.0, CUDA 12.8, and associated libraries (NumPy 1.26.4, SciPy 1.13.1, pandas 2.3.3, scikit-learn 1.6.1, scikit-image 0.24.0, OpenCV 4.12.0, Matplotlib 3.9.4, Seaborn 0.13.2, Pillow 11.3.0). Automated concept discovery experiments were performed on a single NVIDIA RTX 6000 Ada Generation GPU using Python 3.10.19, PyTorch 2.9.1, CUDA 12.8, and core libraries (NumPy 1.26.4, SciPy 1.15.3, pandas 2.3.3, scikit-learn 1.4.1, Matplotlib 3.10.7). We use open\_clip code to develop our foundation model, which can be found in their official github repository (\url{https://github.com/mlfoundations/open_clip}). Pretrained baseline models were obtained from their official github or huggingface repositories: CLIP-L/14 (\url{https://huggingface.co/timm/vit_large_patch14_clip_224.openai}), DINOv3-L16 (\url{https://huggingface.co/facebook/dinov3-vitl16-pretrain-lvd1689m}), DINOv3-7B (\url{https://huggingface.co/facebook/dinov3-vit7b16-pretrain-lvd1689m}), BioMedCLIP (\url{https://huggingface.co/microsoft/BiomedCLIP-PubMedBERT_256-vit_base_patch16_224}), DermLIP (\url{https://huggingface.co/redlessone/DermLIP_ViT-B-16}), MAKE (\url{https://huggingface.co/xieji-x/MAKE}), MONET (\url{https://github.com/suinleelab/MONET}), and PanDerm (\url{https://github.com/SiyuanYan1/PanDerm}).

\noindent\textcolor{black}{\heading{Data availability}}

Most of the datasets used in this study are publicly available. These datasets were used for zero-shot classification, cross-modal retrieval, linear probing, multimodal fine-tuning, and concept discovery tasks. The ISIC Archive (\url{https://challenge.isic-archive.com/data/}) hosts several datasets, including HAM10000, ISIC-Intervention, and ISIC2020. Other benchmark datasets are available through their respective repositories: PH2 (\url{https://www.fc.up.pt/addi/ph2%20database.html}), DAFFODIL (\url{https://data.mendeley.com/datasets/3hckgznc67/1}), PAD-UFES-20 (\url{https://www.kaggle.com/datasets/mahdavi1202/skin-cancer}), SNU Dermatology (\url{https://figshare.com/articles/dataset/6454973}), SD-128 (\url{https://huggingface.co/datasets/resyhgerwshshgdfghsdfgh/SD-198}), Derm1M (\url{https://huggingface.co/datasets/redlessone/Derm1M}), SkinCap (\url{https://huggingface.co/datasets/joshuachou/SkinCAP}), Derm7pt (\url{https://derm.cs.sfu.ca/Welcome.html}), SCIN (\url{https://github.com/google-research-datasets/scin/}), PUMCH-ISD (\url{https://www.kaggle.com/datasets/jcwang10000/pumch-isd/data}), Fitzpatrick17k (\url{https://github.com/mattgroh/fitzpatrick17k}), and DDI (\url{https://ddi-dataset.github.io/}).

\noindent\textcolor{black}{\heading{Code availability}}\\
All experiment code for this study is publicly available. The GitHub repository (\url{https://github.com/SiyuanYan1/DermFM-Zero}) contains implementation code for all downstream tasks, including zero-shot classification, cross-modal retrieval, linear probing, multimodal fine-tuning, and automated concept discovery. The repository also includes benchmark datasets for evaluation, tutorial Jupyter notebooks demonstrating model usage, and detailed documentation. Pretrained model weights for DermFM-Zero are released on HuggingFace (\url{https://huggingface.co/redlessone/DermFM-Zero}). Our code is released to support reproducibility and facilitate future research in dermatology AI.

\heading{Author contributions}\\
"S.Y. and Z.G. conceived and designed the study. S.Y., H.P.S., V.M., H.K., and Z.G. designed the reader studies. S.Y., D.M., P.T., J.L., C.C., J.G.C., J.A., L.T., A.I., X.L., A.B., M.H., and S.Z. conducted the reader studies. S.Y. and X.L. developed the model and performed benchmark evaluations. S.Y., X.L., Y.J., Z.W., M.H., L.J., C.V.A., J.Z., J.A., Y.Z., K.T., and X.T. contributed to data collection and curation. G.T., A.B.N., and S.S. provided computing resources and data management. P.T., C.V.A., C.P., M.J., H.P.S., V.M., and H.K. provided clinical expertise and oversight. S.Y. drafted the manuscript. All authors reviewed and approved the final manuscript.

\heading{Acknowledgements}\\
We thank MoleMap for providing data support for model development.

\heading{Competing interests }\\
Z.G., V.M., H.P.S., M.J., and P.G. are Chief Investigators for the Australian Centre of Excellence for Melanoma Imaging and Diagnosis (ACEMID), which was established via an Australian Cancer Research Foundation Major Infrastructure Grant, with research activities supported by NHMRC grants (Cohort Study Grant APP2001517, Centre of Research Excellence Grant APP2044753, Synergy Grant APP2009923) and MRFF Targeted Health System \& Community Organisation Research Grant (APP1175082).
Z.G. is on the scientific advisory board and consultant for Optain Health. Although Airdoc has philanthropic donation to the AIM for Health Lab, the company was not involved in any aspect of this research.
H.P.S. reported equity in e-derm-consult GmbH and MoleMap NZ Limited; consulting fees from Canfield Scientific Inc and a patent (PCT/AU/2013/000394) licensed to Trajan Medical and Scientific via Uniquest, all outside the submitted work. He is also an Executive Board Member of the International Dermoscopy Society and the Australian Melanoma Clinical Outcome Registry.
M.J. holds National Health and Medical Research Council (NHMRC) TRIP Fellowships (APP2006551, APP2009923, and APP2034422).
P.T. has received speaker fees from AbbVie and unrestricted educational grants from Lilly. He is an Executive Board Member of the International Dermoscopy Society and Past-President of the Austrian Society of Dermatopathology.
S.S., V.T., and A.B.N are employees of NVIDIA and own Restricted Stocks.
H.K. has received speaker fees from Fotofinder, MSD, Novartis, and Pelpharma; license fees from Casio; and equipment from Fotofinder, Casio, and Heine. He has served as an advisor for Fotofinder, La Roche-Posay and AI Medical Technology, and is a member of the executive board of the International Dermoscopy Society.
P.G. has received honoraria from Metaoptima PTY and travel stipend from L'Oreal.
V.M. is supported by an NHMRC Investigator Grant (APP2034976). V.M. has received Victorian Medical Research Acceleration Fund support for the SMARTI Trial with matched contribution from MoleMap, speaker fees from Novartis, Bristol Myers Squibb, Merck and Janssen, conference travel support from L'Oreal and has participated in Advisory Boards for MSD, L'Oreal and SkylineDx. V.M. is a Board Member of the Melanoma and Skin cancer Trials Group and Advisory Member for the Melanoma and Skin Cancer Advocacy Network. The remaining authors declare no competing interests.

\end{spacing}
\clearpage
\setcounter{figure}{0}
\renewcommand{\figurename}{Extended Data Figure}
\vspace{2mm}
\begin{figure*}
\centering
\vspace{-5mm}
\includegraphics[width=0.95\textwidth]{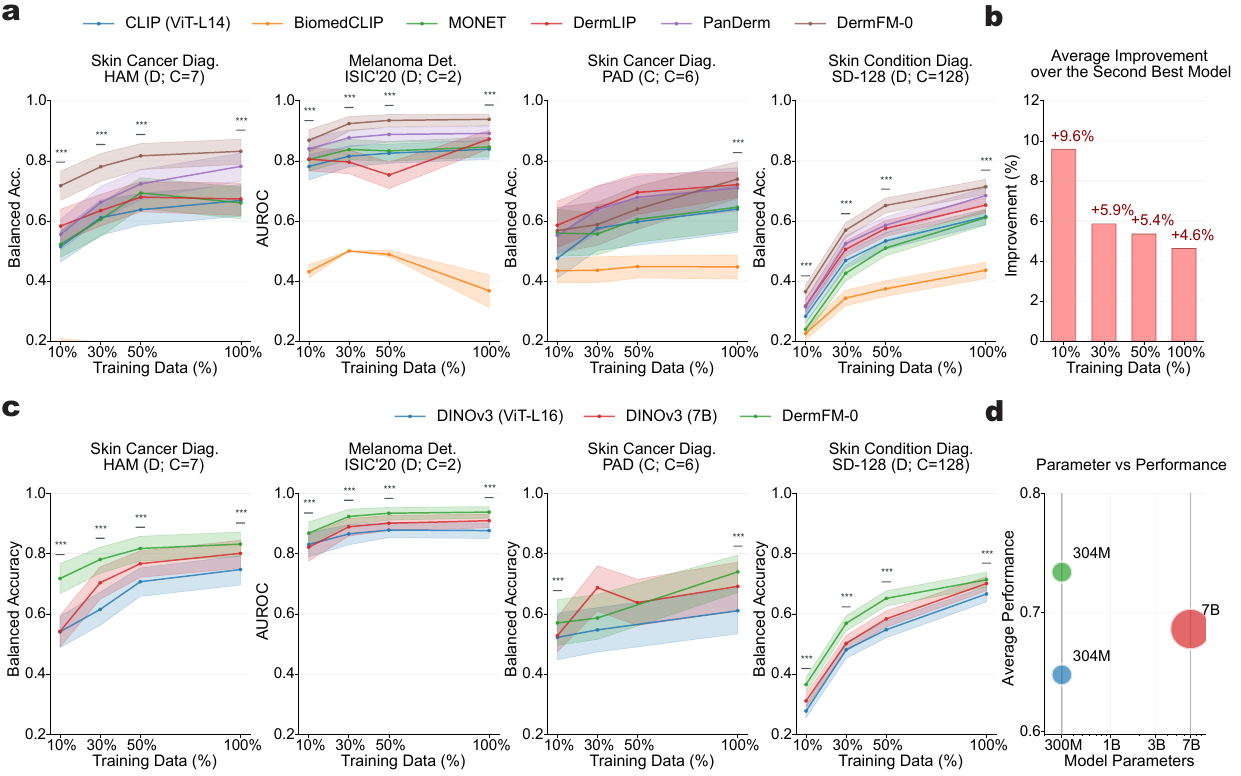}
\caption{\small \textbf{Label-efficient generalization via linear probing.} 
\textbf{a, c,} Label-efficient generalization performance of DermFM-Zero at various training data percentages across diverse tasks. Performance is compared against other medical foundation models (\textbf{a}) and a significantly larger natural-domain model (DINOv3) (\textbf{c}). 
\textbf{b,} Average performance improvement (from \textbf{a}) of DermFM-Zero over the second-best model (PanDerm) across different training data ratios. 
\textbf{d,} Performance versus parameter size, comparing DermFM-Zero (304M parameters) with DINOv3, which is ~23$\times$ larger (7B parameters). 
In \textbf{a, c}, shaded bands indicate 95\% CIs (centre line for the mean), computed via non-parametric bootstrapping (1,000 replicates). Pairwise significance was determined by a two-sided t-test (*\textit{P} < 0.05, **\textit{P} < 0.01, ***\textit{P} < 0.001).
}
\label{supp_lp}
\end{figure*}

\begin{figure*}
\centering
\vspace{-5mm}
\includegraphics[width=0.95\textwidth]{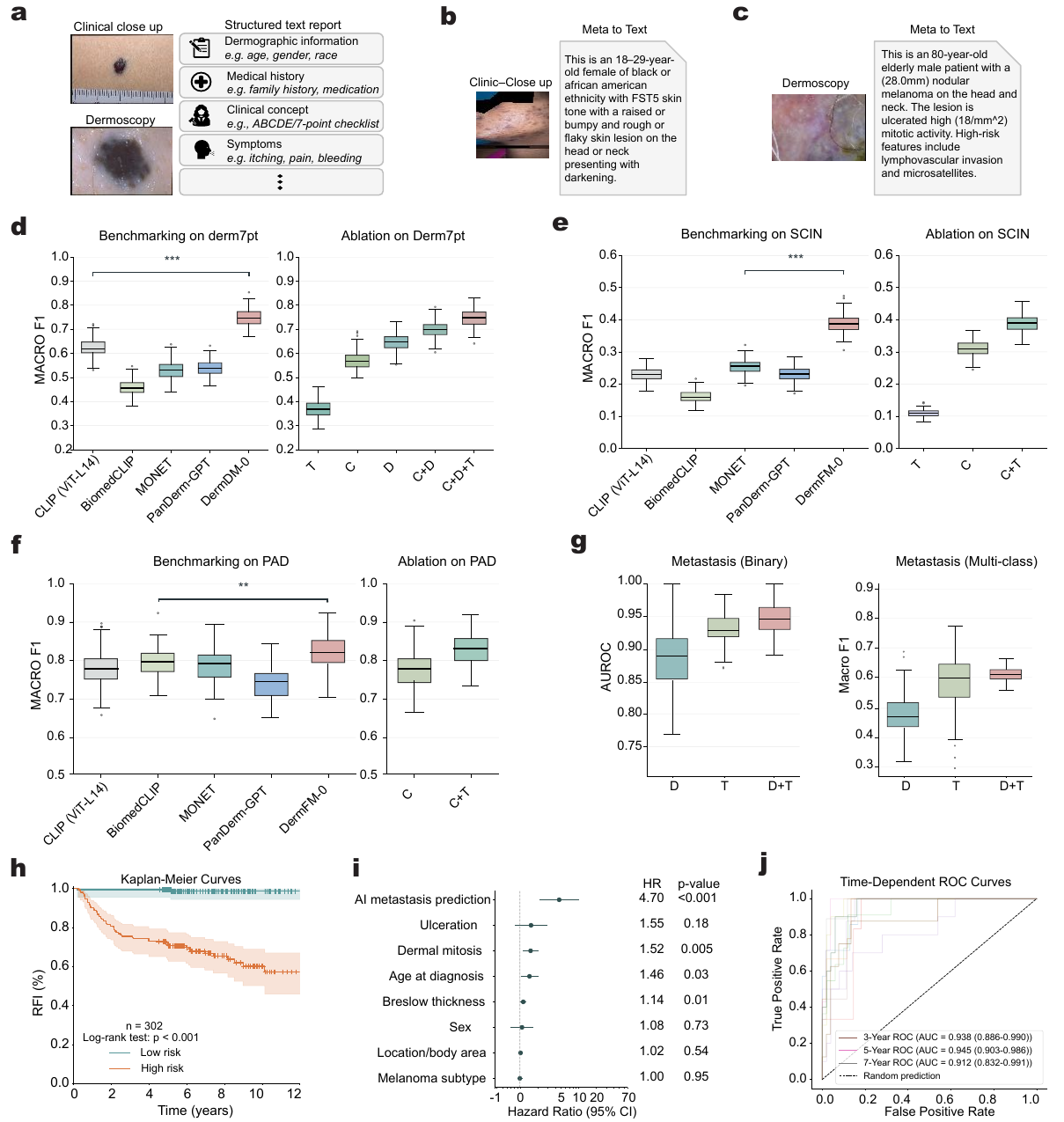}
\caption{\small \textbf{Performance evaluation under real-world multimodal settings.} 
\textbf{a,} The real-world multimodal setting, integrating image modalities (clinical, dermoscopy) and structured text. 
\textbf{b, c,} Example image-text pairs from SCIN (b) and CombinMel (c), with text generated from metadata. 
\textbf{d-f,} Multimodal finetuning performance on Derm7pt (d; skin cancer diagnosis (C+D+T)), SCIN (e; skin condition classification (C+T)), and PAD (f; skin cancer diagnosis (C+T)). Left: Benchmarking DermFM-Zero against other models. Right: Modality ablation (C=clinical, D=dermoscopy, T=text). 
\textbf{g,} Modality ablation for binary and multi-class melanoma metastasis prediction (D+T). 
\textbf{h,} Kaplan-Meier curves for recurrence-free interval (RFI) in CombinMel (n=302), stratified by DermFM-Zero risk predictions. 
\textbf{i,} Forest plot of Hazard Ratios (HRs) for RFI, comparing DermFM-Zero with other clinical variables. 
\textbf{j,} Time-dependent ROC curves for 3, 5, and 7-year RFI prediction. 
In \textbf{d-g}, boxes show median and IQR; whiskers show 1.5$\times$IQR. In \textbf{h}, shaded areas are 95\% CIs; \textit{P}-value ($<0.001$) is from a log-rank test. In \textbf{i}, error bars are 95\% CIs for HRs. In \textbf{j}, values in parentheses are 95\% CIs for AUCs. Asterisks in \textbf{d-f} denote statistical significance (*\textit{P} < 0.05, **\textit{P} < 0.01, ***\textit{P} < 0.001) from a two-sided t-test.
}
\label{supp_multimodal}
\end{figure*}

\begin{figure*}
\centering
\vspace{-5mm}
\includegraphics[width=0.9\textwidth]{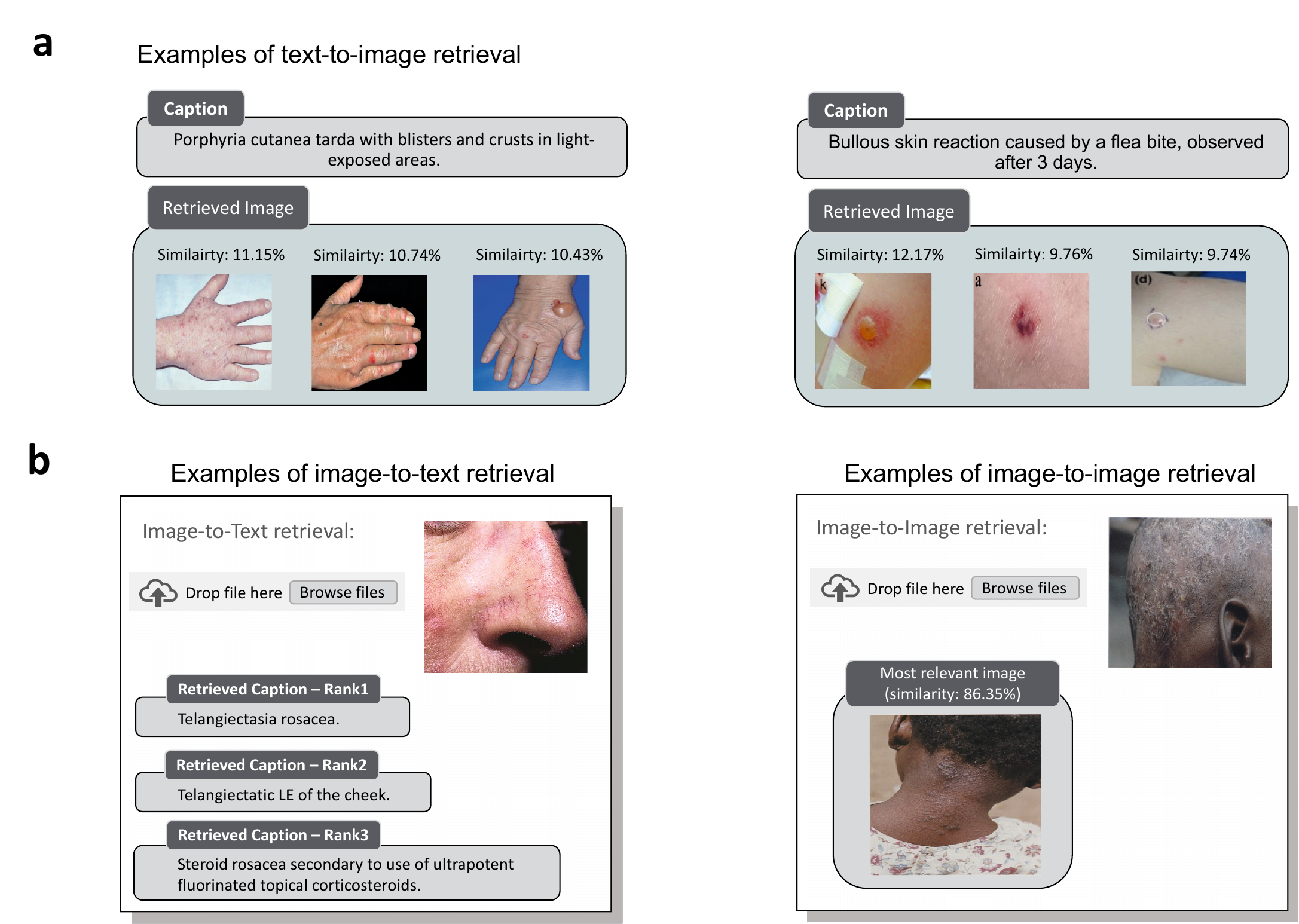}
\caption{\small\textbf{Cross-modal and unimodal retrieval examples.} \textbf{a,} Text-to-image retrieval showing top-3 retrieved images with similarity scores for two clinical text queries: porphyria cutanea tarda (left) and bullous flea bite reaction (right). \textbf{b,} Image-to-text retrieval (left) displaying top-3 retrieved captions for a facial telangiectasia query image, and image-to-image retrieval (right) showing the most similar image match (86.35\% similarity). Retrieved results demonstrate DermFM-Zero's multimodal alignment across text and visual dermatological content.}
\label{supp_retrieval}
\end{figure*}

\begin{figure*}
\centering
\includegraphics[width=0.9\textwidth]{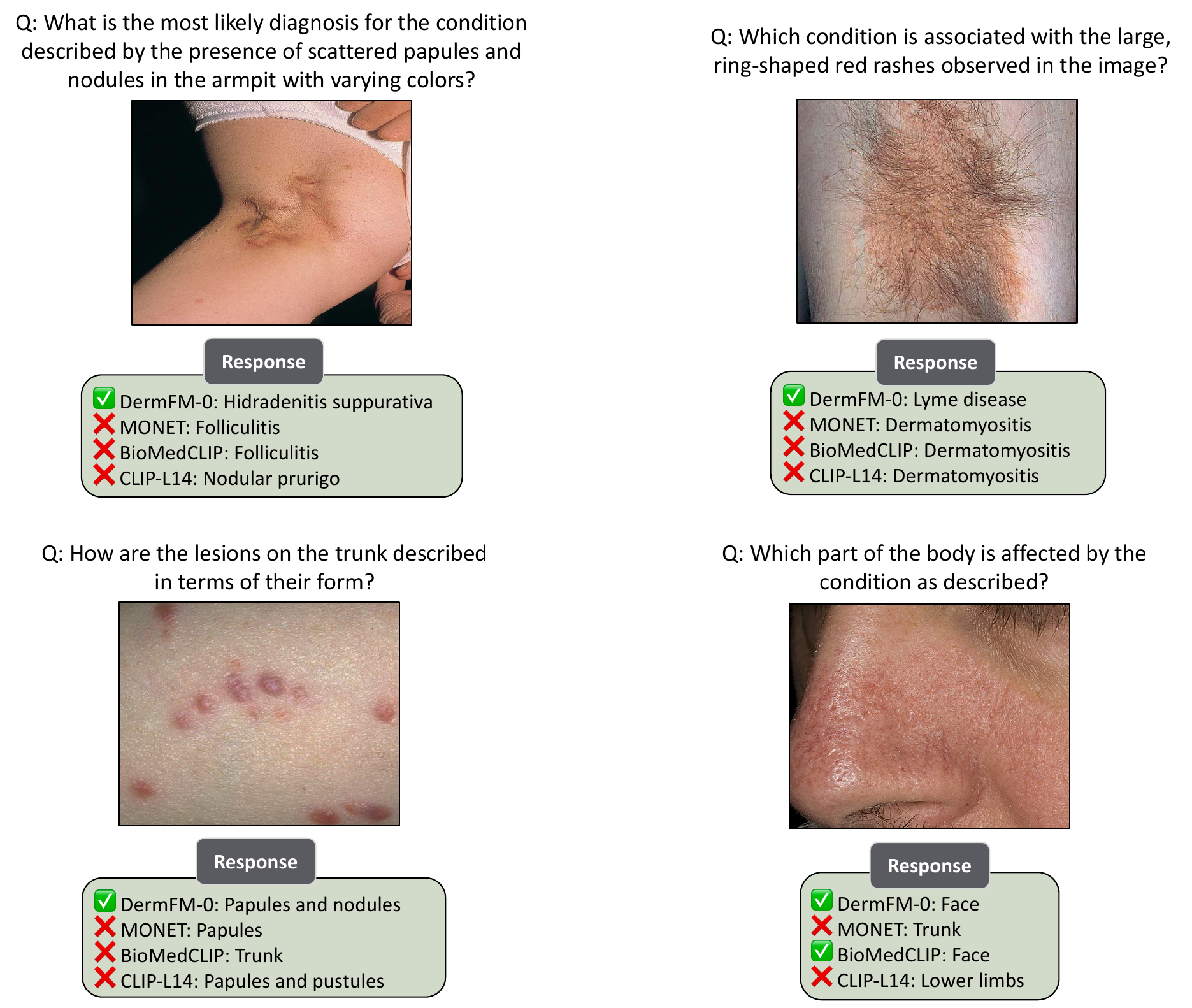}
\caption{\small\textbf{Visual question answering examples on clinical cases.} Representative question-answer pairs across four diagnostic scenarios comparing DermFM-0 with baseline models (MONET, BioMedCLIP, CLIP-L14). Questions span diagnosis identification, lesion morphology description, and anatomical localization. Green checkmarks and red crosses denote correct and incorrect responses, respectively.}
\label{supp_vqa}
\end{figure*}

\begin{figure*}
\centering
\vspace{-3mm}
\includegraphics[width=1.0\textwidth]{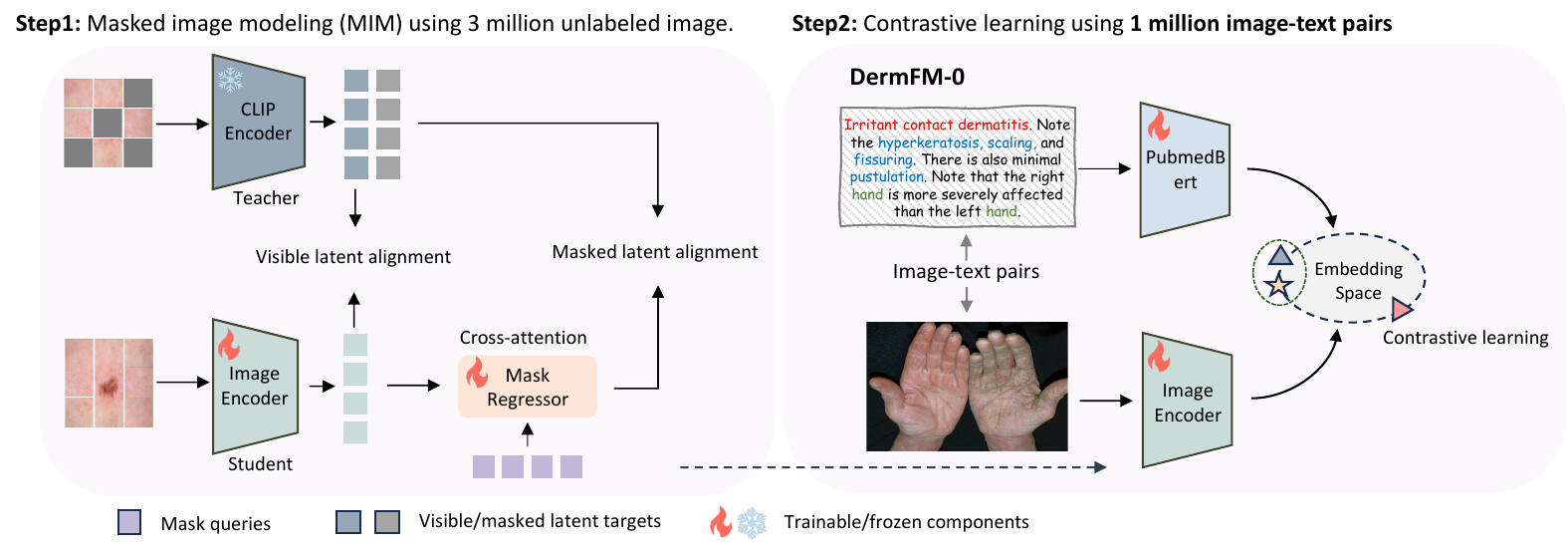}
\caption{\small\textbf{Two-steps pretraining pipeline for DermFM-0.} \textbf{Step 1:} Masked image modeling on 3 million unlabeled images using student-teacher learning with visible and masked latent alignment. \textbf{Step 2:} Contrastive learning on 1 million image-text pairs to align the pretrained vision encoder with a PubMedBERT text encoder in a shared embedding space.}
\label{supp_method1}
\end{figure*}

\begin{figure*}
\centering
\includegraphics[width=0.75\textwidth]{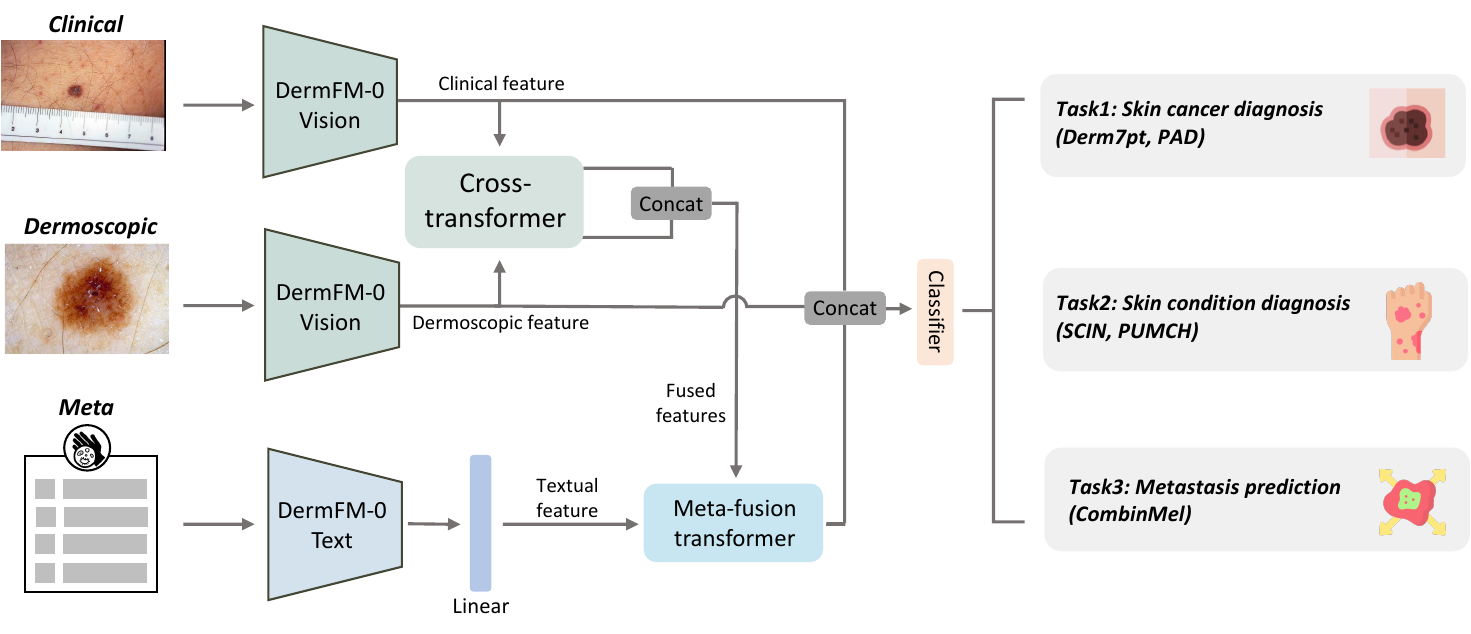}
\caption{\small\textbf{Multimodal fine-tuning architecture.} The framework processes clinical images, dermoscopic images, and metadata through parallel pathways. Visual features from separate DermFM-0 vision encoders are fused via cross-transformer, while text-encoded metadata are integrated through meta-fusion transformer. The concatenated representations feed into a classification head. Three independent tasks demonstrate the framework's versatility: skin cancer diagnosis (Derm7pt, PAD), general skin condition diagnosis (SCIN, PUMCH), and metastasis prediction (CombinMel), each trained separately with modality combinations adapted to dataset composition.}
\label{supp_method2}
\end{figure*}

\begin{figure*}
\centering
\includegraphics[width=0.95\textwidth]{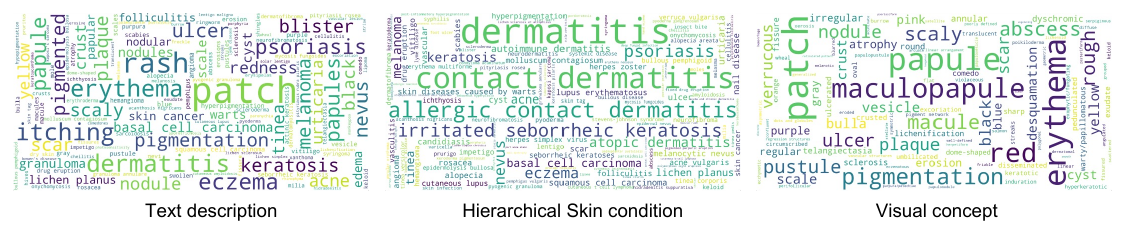}
\caption{\small\textbf{Medical vocabulary in DermFM-Zero pretraining data.} Word clouds showing term frequency across three categories from the pretraining corpus: general clinical descriptions (left), hierarchical skin conditions including inflammatory disorders and malignancies (center), and visual morphological concepts (right). Word size indicates frequency in the 1 million image-text pairs, demonstrating comprehensive coverage of dermatological terminology and clinical concepts.}
\label{supp_word_cloud}
\end{figure*}

\begin{figure*}[t]
    \centering
    \begin{subfigure}[b]{0.49\textwidth}
        \centering
        \includegraphics[width=\linewidth]{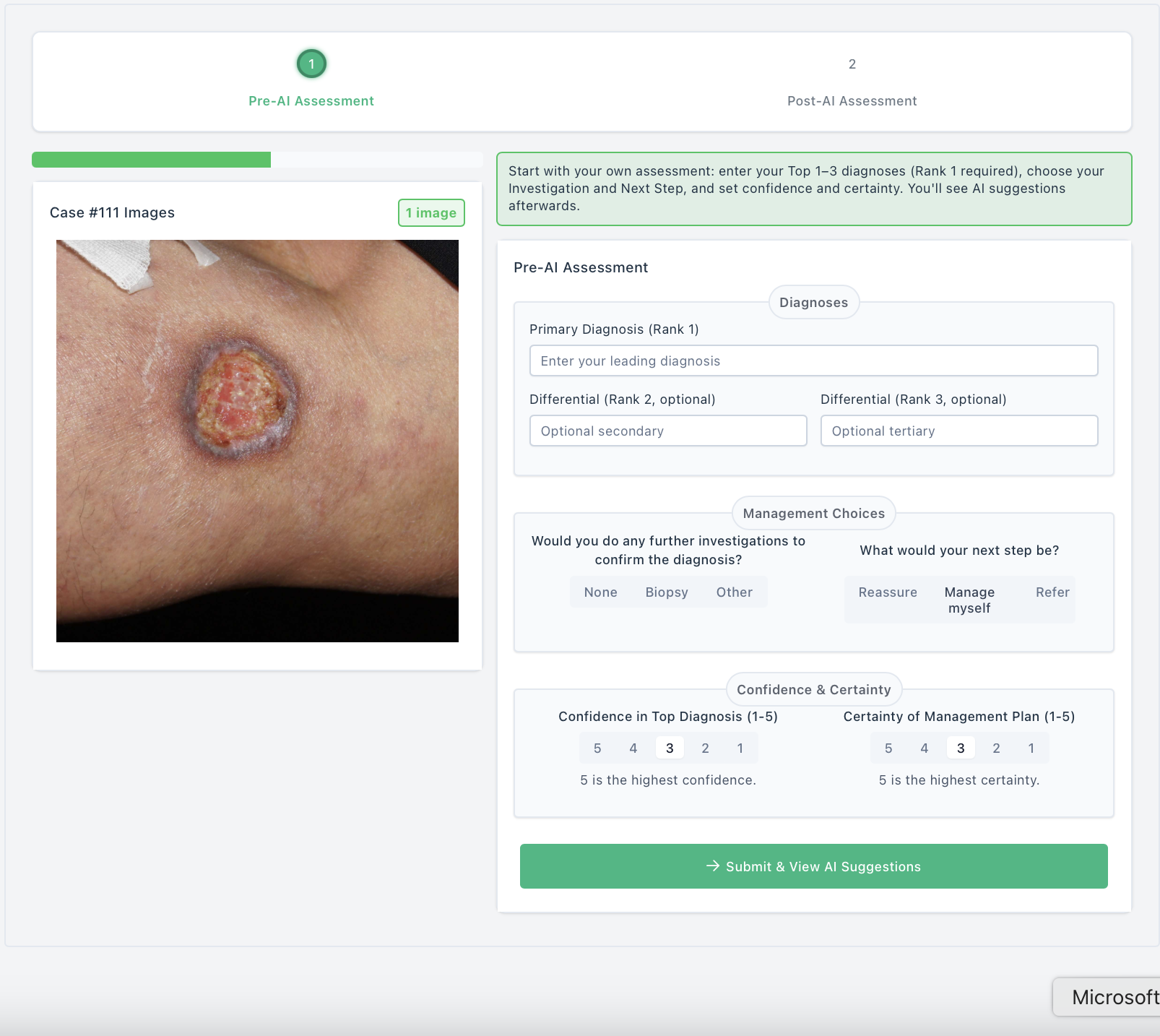} 
        \caption{}
        \label{fig:reader_study_ui_a}
    \end{subfigure}
    \hfill
    \begin{subfigure}[b]{0.49\textwidth}
        \centering
        \includegraphics[width=\linewidth]{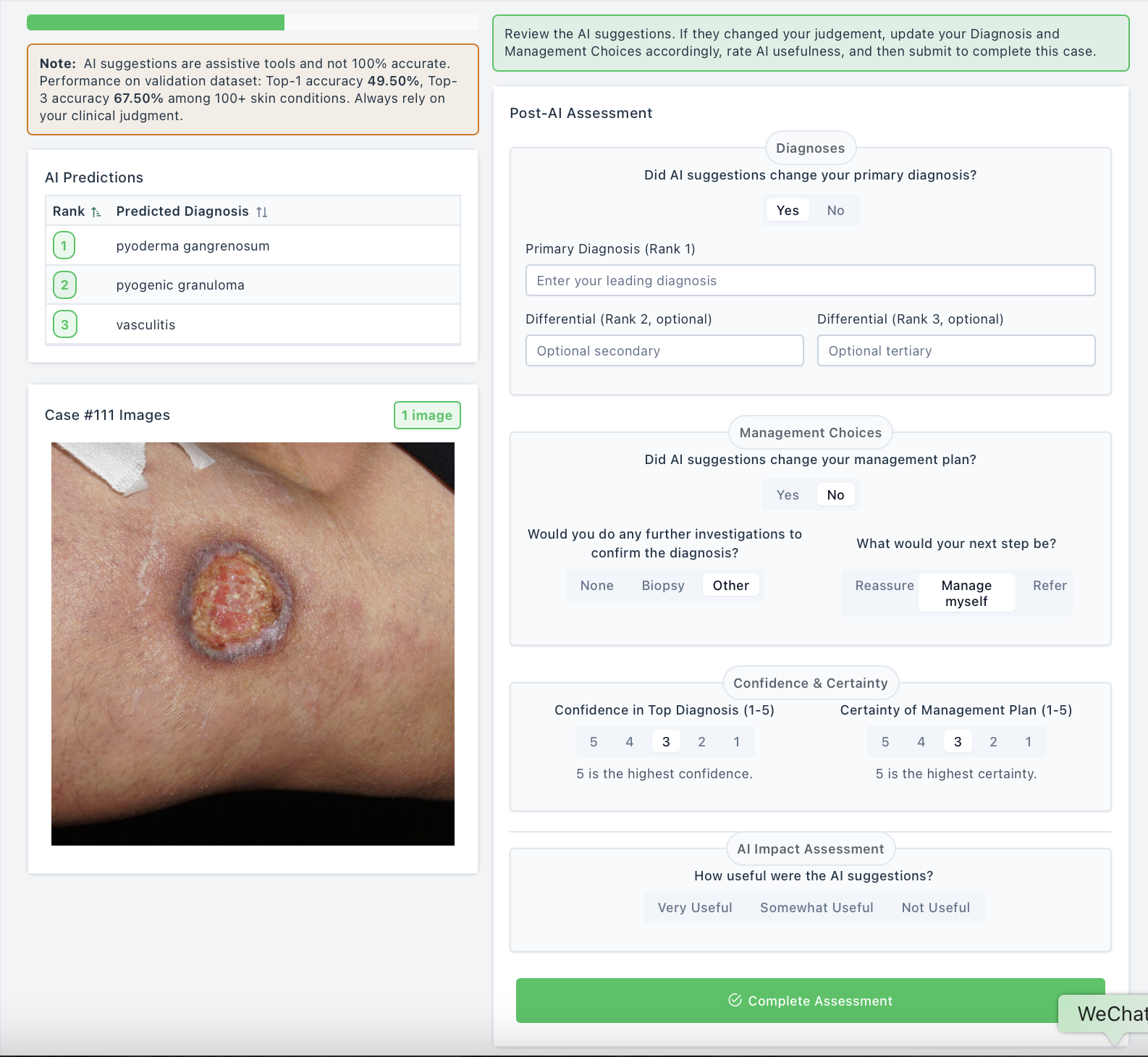}
        \caption{}
        \label{fig:reader_study_ui_b}
    \end{subfigure}
    
    \caption{\textbf{Web-based interface design for the primary care human--AI collaboration reader study 1.} 
    \textbf{a}, Pre-AI assessment phase. General practitioners view the clinical case (e.g., Case \#41 exhibiting herpes zoster) without AI assistance. The interface captures the clinician's independent primary diagnosis (free-text), optional differentials, management decisions (biopsy/investigation and referral/management steps), and self-reported confidence/certainty scores (Likert scale 1--5).
    \textbf{b}, Post-AI assessment phase. Immediately following the initial submission, the interface reveals DermFM-Zero's zero-shot analysis, displaying the top-3 predicted diagnoses (e.g., Rank 1: herpes zoster) along with model performance context. Clinicians are prompted to review these suggestions, indicate whether the AI output altered their diagnosis or management plan (Yes/No toggles), and evaluate the subjective utility of the AI assistance. This two-stage sequential design enables the direct quantification of AI impact on diagnostic accuracy and clinical decision-making.}
    \label{fig:supp_r1_platform}
\end{figure*}

\begin{figure*}[t]
    \centering
    \begin{subfigure}[b]{0.6\textwidth}
        \centering
        \includegraphics[width=\linewidth]{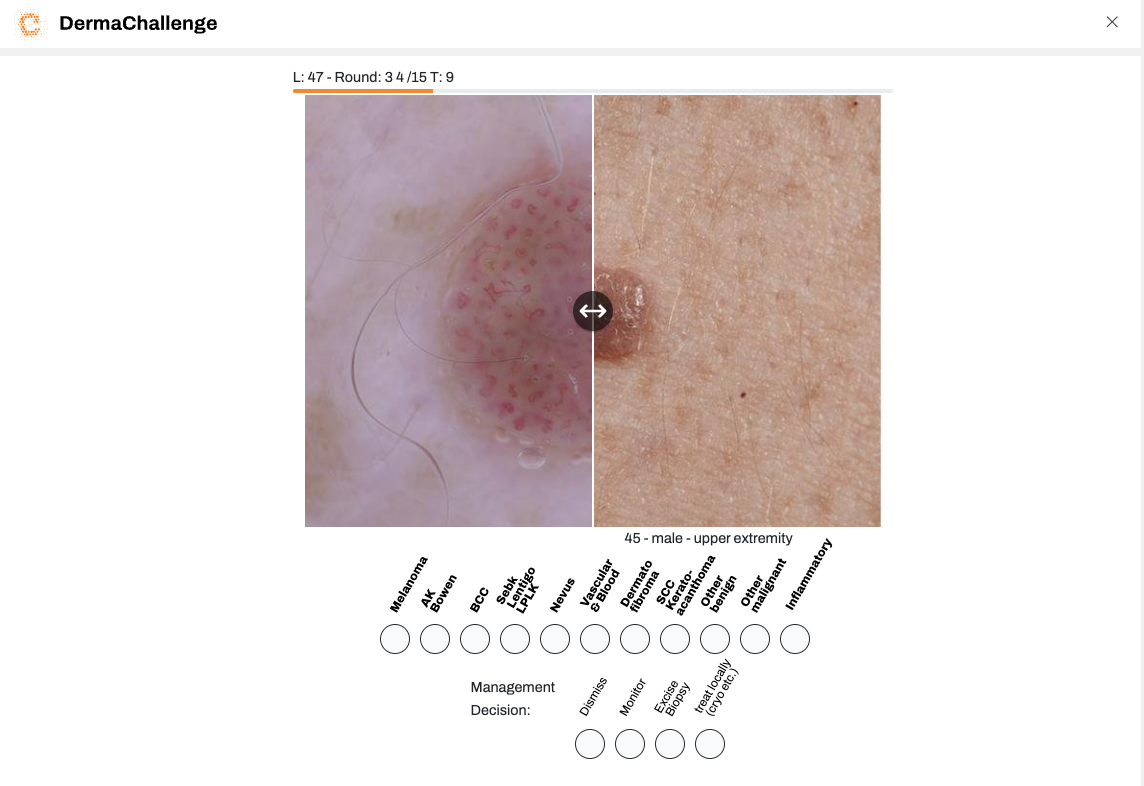} 
        \caption{}
        \label{fig:specialist_ui_a}
    \end{subfigure}

    \par\bigskip 
    
    \begin{subfigure}[b]{0.6\textwidth}
        \centering
        \includegraphics[width=\linewidth]{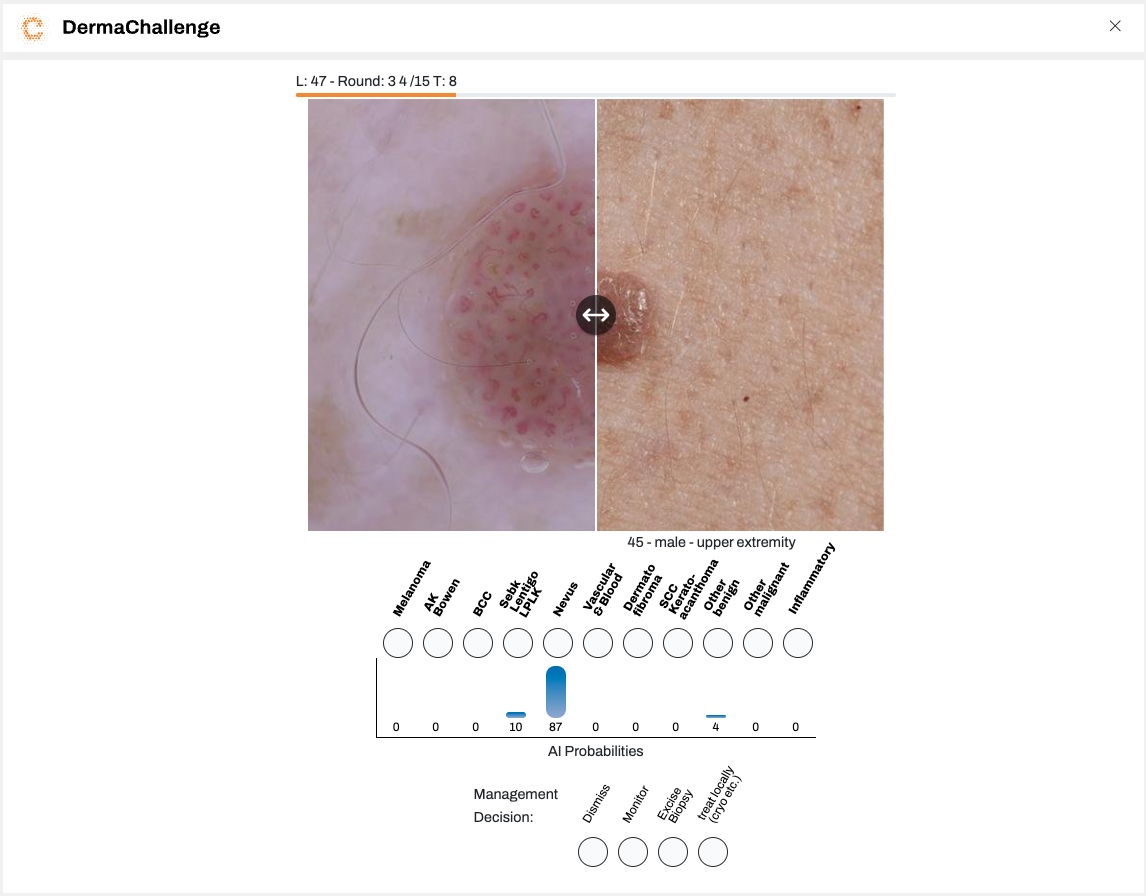}
        \caption{}
        \label{fig:specialist_r}
    \end{subfigure}
    
    \caption{\textbf{Interface design for the multimodal specialist collaboration reader study 2B.} 
    \textbf{a}, Pre-AI assessment phase. Clinicians evaluate complex multimodal cases with basic meta information using an interactive image viewer that allows seamless toggling between dermoscopic (left) and clinical (right) macro images via a central slider. Participants independently select a primary diagnosis from 11 diagnostic categories and choose a management strategy.
    \textbf{b}, Post-AI assessment phase. Upon submitting the initial decision, DermFM-Zero's zero-shot analysis is overlaid as probability bars beneath each diagnostic category (e.g., highlighting `Nevus' with a score of 87). This sequential workflow isolates the specific contribution of AI assistance to specialist-level decision-making.}
    \label{fig:specialist_r2b_platform}
\end{figure*}
\clearpage

\setcounter{table}{0}
\renewcommand{\tablename}{Extended Data Table}
\noindent\textcolor{black}{\heading{Supplementary Methods: Detailed Dataset Descriptions}}\\

\noindent\textbf{Combinmel (Metastasis).} This dataset comprises 663 melanoma cases from 364 patients recruited across 10 hospital sites in Australia and five European countries for binary metastasis prediction. Each case pairs a dermoscopic image with comprehensive clinicopathological metadata: Breslow thickness, ulceration status, mitosis count, lymphovascular invasion, microsatellites presence, melanoma subtype (superficial spreading, nodular, lentigo maligna, acral lentiginous), anatomical location, patient age, and sex. The multi-center collection ensures geographic and demographic diversity. We employed patient-stratified 5-fold cross-validation with 70/10/20 train/validation/test splits per fold to prevent data leakage between splits from the same patient.

\noindent\textbf{DAFFODIL-5\cite{Daffodil}.} The DAFFODIL dataset focuses on rare inflammatory and pigmentary skin conditions. From the original 9,548 images, we sampled a 20\% subset stratified by class label, resulting in 1,910 clinical photographs spanning five diagnostic categories: vitiligo, acne, Stevens-Johnson syndrome/toxic epidermal necrolysis (SJS/TEN), hyperpigmentation, and nail psoriasis. Images were collected from multiple centers and represent challenging diagnostic cases with varying presentation severity and anatomical locations.

\noindent\textbf{Derm1M Validation Set.\cite{derm1m}} The Derm1M validation set contains 9,806 image-text pairs sourced from diverse educational and clinical materials including YouTube videos, medical forums, PubMed articles, public datasets, and educational websites. Text descriptions range from brief captions to detailed case reports. This dataset evaluates cross-modal alignment in realistic scenarios where text varies in length, style, and medical terminology complexity.

\noindent\textbf{Derm7pt.\cite{derm7pt}} The 7-point checklist dataset contains 1,011 cases with paired clinical and dermoscopic images for skin cancer classification across five categories: melanoma, nevus, basal cell carcinoma, miscellaneous, and seborrheic keratosis. Each case includes structured 7-point checklist annotations covering pigment network, dots/globules, streaks, regression areas, blue-whitish veil, vascular structures, and pigmentation patterns, along with patient metadata (sex, lesion elevation, anatomical location). We use the official train/validation/test split of 413/203/395 cases for evaluation.

\noindent\textbf{Derm7pt Melanoma Subset.} From the complete Derm7pt dataset\cite{derm7pt}, we extracted 827 images containing only melanoma and nevus cases for binary melanoma detection. This subset enables focused evaluation of concept discovery for melanoma-related visual features.

\noindent\textbf{Derm7pt-VQA.} We constructed this dataset by applying handcrafted templates to the Derm7pt\cite{derm7pt} dataset's metadata columns, generating 4,000 question-answer pairs. Questions span seven types: concept existence (28.2\%), concept criteria (27.4\%), diagnosis (18.8\%), texture (9.0\%), management (8.4\%), modality (4.1\%), and concept identification (4.0\%). The 49 unique answers cover diagnostic labels, dermoscopic features, management recommendations, and imaging modalities. Dataset splits were stratified to ensure no image appears in multiple splits, preventing information leakage between training and evaluation sets.

\noindent\textbf{Clinical Malignancy Classification.} We combined 3,056 images from Fitzpatrick17k\cite{f17k} and Diverse Dermatology Images (DDI)\cite{ddi} datasets for binary benign versus malignant classification. From F17k, we excluded images annotated as "non-neoplastic" in the three-partition label field, retaining only benign and malignant neoplasms. We applied automated quality filtering using a DenseNet121 model trained on 1,700 manually annotated images (80/20 train/validation split, F1=0.85 on validation) to remove images with imaging artifacts from the F17k subset. From DDI, we used the malignant column directly as the prediction target. This combined dataset provides diverse skin tones (Fitzpatrick I-VI) and imaging conditions for evaluating concept discovery in malignancy detection.

\noindent\textbf{HAM-7.} The HAM10000 dataset\cite{ham10000} contains dermoscopic images of pigmented skin lesions collected from two academic medical centers in Austria. We split the dataset into 80\% training (8,012 images), 5\% validation (500 images), and 15\% test (1,503 images), stratified by class. The dataset covers seven diagnostic categories: actinic keratosis, basal cell carcinoma, benign keratosis, dermatofibroma, melanoma, nevus, and vascular lesions. All images were captured using standardized dermoscopic imaging protocols. The test set was used for zero-shot classification evaluation, while the complete train/validation/test splits were used for linear probing.

\noindent\textbf{ISIC20-2.} The ISIC 2020 Challenge dataset\cite{isic2020} provides dermoscopic images from multiple international institutions for binary malignancy classification. We created stratified train/validation/test splits with 23,188/4,969/4,969 images (70/15/15\%) labeled as benign or malignant. The test set was used for zero-shot classification, while all splits were used for linear probing. Images represent diverse acquisition settings and patient populations, making this dataset particularly suitable for evaluating model generalization across imaging conditions and demographics.

\noindent\textbf{ISIC-Intervention.} To evaluate robustness against spurious visual correlations, we constructed three artifact-biased datasets from 70,853 dermoscopic images in the ISIC Archive (2016-2020 challenges)\cite{isic2016, isic2017, isic2018, isic2019, isic2020}. We extracted visual embeddings using a ViT-B/16 vision transformer\cite{vit} pretrained on ImageNet\cite{imagenet}, reduced dimensionality to 80 components via PCA, performed clustering, and manually reviewed clusters to isolate groups containing a single artifact type (hair, ruler markings, or colored pen annotations) or no artifacts. Each cluster underwent image-level manual inspection to ensure only one artifact category was present. We then constructed intentionally biased train/test splits where artifacts correlate with melanoma labels during training but exhibit reversed correlation during testing. The hair artifact dataset contains 471 training images (non-melanoma without hair, melanoma with hair) and 649 test images (non-melanoma with hair, melanoma without hair). The ruler artifact dataset includes 610 training images (non-melanoma without ruler, melanoma with ruler) and 501 test images (non-melanoma with ruler, melanoma without ruler). The pen annotation dataset comprises 477 training images (non-melanoma without pen marks, melanoma with pen marks) and 246 test images (non-melanoma with pen marks, melanoma without pen marks). These intentional correlations enable direct assessment of whether models rely on spurious features versus pathologically meaningful patterns.

\noindent\textbf{PAD-6.} The PAD-UFES-20 dataset\cite{pad} from Brazil contains clinical images of six skin lesion types: basal cell carcinoma (BCC), squamous cell carcinoma (SCC), actinic keratosis (ACK), seborrheic keratosis (SEK), melanoma (MEL), and nevus (NEV). We created stratified train/validation/test splits with 1,493/344/461 images (65/15/20\%). The test set was used for zero-shot classification, while all splits were used for linear probing. Images were captured under different smartphone devices and lighting conditions, representing real-world clinical photography scenarios. The dataset focuses on skin cancer and precancerous lesion detection in primary care settings.

\noindent\textbf{PAD (Multimodal).} This dataset combines the PAD-UFES-20\cite{pad} clinical images with comprehensive metadata. Beyond the six diagnostic categories (BCC, SCC, ACK, SEK, melanoma, nevus), each case includes patient demographics (age, gender), anatomical location (body region), risk factors, morphological features (diameter, elevation), clinical symptoms (itch, pain, bleeding), and temporal changes (growth, color change). Metadata were collected through structured clinical interviews and physical examinations. The dataset uses a 1,838/230/230 train/validation/test split stratified by lesion type.

\noindent\textbf{PH2-2.} The PH2 dataset\cite{ph2} from Pedro Hispano Hospital in Portugal consists of 200 dermoscopic images for binary melanoma classification. Images were acquired using a Tuebinger Mole Analyzer system under standardized conditions and include manual segmentation masks. The dataset contains two classes: melanoma and common nevus.

\noindent\textbf{PUMCH-ISD.} The Peking Union Medical College Hospital Inflammatory Skin Disease dataset\cite{pumch} contains 1,174 patients with paired clinical and dermoscopic images. From the original 1,950 clinical photographs and 7,798 dermoscopic images, we only selected one image of each modality per patient. The dataset focuses on eight inflammatory conditions: psoriasis, atopic dermatitis, contact dermatitis, seborrheic dermatitis, lichen planus, pityriasis rosea, vitiligo, acne vulgaris, rosacea, and morphea. Images were captured during routine clinical examinations at a tertiary referral center in China, providing high-quality examples of inflammatory dermatoses in Chinese populations.

\noindent\textbf{SCIN-20.} We curated this dataset from the Skin Condition Image Network (SCIN)\cite{scin} collected from internet users in the United States. From the original dataset, we excluded cases without diagnostic labels and filtered images with quality issues. We retained the 20 most frequent disease categories: impetigo, drug rash, urticaria, eczema, tinea versicolor, psoriasis, pityriasis rosea, tinea, folliculitis, insect bite, allergic contact dermatitis, herpes simplex, pigmented purpuric eruption, herpes zoster, acne, irritant contact dermatitis, keratosis pilaris, contact dermatitis, acute dermatitis (not otherwise specified), and lichen simplex chronicus. For cases with multiple images, we randomly selected one clinical photograph. Each case includes metadata: age group, sex, race/ethnicity, skin tone, lesion texture, anatomical location, and patient-reported symptoms. The final dataset of 2,140 cases uses a 1,284/428/428 train/validation/test split stratified by disease class.

\noindent\textbf{SD-128\cite{sd198}.} The Skin Disease dataset encompasses 128 diagnostic categories from general dermatology practice across multiple centers. We created stratified train/validation/test splits with 3,933/281/1,405 images (70/5/25\%). The test set was used for zero-shot classification, while all splits were used for linear probing. Images represent diverse imaging conditions and patient demographics. The large number of fine-grained categories challenges models to discriminate between visually similar conditions, making this dataset suitable for evaluating diagnostic breadth across common and uncommon dermatological conditions.

\noindent\textbf{SkinCap.} The SkinCap dataset\cite{skincap} provides 3,989 expert-annotated image-caption pairs spanning 178 skin diseases across diverse skin tones (Fitzpatrick types I-VI). Each image is paired with detailed clinical descriptions. The dataset emphasizes diagnostic diversity and skin tone representation, making it particularly valuable for evaluating model performance across different patient populations. Annotations include disease labels, anatomical locations, and clinical observations.

\noindent\textbf{SkinCap-VQA.} This dataset comprises 4,665 question-answer pairs derived from the Derm1M instruction set\cite{derm1m}, filtered to retain only samples sourced from SkinCap images\cite{skincap}. We removed multiple-choice options from questions to create open-ended queries and excluded answers appearing fewer than five times to focus on common response patterns. The resulting dataset contains 188 unique answers spanning diagnostic labels, anatomical locations, clinical features, and treatment recommendations. This dataset evaluates the model's ability to answer diverse dermatological questions grounded in images representing multiple skin tones.

\noindent\textbf{SNU-134\cite{snu134}.} The Seoul National University dermatology dataset contains clinical photographs spanning 134 common skin disease categories encountered in general dermatology practice. The dataset emphasizes presentations in Asian populations, with 2,101 test images covering conditions ranging from inflammatory disorders to infectious diseases and neoplasms. Images were collected from dermatology clinics. Due to the large number of categories, we do not enumerate all class labels here.

\begin{table}
\scriptsize
\centering
\begin{tabular}{ll|llll}
\toprule
Dataset & Model & W\_F1 & AUROC & BACC & SENS \\
\midrule\midrule
\multirow{6}{*}{\rotatebox[origin=c]{90}{\shortstack{DAFFODIL-5\\(Dermoscopy)}}} & CLIP-L/14 & 0.482 (0.456-0.506)$^{***}$ & 0.888 (0.877-0.898)$^{***}$ & 0.613 (0.593-0.633)$^{***}$ & 0.613 (0.593-0.633)$^{***}$ \\
 & BiomedCLIP & 0.588 (0.564-0.611)$^{***}$ & 0.856 (0.845-0.868)$^{***}$ & 0.595 (0.571-0.619)$^{***}$ & 0.595 (0.571-0.619)$^{***}$ \\
 & DermLIP & 0.726 (0.706-0.746)$^{***}$ & 0.940 (0.933-0.947)$^{***}$ & 0.737 (0.715-0.759)$^{***}$ & 0.737 (0.715-0.759)$^{***}$ \\
 & MAKE & 0.771 (0.751-0.791)$^{***}$ & 0.961 (0.955-0.967)$^{***}$ & 0.787 (0.766-0.807)$^{***}$ & 0.787 (0.766-0.807)$^{***}$ \\
 & MONET & 0.762 (0.744-0.781)$^{***}$ & 0.942 (0.935-0.949)$^{***}$ & 0.737 (0.713-0.760)$^{***}$ & 0.737 (0.713-0.760)$^{***}$ \\
 & DermFM-Zero & \textbf{0.889 (0.875-0.902)} & \textbf{0.987 (0.983-0.989)} & \textbf{0.893 (0.878-0.908)} & \textbf{0.893 (0.878-0.908)} \\
\midrule
\multirow{6}{*}{\rotatebox[origin=c]{90}{\shortstack{HAM-7\\(Dermoscopy)}}} & CLIP-L/14 & 0.304 (0.280-0.328)$^{***}$ & 0.660 (0.634-0.685)$^{***}$ & 0.202 (0.174-0.235)$^{***}$ & 0.202 (0.174-0.235)$^{***}$ \\
 & BiomedCLIP & 0.567 (0.537-0.595)$^{***}$ & 0.691 (0.662-0.717)$^{***}$ & 0.228 (0.194-0.265)$^{***}$ & 0.228 (0.194-0.265)$^{***}$ \\
 & DermLIP & 0.686 (0.660-0.711)$^{***}$ & 0.881 (0.863-0.896)$^{***}$ & 0.512 (0.460-0.566)$^{***}$ & 0.512 (0.460-0.566)$^{***}$ \\
 & MAKE & 0.501 (0.476-0.527)$^{***}$ & 0.890 (0.876-0.901)$^{***}$ & 0.557 (0.503-0.607)$^{***}$ & 0.557 (0.503-0.607)$^{***}$ \\
 & MONET & 0.394 (0.368-0.419)$^{***}$ & 0.790 (0.759-0.819)$^{***}$ & 0.334 (0.287-0.382)$^{***}$ & 0.334 (0.287-0.382)$^{***}$ \\
 & DermFM-Zero & \textbf{0.805 (0.786-0.824)} & \textbf{0.950 (0.937-0.961)} & \textbf{0.744 (0.702-0.783)} & \textbf{0.744 (0.702-0.783)} \\
\midrule
\multirow{6}{*}{\rotatebox[origin=c]{90}{\shortstack{ISIC20-2\\(Dermoscopy)}}} & CLIP-L/14 & 0.946 (0.940-0.953) & 0.478 (0.419-0.535) & 0.489 (0.473-0.510) & 0.034 (0.003-0.076) \\
 & BiomedCLIP & 0.970 (0.964-0.975)$^{***}$ & 0.729 (0.677-0.780)$^{***}$ & 0.528 (0.504-0.560)$^{***}$ & 0.069 (0.021-0.133)$^{***}$ \\
 & DermLIP & 0.762 (0.752-0.773)$^{***}$ & 0.824 (0.777-0.867)$^{***}$ & \textbf{0.748 (0.709-0.783)$^{***}$} & \textbf{0.863 (0.786-0.934)$^{***}$} \\
 & MAKE & 0.766 (0.756-0.776)$^{***}$ & 0.813 (0.761-0.863)$^{***}$ & 0.722 (0.678-0.763)$^{***}$ & 0.804 (0.721-0.883)$^{***}$ \\
 & MONET & \textbf{0.970 (0.964-0.975)$^{***}$} & 0.694 (0.642-0.748)$^{***}$ & 0.518 (0.499-0.541)$^{***}$ & 0.046 (0.011-0.093)$^{***}$ \\
 & DermFM-Zero & 0.953 (0.948-0.958) & \textbf{0.872 (0.831-0.908)} & 0.745 (0.693-0.799) & 0.548 (0.444-0.657) \\
\midrule
\multirow{6}{*}{\rotatebox[origin=c]{90}{\shortstack{PAD-6\\(Clinical)}}} & CLIP-L/14 & 0.416 (0.369-0.462)$^{***}$ & 0.726 (0.697-0.755)$^{***}$ & 0.384 (0.317-0.443)$^{***}$ & 0.384 (0.317-0.443)$^{***}$ \\
 & BiomedCLIP & 0.474 (0.424-0.519)$^{***}$ & 0.776 (0.745-0.806)$^{***}$ & 0.441 (0.368-0.509)$^{***}$ & 0.441 (0.368-0.509)$^{***}$ \\
 & DermLIP & 0.597 (0.554-0.639)$^{**}$ & 0.870 (0.840-0.895)$^{**}$ & 0.506 (0.422-0.579)$^{**}$ & 0.506 (0.422-0.579)$^{**}$ \\
 & MAKE & 0.597 (0.553-0.641)$^{***}$ & 0.893 (0.873-0.912)$^{***}$ & 0.608 (0.566-0.652)$^{***}$ & 0.608 (0.566-0.652)$^{***}$ \\
 & MONET & 0.445 (0.400-0.495)$^{***}$ & 0.832 (0.810-0.854)$^{***}$ & 0.446 (0.373-0.524)$^{***}$ & 0.446 (0.373-0.524)$^{***}$ \\
 & DermFM-Zero & \textbf{0.709 (0.668-0.750)} & \textbf{0.932 (0.915-0.947)} & \textbf{0.743 (0.697-0.787)} & \textbf{0.743 (0.697-0.787)} \\
\midrule
\multirow{6}{*}{\rotatebox[origin=c]{90}{\shortstack{PH2-2\\(Dermoscopy)}}} & CLIP-L/14 & 0.696 (0.626-0.771)$^{*}$ & 0.384 (0.283-0.485)$^{*}$ & 0.480 (0.453-0.512)$^{*}$ & 0.024 (0.002-0.081)$^{*}$ \\
 & BiomedCLIP & 0.795 (0.725-0.862) & 0.843 (0.760-0.909) & 0.606 (0.544-0.678) & 0.224 (0.100-0.365) \\
 & DermLIP & 0.607 (0.540-0.669)$^{***}$ & 0.829 (0.744-0.902)$^{***}$ & 0.684 (0.613-0.749)$^{***}$ & 0.875 (0.763-0.974)$^{***}$ \\
 & MAKE & \textbf{0.882 (0.837-0.923)} & 0.911 (0.841-0.969) & 0.858 (0.792-0.920) & 0.828 (0.700-0.932) \\
 & MONET & 0.815 (0.746-0.876) & 0.838 (0.754-0.908) & 0.646 (0.575-0.722) & 0.325 (0.184-0.476) \\
 & DermFM-Zero & 0.857 (0.809-0.902) & \textbf{0.929 (0.866-0.975)} & \textbf{0.875 (0.818-0.921)} & \textbf{0.924 (0.833-1.000)} \\
\midrule
\multirow{5}{*}{\rotatebox[origin=c]{90}{\shortstack{SD-128\\(Clinical)}}} & CLIP-L/14 & 0.083 (0.068-0.099)$^{***}$ & 0.772 (0.077-0.829)$^{***}$ & 0.121 (0.107-0.134)$^{***}$ & 0.121 (0.107-0.134)$^{***}$ \\
 & BiomedCLIP & 0.098 (0.083-0.115)$^{***}$ & 0.761 (0.076-0.851)$^{***}$ & 0.112 (0.097-0.129)$^{***}$ & 0.112 (0.097-0.129)$^{***}$ \\
 & DermLIP & 0.255 (0.233-0.278)$^{***}$ & 0.871 (0.087-0.944)$^{***}$ & 0.275 (0.257-0.295)$^{***}$ & 0.275 (0.256-0.294)$^{***}$ \\
 & MONET & 0.172 (0.152-0.193)$^{***}$ & 0.822 (0.082-0.909)$^{***}$ & 0.211 (0.191-0.231)$^{***}$ & 0.211 (0.191-0.231)$^{***}$ \\
 & DermFM-Zero & \textbf{0.473 (0.446-0.502)} & \textbf{0.895 (0.089-0.974)} & \textbf{0.498 (0.475-0.522)} & \textbf{0.498 (0.474-0.522)} \\
\midrule
\multirow{6}{*}{\rotatebox[origin=c]{90}{\shortstack{SNU-134\\(Clinical)}}} & CLIP-L/14 & 0.061 (0.051-0.071)$^{***}$ & 0.783 (0.775-0.793)$^{***}$ & 0.095 (0.084-0.106)$^{***}$ & 0.095 (0.084-0.106)$^{***}$ \\
 & BiomedCLIP & 0.083 (0.071-0.097)$^{***}$ & 0.823 (0.815-0.831)$^{***}$ & 0.096 (0.084-0.108)$^{***}$ & 0.096 (0.084-0.108)$^{***}$ \\
 & DermLIP & 0.227 (0.208-0.244)$^{***}$ & 0.938 (0.933-0.944)$^{***}$ & 0.243 (0.226-0.260)$^{***}$ & 0.243 (0.226-0.260)$^{***}$ \\
 & MAKE & 0.310 (0.290-0.331)$^{***}$ & 0.949 (0.945-0.954)$^{***}$ & 0.322 (0.303-0.340)$^{***}$ & 0.322 (0.303-0.340)$^{***}$ \\
 & MONET & 0.113 (0.099-0.127)$^{***}$ & 0.884 (0.878-0.892)$^{***}$ & 0.150 (0.136-0.164)$^{***}$ & 0.150 (0.136-0.164)$^{***}$ \\
 & DermFM-Zero & \textbf{0.422 (0.400-0.446)} & \textbf{0.960 (0.958-0.965)} & \textbf{0.452 (0.431-0.472)} & \textbf{0.452 (0.431-0.472)} \\
\bottomrule
\end{tabular}
\caption{\textbf{Zero-shot diagnostic performance across dermatology datasets.} Models include CLIP-L/14, BiomedCLIP, DermLIP, MAKE, MONET, and DermFM-Zero. Performance is reported using Weighted F1 score (W\_F1), Area Under the Receiver Operating Characteristic curve (AUROC), Balanced Accuracy (BACC), and Sensitivity (SENS). Further details on the experimental setup, datasets, and metrics are provided in \textbf{Methods}. Best-performing model for each metric and dataset is bolded and highlighted. 95\% CI is included in parentheses. Significance levels for comparisons with the best model: $^{*}p<0.05$, $^{**}p<0.01$, $^{***}p<0.001$.}
\label{tab:zero-shot}
\end{table}

\begin{table}[h]
\scriptsize
\centering
\begin{tabular}{lllll}
\toprule
Model name & Recall@5 & Recall@10 & Recall@50 & Mean Recall \\
\midrule
CLIP-L/14 & 0.054 (0.049, 0.058)$^{***}$ & 0.079 (0.074, 0.084)$^{***}$ & 0.179 (0.172, 0.186)$^{***}$ & 0.104 (0.098, 0.109) \\
BiomedCLIP & 0.110 (0.104, 0.117)$^{***}$ & 0.151 (0.144, 0.158)$^{***}$ & 0.275 (0.266, 0.283)$^{***}$ & 0.179 (0.171, 0.186) \\
MONET & 0.089 (0.084, 0.095)$^{***}$ & 0.127 (0.120, 0.134)$^{***}$ & 0.260 (0.251, 0.269)$^{***}$ & 0.159 (0.152, 0.166) \\
DermFM-Zero & \textbf{0.339 (0.330, 0.347)} & \textbf{0.425 (0.416, 0.434)} & \textbf{0.598 (0.588, 0.607)} & \textbf{0.454 (0.445, 0.463)} \\
\bottomrule
\end{tabular}
\caption{\textbf{Zero-shot text-to-image retrieval performance for Derm1M validation dataset} ($n = $ 9,806) in terms of Recall@$K$ for $K \in \{1, 5, 10\}$ and mean recall over $K$. The best performing model for each metric is bolded. 95\% CI is included in parentheses.}
\label{tab:text_to_image_derm1m_validation}
\end{table}

\begin{table}[h]
\scriptsize
\centering
\begin{tabular}{lllll}
\toprule
Model name & Recall@5 & Recall@10 & Recall@50 & Mean Recall \\
\midrule
CLIP-L/14 & 0.069 (0.064, 0.074)$^{***}$ & 0.098 (0.092, 0.104)$^{***}$ & 0.200 (0.192, 0.208) & 0.122 (0.116, 0.129) \\
BiomedCLIP & 0.123 (0.117, 0.129)$^{***}$ & 0.164 (0.157, 0.171)$^{***}$ & 0.278 (0.270, 0.287) & 0.188 (0.181, 0.196) \\
MONET & 0.101 (0.095, 0.106)$^{***}$ & 0.142 (0.136, 0.149)$^{***}$ & 0.271 (0.264, 0.279) & 0.171 (0.165, 0.178) \\
DermFM-Zero & \textbf{0.343 (0.334, 0.352)} & \textbf{0.427 (0.418, 0.437)} & \textbf{0.601 (0.591, 0.611)} & \textbf{0.457 (0.448, 0.467)} \\
\bottomrule
\end{tabular}
\caption{\textbf{Zero-shot image-to-text retrieval performance for Derm1M validation dataset} ($n = $ 9,806) in terms of Recall@$K$ for $K \in \{5, 10, 50\}$ and mean recall over $K$. The best performing model for each metric is bolded. 95\% CI is included in parentheses.}
\label{tab:image_to_text_derm1m_validation}
\end{table}

\begin{table}[h]
\scriptsize
\centering
\begin{tabular}{lllll}
\toprule
Model name & Recall@5 & Recall@10 & Recall@50 & Mean Recall \\
\midrule
CLIP-L/14 & 0.056 (0.049, 0.062)$^{***}$ & 0.090 (0.082, 0.098)$^{***}$ & 0.234 (0.222, 0.247)$^{***}$ & 0.127 (0.117, 0.136) \\
BiomedCLIP & 0.073 (0.066, 0.081)$^{***}$ & 0.122 (0.113, 0.133)$^{***}$ & 0.328 (0.314, 0.342)$^{***}$ & 0.175 (0.164, 0.185) \\
MONET & 0.096 (0.088, 0.104)$^{***}$ & 0.153 (0.142, 0.163)$^{***}$ & 0.360 (0.345, 0.374)$^{***}$ & 0.203 (0.192, 0.214) \\
DermFM-Zero & \textbf{0.178 (0.168, 0.189)} & \textbf{0.283 (0.271, 0.296)} & \textbf{0.586 (0.571, 0.601)} & \textbf{0.349 (0.337, 0.362)} \\
\bottomrule
\end{tabular}
\caption{\textbf{Zero-shot text-to-image retrieval performance for SkinCap} ($n = $ 4,000) in terms of Recall@$K$ for $K \in \{1, 5, 10\}$ and mean recall over $K$. The best performing model for each metric is bolded. 95\% CI is included in parentheses.}
\label{tab:text_to_image_skincap}
\end{table}

\begin{table}[h]
\scriptsize
\centering
\begin{tabular}{lllll}
\toprule
Model name & Recall@5 & Recall@10 & Recall@50 & Mean Recall \\
\midrule
CLIP-L/14 & 0.078 (0.071, 0.086)$^{***}$ & 0.127 (0.117, 0.136)$^{***}$ & 0.317 (0.303, 0.332) & 0.174 (0.164, 0.185) \\
BiomedCLIP & 0.082 (0.074, 0.090)$^{***}$ & 0.134 (0.124, 0.144)$^{***}$ & 0.344 (0.330, 0.358) & 0.187 (0.176, 0.197) \\
MONET & 0.100 (0.091, 0.108)$^{***}$ & 0.161 (0.150, 0.171)$^{***}$ & 0.385 (0.371, 0.400) & 0.215 (0.204, 0.226) \\
DermFM-Zero & \textbf{0.186 (0.175, 0.197)} & \textbf{0.299 (0.286, 0.313)} & \textbf{0.623 (0.608, 0.637)} & \textbf{0.369 (0.357, 0.382)} \\
\bottomrule
\end{tabular}
\caption{\textbf{Zero-shot image-to-text retrieval performance for SkinCap} ($n = $ 4,000) in terms of Recall@$K$ for $K \in \{5, 10, 50\}$ and mean recall over $K$. The best performing model for each metric is bolded. 95\% CI is included in parentheses.}
\label{tab:image_to_text_skincap}
\end{table}

\begin{table}[!t]
\footnotesize\centering
\setlength{\tabcolsep}{4pt}
\begin{tabular}{llllll}
\toprule
Percent & Model & W\_F1 & AUROC & BACC & M\_F1 \\
\midrule\midrule
\multirow{6}{*}{10\%} & BiomedCLIP & 0.610 (0.581-0.640)$^{***}$ & 0.882 (0.866-0.897)$^{***}$ & 0.194 (0.180-0.209)$^{***}$ & 0.199 (0.178-0.220)$^{***}$ \\
 & OpenCLIP & 0.776 (0.755-0.798)$^{***}$ & 0.927 (0.916-0.938)$^{***}$ & 0.515 (0.464-0.569)$^{***}$ & 0.553 (0.493-0.610)$^{***}$ \\
 & MONET & 0.796 (0.773-0.816)$^{***}$ & 0.939 (0.928-0.949)$^{***}$ & 0.523 (0.479-0.566)$^{***}$ & 0.559 (0.511-0.608)$^{***}$ \\
 & DermLIP & 0.809 (0.788-0.829)$^{***}$ & 0.956 (0.949-0.963)$^{***}$ & 0.583 (0.528-0.639)$^{***}$ & 0.623 (0.568-0.675)$^{***}$ \\
 & PanDerm & 0.788 (0.765-0.810)$^{***}$ & 0.941 (0.928-0.952)$^{***}$ & 0.556 (0.504-0.606)$^{***}$ & 0.585 (0.527-0.640)$^{***}$ \\
 & DermFM-Zero & \textbf{0.842 (0.822-0.862)} & \textbf{0.966 (0.959-0.972)} & \textbf{0.718 (0.669-0.766)} & \textbf{0.737 (0.693-0.778)} \\
\midrule
\multirow{6}{*}{30\%} & BiomedCLIP & 0.595 (0.566-0.627)$^{***}$ & 0.880 (0.861-0.896)$^{***}$ & 0.182 (0.171-0.197)$^{***}$ & 0.183 (0.165-0.204)$^{***}$ \\
 & OpenCLIP & 0.801 (0.779-0.822)$^{***}$ & 0.948 (0.940-0.956)$^{***}$ & 0.611 (0.552-0.670)$^{***}$ & 0.641 (0.584-0.692)$^{***}$ \\
 & MONET & 0.827 (0.805-0.848)$^{***}$ & 0.959 (0.951-0.966)$^{***}$ & 0.605 (0.548-0.659)$^{***}$ & 0.654 (0.596-0.707)$^{***}$ \\
 & DermLIP & 0.829 (0.806-0.849)$^{***}$ & 0.969 (0.962-0.975)$^{***}$ & 0.635 (0.584-0.689)$^{***}$ & 0.676 (0.621-0.725)$^{***}$ \\
 & PanDerm & 0.830 (0.809-0.850)$^{***}$ & 0.965 (0.956-0.972)$^{***}$ & 0.662 (0.605-0.716)$^{***}$ & 0.697 (0.644-0.746)$^{***}$ \\
 & DermFM-Zero & \textbf{0.875 (0.859-0.892)} & \textbf{0.981 (0.977-0.984)} & \textbf{0.781 (0.732-0.825)} & \textbf{0.801 (0.760-0.836)} \\
\midrule
\multirow{6}{*}{50\%} & BiomedCLIP & 0.584 (0.552-0.613)$^{***}$ & 0.881 (0.863-0.898)$^{***}$ & 0.174 (0.164-0.187)$^{***}$ & 0.170 (0.153-0.191)$^{***}$ \\
 & OpenCLIP & 0.813 (0.791-0.833)$^{***}$ & 0.960 (0.953-0.967)$^{***}$ & 0.637 (0.587-0.692)$^{***}$ & 0.668 (0.620-0.713)$^{***}$ \\
 & MONET & 0.841 (0.822-0.861)$^{***}$ & 0.964 (0.956-0.972)$^{***}$ & 0.694 (0.643-0.744)$^{***}$ & 0.732 (0.685-0.775)$^{***}$ \\
 & DermLIP & 0.837 (0.816-0.856)$^{***}$ & 0.972 (0.966-0.977)$^{***}$ & 0.679 (0.631-0.731)$^{***}$ & 0.716 (0.671-0.760)$^{***}$ \\
 & PanDerm & 0.855 (0.837-0.872)$^{***}$ & 0.975 (0.969-0.981)$^{***}$ & 0.723 (0.671-0.771)$^{***}$ & 0.755 (0.709-0.794)$^{***}$ \\
 & DermFM-Zero & \textbf{0.891 (0.874-0.908)} & \textbf{0.985 (0.981-0.988)} & \textbf{0.817 (0.771-0.858)} & \textbf{0.820 (0.781-0.854)} \\
\midrule
\multirow{6}{*}{100\%} & BiomedCLIP & 0.586 (0.557-0.618)$^{***}$ & 0.880 (0.861-0.896)$^{***}$ & 0.176 (0.164-0.189)$^{***}$ & 0.173 (0.154-0.194)$^{***}$ \\
 & OpenCLIP & 0.841 (0.823-0.860)$^{***}$ & 0.968 (0.962-0.973)$^{***}$ & 0.668 (0.616-0.721)$^{***}$ & 0.704 (0.655-0.753)$^{***}$ \\
 & MONET & 0.845 (0.825-0.864)$^{***}$ & 0.970 (0.963-0.975)$^{***}$ & 0.660 (0.607-0.713)$^{***}$ & 0.709 (0.661-0.755)$^{***}$ \\
 & DermLIP & 0.844 (0.825-0.862)$^{***}$ & 0.973 (0.969-0.978)$^{***}$ & 0.673 (0.619-0.727)$^{***}$ & 0.712 (0.661-0.759)$^{***}$ \\
 & PanDerm & 0.881 (0.864-0.898)$^{***}$ & 0.981 (0.977-0.986)$^{***}$ & 0.782 (0.733-0.826)$^{***}$ & 0.816 (0.775-0.849)$^{***}$ \\
 & DermFM-Zero & \textbf{0.894 (0.878-0.910)} & \textbf{0.989 (0.986-0.991)} & \textbf{0.832 (0.789-0.873)} & \textbf{0.827 (0.791-0.860)} \\
\bottomrule
\end{tabular}
\caption{\textbf{Label efficiency generalization (linear probing) performance for dermoscopic image-based skin lesion classification based on HAM10000 dataset.} 
Metrics: W\_F1 (Weighted F1), AUROC, BACC (Balanced Accuracy), M\_F1 (Macro F1). 
Best performance for each setting is bolded. 95\% CI in parentheses. 
$^{*}p<0.05$, $^{**}p<0.01$, $^{***}p<0.001$ compared to DermFM-Zero.}
\label{tab:label_efficiency_ham}
\end{table}


\begin{table}[!t]
\footnotesize\centering
\setlength{\tabcolsep}{4pt}
\begin{tabular}{llllll}
\toprule
Percent & Model & W\_F1 & AUROC & BACC & M\_F1 \\
\midrule\midrule
\multirow{6}{*}{10\%} & BiomedCLIP & 0.611 (0.562-0.659)$^{***}$ & 0.871 (0.850-0.893)$^{***}$ & 0.435 (0.394-0.476)$^{***}$ & 0.434 (0.392-0.473)$^{***}$ \\
 & OpenCLIP & 0.606 (0.560-0.651)$^{***}$ & 0.855 (0.828-0.877)$^{***}$ & 0.475 (0.409-0.546)$^{***}$ & 0.499 (0.421-0.574)$^{***}$ \\
 & MONET & 0.641 (0.596-0.684)$^{***}$ & 0.880 (0.857-0.903)$^{***}$ & 0.560 (0.483-0.639)$^{***}$ & \textbf{0.597 (0.517-0.668)} \\
 & DermLIP & 0.659 (0.616-0.700)$^{***}$ & \textbf{0.909 (0.891-0.928)} & \textbf{0.586 (0.512-0.666)} & 0.587 (0.514-0.655) \\
 & PanDerm & 0.644 (0.603-0.689)$^{***}$ & 0.883 (0.864-0.902)$^{***}$ & 0.552 (0.478-0.637)$^{***}$ & 0.544 (0.477-0.612)$^{***}$ \\
 & DermFM-Zero & \textbf{0.678 (0.632-0.721)} & 0.896 (0.876-0.917) & 0.568 (0.493-0.645) & 0.574 (0.490-0.646) \\
\midrule
\multirow{6}{*}{30\%} & BiomedCLIP & 0.618 (0.567-0.668)$^{***}$ & 0.884 (0.865-0.905)$^{***}$ & 0.436 (0.395-0.481)$^{***}$ & 0.436 (0.393-0.485)$^{***}$ \\
 & OpenCLIP & 0.690 (0.646-0.734)$^{***}$ & 0.902 (0.884-0.919)$^{***}$ & 0.576 (0.502-0.653)$^{***}$ & 0.594 (0.510-0.670)$^{***}$ \\
 & MONET & 0.683 (0.641-0.727)$^{***}$ & 0.894 (0.871-0.917)$^{***}$ & 0.557 (0.487-0.626)$^{***}$ & 0.574 (0.494-0.651)$^{***}$ \\
 & DermLIP & 0.711 (0.669-0.750)$^{***}$ & 0.922 (0.904-0.939)$^{***}$ & \textbf{0.642 (0.567-0.717)} & \textbf{0.647 (0.568-0.714)} \\
 & PanDerm & \textbf{0.722 (0.682-0.761)} & 0.912 (0.894-0.929)$^{***}$ & 0.638 (0.560-0.716) & 0.642 (0.562-0.712) \\
 & DermFM-Zero & 0.720 (0.675-0.764) & \textbf{0.928 (0.911-0.945)} & 0.588 (0.519-0.663) & 0.602 (0.520-0.678) \\
\midrule
\multirow{6}{*}{50\%} & BiomedCLIP & 0.637 (0.591-0.683)$^{***}$ & 0.890 (0.869-0.910)$^{***}$ & 0.448 (0.410-0.484)$^{***}$ & 0.448 (0.411-0.482)$^{***}$ \\
 & OpenCLIP & 0.677 (0.634-0.721)$^{***}$ & 0.901 (0.885-0.921)$^{***}$ & 0.597 (0.519-0.670)$^{***}$ & 0.596 (0.516-0.661)$^{***}$ \\
 & MONET & 0.703 (0.662-0.741)$^{***}$ & 0.906 (0.884-0.927)$^{***}$ & 0.606 (0.529-0.680)$^{***}$ & 0.627 (0.541-0.696)$^{***}$ \\
 & DermLIP & 0.743 (0.704-0.781)$^{***}$ & 0.925 (0.908-0.944)$^{***}$ & \textbf{0.695 (0.622-0.757)} & \textbf{0.704 (0.639-0.758)} \\
 & PanDerm & 0.719 (0.680-0.759)$^{***}$ & 0.925 (0.908-0.940)$^{***}$ & 0.679 (0.602-0.749) & 0.683 (0.613-0.740) \\
 & DermFM-Zero & \textbf{0.748 (0.709-0.789)} & \textbf{0.929 (0.912-0.947)} & 0.639 (0.572-0.716) & 0.654 (0.573-0.739) \\
\midrule
\multirow{6}{*}{100\%} & BiomedCLIP & 0.640 (0.593-0.687)$^{***}$ & 0.892 (0.874-0.911)$^{***}$ & 0.447 (0.407-0.487)$^{***}$ & 0.446 (0.409-0.482)$^{***}$ \\
 & OpenCLIP & 0.722 (0.682-0.763)$^{***}$ & 0.926 (0.909-0.943)$^{***}$ & 0.640 (0.562-0.717)$^{***}$ & 0.659 (0.584-0.723)$^{***}$ \\
 & MONET & 0.725 (0.682-0.766)$^{***}$ & 0.916 (0.898-0.934)$^{***}$ & 0.646 (0.569-0.721)$^{***}$ & 0.667 (0.588-0.735)$^{***}$ \\
 & DermLIP & 0.748 (0.704-0.789)$^{***}$ & 0.936 (0.918-0.952) & 0.721 (0.679-0.763)$^{***}$ & 0.712 (0.658-0.765)$^{***}$ \\
 & PanDerm & \textbf{0.758 (0.720-0.800)} & \textbf{0.942 (0.929-0.959)} & 0.710 (0.632-0.778)$^{***}$ & 0.716 (0.644-0.775)$^{***}$ \\
 & DermFM-Zero & 0.758 (0.719-0.797) & 0.935 (0.917-0.952) & \textbf{0.740 (0.675-0.798)} & \textbf{0.733 (0.675-0.786)} \\
\bottomrule
\end{tabular}
\caption{\textbf{Label efficiency generalization (linear probing) performance for clinical photo-based skin lesion classification based on PAD-UFES-20 dataset.} 
Metrics: W\_F1 (Weighted F1), AUROC, BACC (Balanced Accuracy), M\_F1 (Macro F1). 
Best performance for each setting is bolded. 95\% CI in parentheses. 
$^{*}p<0.05$, $^{**}p<0.01$, $^{***}p<0.001$ compared to DermFM-Zero.}
\label{tab:label_efficiency_pad}
\end{table}


\begin{table}[!t]
\footnotesize\centering
\setlength{\tabcolsep}{4pt}
\begin{tabular}{llllll}
\toprule
Percent & Model & W\_F1 & AUROC & BACC & M\_F1 \\
\midrule\midrule
\multirow{6}{*}{10\%} & BiomedCLIP & 0.237 (0.215-0.260)$^{***}$ & 0.786 (0.000-0.904)$^{***}$ & 0.226 (0.205-0.246)$^{***}$ & 0.209 (0.189-0.230)$^{***}$ \\
 & OpenCLIP & 0.296 (0.272-0.320)$^{***}$ & 0.790 (0.000-0.923)$^{***}$ & 0.283 (0.262-0.305)$^{***}$ & 0.266 (0.244-0.286)$^{***}$ \\
 & MONET & 0.259 (0.235-0.284)$^{***}$ & 0.796 (0.000-0.912)$^{***}$ & 0.239 (0.219-0.260)$^{***}$ & 0.225 (0.205-0.246)$^{***}$ \\
 & DermLIP & 0.332 (0.305-0.357)$^{***}$ & \textbf{0.820 (0.000-0.940)} & 0.317 (0.295-0.339)$^{***}$ & 0.294 (0.272-0.317)$^{***}$ \\
 & PanDerm & 0.328 (0.303-0.353)$^{***}$ & 0.793 (0.000-0.931) & 0.315 (0.291-0.339)$^{***}$ & 0.299 (0.274-0.323)$^{***}$ \\
 & DermFM-Zero & \textbf{0.383 (0.357-0.409)} & 0.780 (0.000-0.928) & \textbf{0.365 (0.341-0.387)} & \textbf{0.347 (0.325-0.369)} \\
\midrule
\multirow{6}{*}{30\%} & BiomedCLIP & 0.349 (0.323-0.376)$^{***}$ & 0.811 (0.000-0.949)$^{***}$ & 0.343 (0.318-0.370)$^{***}$ & 0.333 (0.309-0.360)$^{***}$ \\
 & OpenCLIP & 0.471 (0.443-0.498)$^{***}$ & 0.836 (0.000-0.970)$^{***}$ & 0.468 (0.442-0.494)$^{***}$ & 0.455 (0.429-0.479)$^{***}$ \\
 & MONET & 0.439 (0.412-0.464)$^{***}$ & 0.839 (0.000-0.963)$^{***}$ & 0.426 (0.400-0.453)$^{***}$ & 0.416 (0.388-0.442)$^{***}$ \\
 & DermLIP & 0.512 (0.485-0.540)$^{***}$ & \textbf{0.848 (0.000-0.976)$^{***}$} & 0.506 (0.481-0.533)$^{***}$ & 0.489 (0.463-0.516)$^{***}$ \\
 & PanDerm & 0.523 (0.495-0.549)$^{***}$ & 0.829 (0.000-0.976)$^{***}$ & 0.524 (0.499-0.552)$^{***}$ & 0.509 (0.484-0.535)$^{***}$ \\
 & DermFM-Zero & \textbf{0.578 (0.551-0.605)} & 0.837 (0.000-0.978) & \textbf{0.569 (0.544-0.594)} & \textbf{0.558 (0.532-0.583)} \\
\midrule
\multirow{6}{*}{50\%} & BiomedCLIP & 0.384 (0.359-0.410)$^{***}$ & 0.828 (0.000-0.958)$^{***}$ & 0.374 (0.349-0.401)$^{***}$ & 0.363 (0.338-0.389)$^{***}$ \\
 & OpenCLIP & 0.536 (0.510-0.564)$^{***}$ & 0.836 (0.000-0.979)$^{***}$ & 0.533 (0.506-0.561)$^{***}$ & 0.517 (0.491-0.544)$^{***}$ \\
 & MONET & 0.517 (0.490-0.544)$^{***}$ & 0.848 (0.000-0.974)$^{***}$ & 0.510 (0.484-0.535)$^{***}$ & 0.495 (0.468-0.520)$^{***}$ \\
 & DermLIP & 0.572 (0.545-0.600)$^{***}$ & 0.866 (0.000-0.985)$^{**}$ & 0.575 (0.548-0.602)$^{***}$ & 0.558 (0.529-0.586)$^{***}$ \\
 & PanDerm & 0.578 (0.550-0.607)$^{***}$ & \textbf{0.868 (0.000-0.985)$^{***}$} & 0.586 (0.560-0.614)$^{***}$ & 0.568 (0.540-0.597)$^{***}$ \\
 & DermFM-Zero & \textbf{0.647 (0.621-0.672)} & 0.851 (0.000-0.985) & \textbf{0.652 (0.627-0.676)} & \textbf{0.637 (0.611-0.662)} \\
\midrule
\multirow{6}{*}{100\%} & BiomedCLIP & 0.431 (0.405-0.456)$^{***}$ & 0.843 (0.000-0.968)$^{***}$ & 0.436 (0.408-0.463)$^{***}$ & 0.422 (0.395-0.448)$^{***}$ \\
 & OpenCLIP & 0.609 (0.582-0.634)$^{***}$ & 0.849 (0.000-0.986)$^{***}$ & 0.614 (0.588-0.640)$^{***}$ & 0.600 (0.575-0.625)$^{***}$ \\
 & MONET & 0.607 (0.582-0.634)$^{***}$ & 0.861 (0.000-0.984)$^{***}$ & 0.611 (0.586-0.637)$^{***}$ & 0.598 (0.572-0.624)$^{***}$ \\
 & DermLIP & 0.645 (0.621-0.670)$^{***}$ & 0.851 (0.000-0.990)$^{***}$ & 0.653 (0.627-0.677)$^{***}$ & 0.637 (0.612-0.663)$^{***}$ \\
 & PanDerm & 0.679 (0.654-0.705)$^{***}$ & 0.847 (0.000-0.990)$^{***}$ & 0.685 (0.659-0.712)$^{***}$ & 0.671 (0.646-0.697)$^{***}$ \\
 & DermFM-Zero & \textbf{0.711 (0.685-0.737)} & \textbf{0.869 (0.000-0.991)} & \textbf{0.714 (0.689-0.739)} & \textbf{0.706 (0.680-0.731)} \\
\bottomrule
\end{tabular}
\caption{\textbf{Label efficiency generalization (linear probing) performance for skin condition diagnosis based on SD-128 dataset.} 
Metrics: W\_F1 (Weighted F1), AUROC, BACC (Balanced Accuracy), M\_F1 (Macro F1). 
Best performance for each setting is bolded. 95\% CI in parentheses. 
$^{*}p<0.05$, $^{**}p<0.01$, $^{***}p<0.001$ compared to DermFM-Zero.}
\label{tab:label_efficiency_sd128}
\end{table}


\begin{table}[!t]
\footnotesize\centering
\setlength{\tabcolsep}{4pt}
\begin{tabular}{llllll}
\toprule
Percent & Model & W\_F1 & AUROC & BACC & M\_F1 \\
\midrule\midrule
\multirow{6}{*}{10\%} & BiomedCLIP & 0.974 (0.968-0.980)$^{***}$ & 0.431 (0.411-0.456)$^{***}$ & 0.500 (0.500-0.500)$^{***}$ & 0.496 (0.495-0.497)$^{***}$ \\
 & OpenCLIP & 0.974 (0.969-0.979)$^{***}$ & 0.782 (0.736-0.822)$^{***}$ & 0.500 (0.500-0.500)$^{***}$ & 0.496 (0.495-0.496)$^{***}$ \\
 & MONET & 0.974 (0.968-0.979)$^{***}$ & 0.804 (0.766-0.838)$^{***}$ & 0.500 (0.500-0.500)$^{***}$ & 0.496 (0.495-0.496)$^{***}$ \\
 & DermLIP & 0.974 (0.968-0.979)$^{***}$ & 0.806 (0.767-0.843)$^{***}$ & 0.500 (0.500-0.500)$^{***}$ & 0.496 (0.495-0.496)$^{***}$ \\
 & PanDerm & 0.974 (0.968-0.979)$^{***}$ & 0.841 (0.803-0.876)$^{***}$ & 0.500 (0.500-0.500)$^{***}$ & 0.496 (0.495-0.496)$^{***}$ \\
 & DermFM-Zero & \textbf{0.977 (0.972-0.982)} & \textbf{0.869 (0.829-0.904)} & \textbf{0.556 (0.524-0.592)} & \textbf{0.586 (0.538-0.637)} \\
\midrule
\multirow{6}{*}{30\%} & BiomedCLIP & 0.974 (0.968-0.979)$^{***}$ & 0.500 (0.500-0.500)$^{***}$ & 0.500 (0.500-0.500)$^{***}$ & 0.496 (0.495-0.496)$^{***}$ \\
 & OpenCLIP & 0.974 (0.969-0.979)$^{***}$ & 0.816 (0.780-0.850)$^{***}$ & 0.500 (0.500-0.500)$^{***}$ & 0.496 (0.495-0.496)$^{***}$ \\
 & MONET & 0.974 (0.968-0.979)$^{***}$ & 0.838 (0.804-0.872)$^{***}$ & 0.500 (0.500-0.500)$^{***}$ & 0.496 (0.495-0.496)$^{***}$ \\
 & DermLIP & 0.974 (0.968-0.979)$^{***}$ & 0.796 (0.757-0.837)$^{***}$ & 0.500 (0.500-0.500)$^{***}$ & 0.496 (0.495-0.496)$^{***}$ \\
 & PanDerm & 0.974 (0.968-0.979)$^{***}$ & 0.877 (0.847-0.903)$^{***}$ & 0.500 (0.500-0.500)$^{***}$ & 0.496 (0.495-0.496)$^{***}$ \\
 & DermFM-Zero & \textbf{0.976 (0.971-0.982)} & \textbf{0.924 (0.899-0.948)} & \textbf{0.540 (0.516-0.573)} & \textbf{0.566 (0.524-0.615)} \\
\midrule
\multirow{6}{*}{50\%} & BiomedCLIP & 0.974 (0.969-0.979)$^{***}$ & 0.489 (0.481-0.503)$^{***}$ & 0.500 (0.500-0.500)$^{***}$ & 0.496 (0.495-0.496)$^{***}$ \\
 & OpenCLIP & 0.974 (0.968-0.979)$^{***}$ & 0.826 (0.792-0.856)$^{***}$ & 0.500 (0.500-0.500)$^{***}$ & 0.496 (0.495-0.496)$^{***}$ \\
 & MONET & 0.974 (0.969-0.979)$^{***}$ & 0.833 (0.799-0.865)$^{***}$ & 0.500 (0.500-0.500)$^{***}$ & 0.496 (0.495-0.496)$^{***}$ \\
 & DermLIP & 0.974 (0.968-0.979)$^{***}$ & 0.753 (0.708-0.796)$^{***}$ & 0.500 (0.500-0.500)$^{***}$ & 0.496 (0.495-0.496)$^{***}$ \\
 & PanDerm & 0.974 (0.968-0.979)$^{***}$ & 0.888 (0.863-0.911)$^{***}$ & 0.500 (0.500-0.500)$^{***}$ & 0.496 (0.495-0.496)$^{***}$ \\
 & DermFM-Zero & \textbf{0.976 (0.970-0.981)} & \textbf{0.935 (0.914-0.955)} & \textbf{0.528 (0.506-0.556)} & \textbf{0.547 (0.507-0.593)} \\
\midrule
\multirow{6}{*}{100\%} & BiomedCLIP & 0.974 (0.968-0.979)$^{***}$ & 0.368 (0.312-0.420)$^{***}$ & 0.500 (0.500-0.500)$^{***}$ & 0.496 (0.495-0.496)$^{***}$ \\
 & OpenCLIP & 0.974 (0.968-0.979)$^{***}$ & 0.839 (0.805-0.868)$^{***}$ & 0.500 (0.500-0.500)$^{***}$ & 0.496 (0.495-0.496)$^{***}$ \\
 & MONET & 0.974 (0.968-0.979)$^{***}$ & 0.846 (0.814-0.879)$^{***}$ & 0.500 (0.500-0.500)$^{***}$ & 0.496 (0.495-0.496)$^{***}$ \\
 & DermLIP & 0.974 (0.968-0.979)$^{***}$ & 0.873 (0.841-0.902)$^{***}$ & 0.500 (0.500-0.500)$^{***}$ & 0.496 (0.495-0.496)$^{***}$ \\
 & PanDerm & 0.974 (0.968-0.979)$^{***}$ & 0.892 (0.866-0.916)$^{***}$ & 0.500 (0.500-0.500)$^{***}$ & 0.496 (0.495-0.496)$^{***}$ \\
 & DermFM-Zero & \textbf{0.976 (0.971-0.981)} & \textbf{0.939 (0.917-0.956)} & \textbf{0.534 (0.510-0.564)} & \textbf{0.557 (0.516-0.604)} \\
\bottomrule
\end{tabular}
\caption{\textbf{Label efficiency generalization (linear probing) performance for binary melanoma detection based on ISIC2020 dataset.} 
Metrics: W\_F1 (Weighted F1), AUROC, BACC (Balanced Accuracy), M\_F1 (Macro F1). 
Best performance for each setting is bolded. 95\% CI in parentheses. 
$^{*}p<0.05$, $^{**}p<0.01$, $^{***}p<0.001$ compared to DermFM-Zero.}
\label{tab:label_efficiency_isic2020}
\end{table}

\begin{table}[h]
\footnotesize\centering
\setlength{\tabcolsep}{3pt}
\begin{tabular}{llllll}
\toprule
Metric & Condition & Mean (95\% CI) & Median & IQR & p-value \\
\midrule
\multirow{2}{*}{Diagnostic Accuracy} 
  & Without AI & 2.24 (1.90-2.58) & 2.22 & 0.41 & \multirow{2}{*}{0.0058$^{**}$} \\
  & With AI & \textbf{3.05 (2.67-3.43)} & \textbf{3.08} & \textbf{0.90} & \\
\midrule
\multirow{2}{*}{Management Quality}
  & Without AI & 2.13 (1.83-2.43) & 2.02 & 0.80 & \multirow{2}{*}{0.0142$^{*}$} \\
  & With AI & \textbf{2.35 (1.98-2.72)} & \textbf{2.37} & \textbf{0.57} & \\
\midrule
\multirow{2}{*}{Top-3 Utility (\%)}
  & Without AI & 25.60 (16.59-34.60) & 26.67 & 13.51 & \multirow{2}{*}{0.0038$^{**}$} \\
  & With AI & \textbf{48.24 (38.03-58.46)} & \textbf{45.00} & \textbf{22.29} & \\
\midrule
\multirow{2}{*}{Harm Rate (\%)}
  & Without AI & 40.02 (24.22-55.83) & 43.65 & 37.65 & \multirow{2}{*}{0.0479$^{*}$} \\
  & With AI & 34.10 (19.07-49.13) & 30.95 & 27.71 & \\
\midrule
\multirow{2}{*}{Success Rate (\%)}
  & Without AI & 50.37 (35.01-65.73) & 40.08 & 45.65 & \multirow{2}{*}{0.0177$^{*}$} \\
  & With AI & \textbf{59.23 (43.07-75.39)} & \textbf{60.71} & \textbf{32.50} & \\
\midrule
\multirow{2}{*}{Diagnosis Confidence}
  & Without AI & 2.89 (2.46-3.33) & 3.00 & 0.46 & \multirow{2}{*}{0.0139$^{*}$} \\
  & With AI & \textbf{3.28 (2.86-3.69)} & \textbf{3.10} & \textbf{0.66} & \\
\midrule
\multirow{2}{*}{Management Confidence}
  & Without AI & 2.90 (2.45-3.36) & 3.00 & 0.16 & \multirow{2}{*}{0.0232$^{*}$} \\
  & With AI & \textbf{3.11 (2.65-3.56)} & \textbf{3.00} & \textbf{0.23} & \\
\bottomrule
\end{tabular}
\caption{\textbf{Reader Study 1: Comprehensive analysis of human-AI collaboration performance on skin condition differential diagnosis}. Statistical analysis including mean with 95\% confidence interval, median, and interquartile range (IQR) for all performance metrics. Diagnostic accuracy scored 1-5 (1=potential harm, 5=spot on); Management quality scored 1-4 (1=inadequate/dangerous, 4=perfect); Top-3 utility represents percentage of cases with correct diagnosis in top-3 differentials; Harm rate indicates percentage of inadequate/dangerous management decisions; Success rate represents percentage of adequate or perfect management; Confidence scored 1-5 (1=not confident, 5=very confident). n=12 readers. $^{*}p<0.05$, $^{**}p<0.01$, $^{***}p<0.001$ (Wilcoxon signed-rank test, one-tailed). ns = not significant.}
\label{tab:rs1_comprehensive_analysis}
\end{table}

\begin{table}[h]
\footnotesize\centering
\setlength{\tabcolsep}{4pt}
\begin{tabular}{lll}
\toprule
Group & TODIV Score (\%) & p-value \\
\midrule
Humans (All) & 66.26 (65.66-66.87) & $<$0.0001$^{***}$ \\
General Practitioners & 59.57 (58.27-60.88) & $<$0.0001$^{***}$ \\
Dermatologists & 69.35 (68.76-69.95) & $<$0.0001$^{***}$ \\
Ypsono (AI) & 56.64 (56.56-56.71) & $<$0.0001$^{***}$ \\
\textbf{DermFM-Zero Zero-Shot} & \textbf{71.74 (71.70-71.79)} & \\
\bottomrule
\end{tabular}
\caption{\textbf{Reader Stduy 2A: Performance comparison by group on TODIV cohort}. TODIV scores with 95\% CI in parentheses. P-values from comparison to DermFM-Zero Zero-Shot. $^{*}p<0.05$, $^{**}p<0.01$, $^{***}p<0.001$.}
\label{tab:rs2a_performance_by_group}
\end{table}

\begin{table}[h]
\footnotesize\centering
\setlength{\tabcolsep}{4pt}
\begin{tabular}{lll}
\toprule
Experience Level & TODIV Score (\%) & p-value \\
\midrule
Humans (All) & 66.26 (65.66-66.87) & $<$0.0001$^{***}$ \\
<1 year & 58.51 (57.51-59.50) & $<$0.0001$^{***}$ \\
1-3 years & 68.51 (67.63-69.40) & $<$0.0001$^{***}$ \\
3-10 years & 73.17 (72.39-73.95) &  \\
>10 years & 73.66 (72.46-74.87) &  \\
Ypsono (AI) & 56.64 (56.56-56.71) & $<$0.0001$^{***}$ \\
\textbf{DermFM-Zero Zero-Shot} & \textbf{71.74 (71.70-71.79)} &  \\
\bottomrule
\end{tabular}
\caption{\textbf{Reader Stduy 2A: Performance comparison by experience level on TODIV chort}. TODIV scores with 95\% CI in parentheses. P-values from comparison to DermFM-Zero Zero-Shot. $^{*}p<0.05$, $^{**}p<0.01$, $^{***}p<0.001$.}
\label{tab:rs2a_performance_by_experience}
\end{table}

\begin{table}[h]
\footnotesize\centering
\setlength{\tabcolsep}{4pt}
\begin{tabular}{lll}
\toprule
Condition & Accuracy & p-value \\
\midrule
Without AI assistance & 0.50 (0.45-0.54) & — \\
\textbf{With AI assistance} & \textbf{0.61 (0.58-0.65)} & $<$0.001$^{***}$ \\
\bottomrule
\end{tabular}
\caption{\textbf{Reader Study 2B: Human-AI collaboration performance on multimodal skin cancer diagnosis}. Comparison of accuracy with and without AI assistance. 95\% CI in parentheses. $^{*}p<0.05$, $^{**}p<0.01$, $^{***}p<0.001$. $p<0.001$ for paired t-test comparing with vs. without AI.}
\label{tab:rs2b_overall_accuracy}
\end{table}

\begin{table}[h]
\footnotesize\centering
\setlength{\tabcolsep}{4pt}
\begin{tabular}{llll}
\toprule
Class & Without AI & With AI & p-value \\
\midrule
NV & 0.55 (0.48-0.62) & \textbf{0.64 (0.58-0.71)} & 0.0559 \\
MEL & 0.69 (0.61-0.77) & \textbf{0.75 (0.68-0.83)} & 0.2286 \\
BKL & 0.35 (0.27-0.43) & \textbf{0.46 (0.38-0.55)} & 0.0501 \\
BCC & 0.64 (0.56-0.71) & \textbf{0.76 (0.69-0.82)} & 0.0150$^{*}$ \\
AKIEC & 0.49 (0.37-0.60) & \textbf{0.61 (0.50-0.73)} & 0.1296 \\
DF & 0.48 (0.35-0.60) & \textbf{0.71 (0.60-0.82)} & 0.0058$^{**}$ \\
INF & 0.19 (0.04-0.35) & \textbf{0.31 (0.13-0.49)} & 0.3464 \\
VASC & 0.73 (0.62-0.84) & 0.72 (0.61-0.83) & 0.8900 \\
SCCKA & 0.37 (0.25-0.48) & \textbf{0.47 (0.36-0.59)} & 0.2016 \\
OTHER\_BEN & 0.26 (0.11-0.42) & \textbf{0.32 (0.16-0.48)} & 0.6010 \\
\bottomrule
\end{tabular}
\caption{\textbf{Reader Study 2B: Class-specific performance on human-AI collaborative multimodal skin cancer diagnosis}. Comparison of accuracy with and without AI assistance for each diagnostic class. Values represent mean accuracy with 95\% CI in parentheses. $^{*}p<0.05$, $^{**}p<0.01$, $^{***}p<0.001$ (paired t-test) compared to performance without AI assistance.}
\label{tab:rs2b_class_specific_accuracy}
\end{table}

\begin{table}[h]
\footnotesize\centering
\setlength{\tabcolsep}{4pt}
\begin{tabular}{llll}
\toprule
Experience Group & Without AI & With AI & p-value \\
\midrule
Non-expert & 0.45 (0.39-0.51) & \textbf{0.59 (0.54-0.64)} & $<$0.0001$^{***}$ \\
Expert & 0.55 (0.48-0.62) & \textbf{0.65 (0.60-0.69)} & 0.0241$^{*}$ \\
\bottomrule
\end{tabular}
\caption{\textbf{Reader Study 2B: Performance by experience level on human-AI collaborative multimodal skin cancer diagnosis}. Comparison of accuracy with and without AI assistance for each experience group. Values represent mean accuracy with 95\% CI in parentheses. $^{*}p<0.05$, $^{**}p<0.01$, $^{***}p<0.001$ (paired t-test) compared to performance without AI assistance.}
\label{tab:rs2b_accuracy_by_experience}
\end{table}

\begin{table}[h]
\footnotesize\centering
\setlength{\tabcolsep}{4pt}
\begin{tabular}{lll}
\toprule
Condition & Appropriate Management Rate & p-value \\
\midrule
Without AI assistance & 0.70 (0.67-0.73) & — \\
\textbf{With AI assistance} & \textbf{0.73 (0.69-0.76)} & 0.0100$^{**}$ \\
\bottomrule
\end{tabular}
\caption{\textbf{Reader Study 2B: Management appropriateness comparison on human-AI collaborative Study}. Comparison of appropriate management rates with and without AI assistance. 95\% CI in parentheses. $^{*}p<0.05$, $^{**}p<0.01$, $^{***}p<0.001$. $p=0.0100$ for paired t-test comparing with vs. without AI.}
\label{tab:rs2b_overall_management}
\end{table}

\begin{table}[h]
\footnotesize\centering
\setlength{\tabcolsep}{4pt}
\begin{tabular}{llll}
\toprule
Experience Group & Without AI & With AI & p-value \\
\midrule
Non-expert & 0.67 (0.62-0.71) & \textbf{0.70 (0.65-0.74)} & 0.0148$^{*}$ \\
Expert & 0.74 (0.70-0.77) & \textbf{0.76 (0.71-0.81)} & 0.1915 \\
\bottomrule
\end{tabular}
\caption{\textbf{Reader Study 2B: Management appropriateness by experience level}. Comparison of appropriate management rates with and without AI assistance for each experience group. Values represent mean appropriate management rate with 95\% CI in parentheses. $^{*}p<0.05$, $^{**}p<0.01$, $^{***}p<0.001$ (paired t-test) compared to performance without AI assistance.}
\label{tab:rs2b_management_by_experience}
\end{table}

\begin{table}[h]
\footnotesize\centering
\setlength{\tabcolsep}{4pt}
\begin{tabular}{lcccc}
\toprule
\textbf{Model} & \textbf{Streaks} & \textbf{Dots and Globules} & \textbf{Pigment Network} & \textbf{Pigmentation} \\
\midrule
\multicolumn{5}{l}{\textit{Precision@10}} \\
CLIP & 0.20 & 0.90 & 0.90 & 0.30 \\
MONET & 0.80 & 0.80 & 0.80 & 0.80 \\
DermFM-Zero & \textbf{1.00} & \textbf{0.90} & \textbf{0.90} & \textbf{0.80} \\
\midrule
\multicolumn{5}{l}{\textit{Precision@30}} \\
CLIP & 0.33 & 0.83 & 0.80 & 0.26 \\
MONET & 0.60 & 0.86 & 0.73 & 0.70 \\
DermFM-Zero & \textbf{0.70} & \textbf{0.96} & \textbf{0.93} & \textbf{0.76} \\
\midrule
\multicolumn{5}{l}{\textit{Precision@50}} \\
CLIP & 0.22 & 0.84 & 0.52 & 0.44 \\
MONET & 0.58 & 0.90 & 0.80 & 0.70 \\
DermFM-Zero & \textbf{0.64} & \textbf{0.94} & \textbf{0.94} & \textbf{0.78} \\
\bottomrule
\end{tabular}
\caption{\textbf{Concept retrieval performance on Derm7pt dataset.} Precision@K scores measuring the proportion of correctly retrieved images containing target dermoscopic concepts among top-K highest-activated images. The dataset contains 827 images with varying concept prevalence (streaks: 40.3\%, dots and globules: 83.8\%, pigment network: 66.6\%, pigmentation: 41.5\%). Bold values indicate best performance for each concept-K combination.}
\label{tab:concept_retrieval}
\end{table}

\begin{table}[t]
\centering
\begin{tabular}{ll}
\hline
\textbf{Category} & \textbf{Setting} \\
\hline
Vision encoder
& PanDerm \\
Text encoder & PubMedBERT \\

Sampling strategy 
& Source-balanced dataset resampling \\

Gradient accumulation
& 4 \\

Batch size & 1024 (effective: 4096) \\

Optimizer 
& AdamW \\

Learning rate 
& $5 \times 10^{-5}$ \\

Weight decay 
& 0.1 \\

Warm-up steps 
& 1000 \\

Training epochs 
& 30 \\

Image augmentation 
& Random resized crop (scale: 0.4--1.0), \\
& color jitter (0.32, 0.32, 0.32, 0.08; prob 0.8), \\
& grayscale (prob 0.2) \\
Learning rate schedule & Cosine decay (with weight decay = 1.0) \\

\hline
\label{tab:pretrain_hyperparams}
\end{tabular}
\caption{\textbf{Key hyper-parameters used for image-text pretraining.}}
\end{table}

\begin{table}[t]
\centering
\small
\begin{tabular}{cl}
\hline
\textbf{ID} & \textbf{Zero-shot Prompt Template} \\
\hline
1 & This is a skin image of \{\textit{disease}\}. \\
2 & A skin image of \{\textit{disease}\}. \\
3 & An image of \{\textit{disease}\}, a skin condition. \\
4 & \{\textit{disease}\}, a skin disorder, is shown in this image. \\
5 & The skin lesion depicted is \{\textit{disease}\}. \\
6 & The skin cancer in this image is \{\textit{disease}\}. \\
7 & This image depicts \{\textit{disease}\}, a type of skin cancer. \\
\hline
\end{tabular}
\caption{\textbf{Zero-shot prompt templates used for skin disease classification.}
Here, \{\textit{disease}\} denotes the target disease or condition name.}
\label{tab:zs_prompt}
\end{table}

\begin{onecolumn}

\footnotesize

\centering

\begin{longtable}{|p{8.5cm}|p{8.5cm}|}

\hline

\cellcolor{gray!25}\textbf{A--C} & \cellcolor{gray!25}\textbf{D--F} \\
\hline
abscess & darier-white disease \\
acanthosis nigricans & dariers disease \\
acne & deep fungal infection \\
acne keloidalis nuchae & degos disease \\
acne urticata & dermatitis \\
acne vulgaris & dermatitis herpetiformis \\
acquired autoimmune bullous diseaseherpes gestationis & dermatofibroma \\
acrodermatitis enteropathica & dermatosis papulosa nigra \\
acrodermatitis infantile papular & desquamation \\
acrokeratosis verruciformis & diffuse xanthoma \\
actinic granuloma & digital fibroma \\
actinic solar damage(actinic keratosis) & dilated pore of winer \\
actinic solar damage(cutis rhomboidalis nuchae) & discoid eczema \\
actinic solar damage(pigmentation) & disseminated actinic porokeratosis \\
actinic solar damage(solar elastosis) & drug eruption \\
actinic solar damage(solar purpura) & drug eruptions \& reactions \\
actinic solar damage(telangiectasia) & drug-induced pigmentary changes \\
acute and chronic dermatitis & dry skin \\
acute constitutional eczema & dry skin eczema \\
acute dermatitis & dyshidrosiform eczema \\
acute dermatitis, nos & dysplastic nevus \\
acute generalized exanthematous pustulosis & ecthyma \\
acute vesicular dermatitis & ecthyma gangrenosum \\
adnexal neoplasm & eczema \\
allergic contact dermatitis & eczema herpeticum \\
allergic reaction & ehlers danlos syndrome \\
alopecia & elephantiasis nostras \\
alopecia areata & epidermal nevus \\
alopecia mucinosa & epidermoid cyst \\
amyloidosis & epidermolysis bullosa \\
angiofibroma & erosion of skin \\
angiokeratoma & erosive pustular dermatosis of the scalp \\
angioma & eruptive odontogenic cyst \\
angular cheilitis & eruptive xanthoma \\
animal bite - wound & erythema ab igne \\
annular erythema & erythema annulare centrifugum \\
apocrine hydrocystoma & erythema craquele \\
arsenical keratosis & erythema dyschromicum perstans \\
atopic dermatitis & erythema elevatum diutinum \\
atopic winter feet & erythema gyratum repens \\
basal cell carcinoma & erythema migrans \\
basal cell carcinoma morpheiform & erythema multiforme \\
beau's lines & erythema nodosum \\huo
becker nevus & exfoliative dermatitis \\
behcets disease & exfoliative erythroderma \\
benign keratosis & factitial dermatitis \\
blister & favre racouchot \\
blue nevus & fibroma molle \\
bowen's disease & fixed drug eruption \\
bullous disease & fixed eruptions \\
bullous pemphigoid & flat wart \\
burn of forearm & flushing \\
burn of skin & follicular mucinosis \\
café au lait macule & folliculitis \\
calcinosis cutis & foot ulcer \\
callus & foreign body reaction of the skin \\
campbell de morgan spots & fox-fordyce disease \\
candida intertrigo & freckle \\
candidiasis & fungal dermatitis \\
cellulitis & fungal dermatosis \\
cheilitis & furuncle \\
chilblain &  \\
cholestasis of pregnancy &  \\
chondrodermatitis nodularis helicis &  \\
chronic actinic dermatitis &  \\
chronic dermatitis, nos &  \\
chronic palmoplantar pustular psoriasis &  \\
clubbing of fingers &  \\
compound nevus &  \\
condyloma &  \\
condyloma acuminatum &  \\
confluent and reticulated papillomatosis &  \\
congenital nevus &  \\
contact dermatitis &  \\
contact dermatitis caused by rhus diversiloba &  \\
contact dermatitis, nos &  \\
contact purpura &  \\
crowe's sign &  \\
cutaneous b-cell lymphoma &  \\
cutaneous horn &  \\
cutaneous larva migrans &  \\
cutaneous leishmaniasis &  \\
cutaneous lupus &  \\
cutaneous sarcoidosis &  \\
cutaneous t cell lymphoma &  \\
cyst &  \\

\cellcolor{gray!25}\textbf{G--I} & \cellcolor{gray!25}\textbf{J--L} \\
\hline
geographic tongue & junction nevus \\
granulation tissue & juvenile plantar dermatosis \\
granuloma annulare & juvenile xanthogranuloma \\
granuloma faciale & kaposi sarcoma \\
grover's disease & kaposi's sarcoma of skin \\
guttate psoriasis & keloid \\
hailey hailey disease & keratoacanthoma \\
halo nevus & keratoderma \\
hand eczema & keratolysis exfoliativa of wende \\
hand foot and mouth disease & keratosis \\
hemangioma & keratosis pilaris \\
hematoma of skin & keratosis pilaris rubra faciei \\
hemosiderin pigmentation of lower limb due to varicose veins of lower limb & kerion \\
hemosiderin pigmentation of skin due to venous insufficiency & knuckle pads \\
herpes simplex virus & koilonychia \\
herpes zoster & langerhans cell histiocytosis \\
hidradenitis suppurativa & leg veins \\
histiocytosis of skin & lentigo \\
hormonal acne & lentigo maligna \\
hyperkeratosis palmaris et plantaris & lentigo maligna melanoma \\
hyperpigmentation & leprosy borderline \\
hypersensitivity & leprosy lepromatous \\
hypertrichosis & leprosy tuberculoid \\
hypertrophic scar & leukocytoclastic vasculitis \\
ichthyosis & leukonychia \\
idiopathic guttate hypomelanosis & lichen amyloidosis \\
immunological disorder & lichen nitidus \\
impetigo & lichen planus \\
infantile atopic dermatitis & lichen sclerosis et atrophicus \\
infected eczema & lichen simplex chronicus \\
inflammatory dermatosis & lichen spinulosus \\
insect bite & lichen striatus \\
intertrigo & lipoma \\
inverse psoriasis & livedo reticularis \\
irritant contact dermatitis & local infection of wound \\
irritated seborrheic keratosis (from "sk/isk") & localized cutaneous vasculitis \\
 & localized skin infection \\
 & lupus erythematosus \\
 & lyme disease \\
 & lymphangioma \\
 & lymphocytic infiltrate of jessner \\

\cellcolor{gray!25}\textbf{M--O} & \cellcolor{gray!25}\textbf{P--R} \\
\hline
majocchi granuloma & palmoplantar pustulosis \\
median nail dystrophy & palpable migrating erythema \\
medication-induced cutaneous pigmentation & papular dermatoses of pregnancy \\
melanin pigmentation due to exogenous substance & parapsoriasis \\
melanocytic nevus & paronychia \\
melanoma & parvovirus b19 infection \\
melasma & pediculosis capitis \\
merkel cell carcinoma & pediculosis lids \\
metastatic carcinoma & pellagra \\
milia & pemphigus foliaceus \\
miliaria & pemphigus vulgaris \\
molluscum contagiosum & phototherapy \\
morphea & phytophotodermatitis \\
mucinosis & pigmentation of pregnancy \\
mucocele & pigmented progressive purpuric dermatosis \\
mucosal melanotic macule & pigmented purpuric eruption \\
muzzle rash & pilar cyst \\
mycosis fungoides & pincer nail deformity \\
myxoid cyst & pityriasis alba \\
naevus comedonicus & pityriasis lichenoides \\
nail disease & pityriasis lichenoides chronica \\
nail dystrophy & pityriasis lichenoides et varioliformis acuta \\
nail psoriasis & pityriasis rosea \\
necrobiosis lipoidica & pityriasis rubra pilaris \\
nematode infection & pityrosporum folliculitis \\
neurodermatitis & poikiloderma \\
neurofibroma & poikiloderma of civatte \\
neurofibromatosis & poisoning by nematocyst \\
neutrophilic dermatoses & polymorphic eruption of pregnancy \\
nevus & polymorphous light eruption \\
nevus sebaceous of jadassohn & porokeratosis \\
nevus spilus & porokeratosis actinic \\
no definitive diagnosis & porokeratosis of mibelli \\
norweigian scabies & poroma \\
nummular eczema & porphyria \\
onycholysis & port wine stain \\
onychomycosis & post-inflammatory hyperpigmentation \\
onychoschizia & post-inflammatory hypopigmentation \\
 & post-inflammatory pigmentation \\
 & pressure ulcer \\
 & prurigo \\
 & prurigo nodularis \\
 & prurigo pigmentosa \\
 & pruritic urticarial papules and plaques of pregnancy \\
 & pruritus ani \\
 & pseudofolliculitis barbae \\
 & pseudorhinophyma \\
 & psoriasis \\
 & pustular psoriasis \\
 & pyoderma \\
 & pyoderma gangrenosum \\
 & pyogenic granuloma \\
 & radiodermatitis \\
 & raynaud phenomenon \\
 & red stretch marks \\
 & reiter's syndrome \\
 & relapsing polychondritis \\
 & rheumatoid nodule \\
 & rhinophyma \\
 & riehl melanosis \\
 & rosacea \\

\cellcolor{gray!25}\textbf{S--U} & \cellcolor{gray!25}\textbf{V--X} \\
\hline
sand-worm eruption & varicella \\
sarcoidosis & varicose veins of lower extremity \\
scabies & vascular \\
scalp psoriasis & vasculitis \\
scar & venous lake \\
scleroderma & verruca vulgaris \\
scleromyxedema & viral exanthem \\
sebaceous hyperplasia & viral exanthems: roseola \\
seborrheic keratoses & vitiligo \\
sixth disease & wound/abrasion \\
skin and soft tissue atypical mycobacterial infection & xanthelasma \\
skin cancer & xeroderma pigmentosum \\
skin diseases caused by warts & xerosis \\
skin infection & xerotic eczema \\
skin lesion in drug addict &  \\
skin tag &  \\
solid cystic basal cell carcinoma &  \\
spider veins &  \\
squamous cell carcinoma &  \\
staphylococcal scalded skin syndrome &  \\
stasis dermatitis &  \\
stasis edema &  \\
stasis ulcer &  \\
steatocystoma multiplex &  \\
steroid acne &  \\
steroid use abusemisuse dermatitis &  \\
stevens-johnson syndrome &  \\
strawberry birthmarks &  \\
striae &  \\
subungual hematoma &  \\
sun spots &  \\
sunburn &  \\
superficial gyrate erythema &  \\
superficial spreading melanoma ssm &  \\
superficial wound of body region &  \\
sweet syndrome &  \\
sweet’s syndrome &  \\
syphilis &  \\
syringoma &  \\
systemic disease &  \\
telangiectasia macularis eruptiva perstans &  \\
tick bite &  \\
tinea &  \\
tinea corporis &  \\
tinea cruris &  \\
tinea faciale &  \\
tinea manus &  \\
tinea pedis &  \\
tinea versicolor &  \\
transient acantholytic dermatosis &  \\
traumatic blister &  \\
traumatic ulcer &  \\
tuberous sclerosis &  \\
tungiasis &  \\
ulcer &  \\
unilateral laterothoracic exanthem &  \\
urticaria &  \\
urticaria pigmentosa &  \\
urticarial vasculitis &  \\
\hline
\caption{\textbf{Complete alphabetical listing of skin conditions in the training dataset.}}
\label{tab:all_disease}
\end{longtable}

\end{onecolumn}
\clearpage

\Heading{References}
\begin{spacing}{0.9}
\bibliographystyle{naturemag}
\bibliography{main}
\end{spacing}

\end{document}